# *Towards Interpretable Physical-Conceptual Catchment-Scale Hydrological Modeling using the Mass-Conserving-Perceptron*

Yuan-Heng Wang[1,2] and Hoshin V. Gupta[1]

[1] Department of Hydrology and Atmospheric Science, The University of Arizona, Tucson, AZ
[2] Earth and Environmental Sciences Area, Lawrence Berkeley National Laboratory, Berkeley, CA

Corresponding Author: Yuan-Heng Wang, Ph.D.
Email: yhwang0730@gmail.com | YuanHengWang@lbl.gov | yhwang0730@arizona.edu

## Abstract

We investigate the applicability of machine learning technologies to the development of parsimonious, interpretable, catchment-scale hydrologic models using directed-graph architectures based on the mass-conserving perceptron (MCP) as the fundamental computational unit. Here, we focus on architectural complexity (depth) at a single location, rather than universal applicability (breadth) across large samples of catchments. The goal is to discover a minimal representation (numbers of cell-states and flow paths) that represents the dominant processes that can explain the input-state-output behaviors of a given catchment, with particular emphasis given to simulating the full range (high, medium, and low) of flow dynamics. We find that a "*HyMod Like*" architecture with three cell-states and two major flow pathways achieves such a representation at our study location, but that the additional incorporation of an input-bypass mechanism significantly improves the timing and shape of the hydrograph, while the inclusion of bi-directional groundwater mass exchanges significantly enhances the simulation of baseflow. Overall, our results demonstrate the importance of using multiple diagnostic metrics for model evaluation, while highlighting the need for properly selecting and designing the training metrics based on information-theoretic foundations that are better suited to extracting information across the full range of flow dynamics. This study sets the stage for interpretable regional-scale MCP-based hydrological modeling (using large sample data) by using neural architecture search to determine appropriate minimal representations for catchments in different hydroclimatic regimes.

## Plain Language Summary

We show that conventional machine learning technologies can be used to develop parsimonious, interpretable, catchment-scale hydrologic models using the mass-conserving perceptron (MCP) as a fundamental computational unit. Using data from the Leaf River Basin, we test a variety of minimal, dominant process, representations that can explain the input-state-output dynamics of the catchment. Our results demonstrate the importance of using multiple diagnostic metrics for evaluation and comparison of different model architectures, and highlight the importance of choosing (or designing) objective functions for model training that are properly suited to the task of extracting information across the full range of flow dynamics. This depth-focus study sets the stage for interpretable regional-scale MCP-based hydrological modeling (using large sample data) by using neural architecture search to determine appropriate minimal representations for catchments in different hydroclimatic regimes.

# 1. Introduction and Scope

[1]    A spatially-lumped catchment-scale conceptual rainfall-runoff (CRR) model is a simplified representation of the complicated natural system that dynamically partitions rainfall into runoff, evapotranspirative loss, and moisture stored within the system (e.g., as groundwater); see for example the *Sacramento Soil Moisture Accounting Model* (SAC-SMA; *Burnash & Ferral, 1973*). The development of CRR models dates back more than half a century; for example, *Nash (1957)* proposed the idea that the RR process can be represented as a sequential storage-release process that can be modeled using several (linear) reservoirs in series. More generally, the use of simple mass-conserving buckets has become widespread for CRR model development (*Gupta & Sorooshian, 1983; Gupta & Sorooshian, 1985; Singh, 1988*) with emphasis placed on representing different dominant processes, such as percolation to the subsurface groundwater system (*Burnash & Ferral, 1973; Sorooshian & Gupta, 1983*), and surface routing (*Boyle, 2000*), etc. (see *Bergström, 1976; Perrin et al., 2003*). Such models can generally be viewed as directed-graph-like structures made up of computational units (nodes), and variations of the directed-graph architectures can be used to explore alternative hypotheses regarding how the internal workings of a catchment can be usefully conceptualized (*Gupta & Nearing, 2014*). In a nutshell, CRR models provide simplified, lumped descriptions of the dominant sub-watershed-scale processes that act to generate the overall watershed-scale hydrologic response of the system (*Singh, 1988; Gupta et al., 2003; Gupta et al., 2012*). Such models are widely used in practice for a variety of purposes ranging from operational flood forecasting (*Lindström et al., 2010; Pagano et al., 2016*) to climate change impact studies (*Ji et al., 2023*) and forecasting probable changes to streamflow under anticipated changes to climate (*Schiermeier 2011*).

[2]    As might be expected, CRR modeling frameworks are under continuous development (*Harrigan et al., 2023*), thereby representing an important approach for addressing the "one-size-fits-all" problem by using the method of multiple working hypotheses (e.g. *Clark et al., 2008; Fenicia et al., 2011; Prieto et al., 2021*) to deal with uncertainty regarding appropriate/adequate model structure (*Gupta et al., 2012*). By extension to spatially-distributed configurations, and by incorporating a greater variety of physical processes, such frameworks can enable timely and accurate flood forecasting over larger spatiotemporal domains; see for example, the US National Water Model (*Salas et al., 2018*), the Raven software framework (*Craig et al., 2020*), and the Global Flood Awareness System (*Alfieri et al., 2013*). Meanwhile, such frameworks, which are based on the use of prior physical-conceptual (PC) knowledge to specify model architectures and process relationships, face strong competition from data-based machine learning (ML) approaches which seek to '*learn*' the appropriate model architectures, and relationships, directly from data.

[3]    Although simple ML-based models, such as the time-delay artificial neural network (ANN), have (since at least the mid 1990's) been clearly shown to outperform PC-based modeling approaches when applied to RR modeling at individual locations (e.g., *Hsu et al., 1995*), a particularly important challenge occurred recently with the demonstration that the *Long Short-Term Memory* network (LSTM; *Hochreiter & Schmidhuber, 1997*) – originally utilized mainly for sequence prediction in the context of Natural Language Processing – can provide a serious alternative to PC-based modeling when applied to catchment-scale RR modeling over continental extents by simultaneously extracting information from large numbers of catchments (*Kratzert et al., 2018; Kratzert et al., 2019a; Kratzert et al., 2019b; Feng et al., 2020; Arsenault et al., 2023*) and, in particular, can help to address the "*prediction in ungauged basins*" problem (PUB; *Sivapalan et al., 2003; Hrachowitz et al., 2013*).

[4]    Notwithstanding such success, the application of ML-based modeling to the domain sciences has been critiqued as being able to provide only limited insight into the underlying nature of the physical processes that determine the input-state response of a physical system (*Jiang. et al., 2020*). Authors have highlighted the dangers of relying on modeling systems that lack strong physical interpretability (*Reichstein et al., 2019; Fleming et al., 2021*), resulting in research that attempts to improve the physical interpretability of ML-based representations while still retaining the ability to learn system structures and process relationships directly from data (*Jiang. et al., 2022; De la Fuente et al., 2024; Wang & Gupta 2024*).

[5]    Broadly speaking, the efforts to incorporate PC knowledge/constraints into ML-based learning are referred to as "*physics-informed machine learning*" (PIML). Such efforts include postprocessing of the model outputs (*Jiang et al., 2020; Frame et al., 2021; Zhong et al., 2023*) or internal storage variables (*Kapoor et al., 2023*) generated by physically-based models as additional inputs to ML-based models (*Nearing et al., 2020a*), modifying the training loss

function to penalize physically inconsistent behaviors (*Read et al., 2019; Pokharel et al., 2023*), embedding physical constraints directly into the ML-based architectures (*Hoedt et al., 2021; Quiñones et al., 2021; Sturm and Wexler, 2022*), embedding physics layers into a tensor network (*Wang et al., 2020; Mahesh et al., 2022; Nazari et al., 2022; Feng et al., 2023c*) or, conversely, embedding neural network layers into physically-based models (*Bennett & Nijssen, 2021*). We refer the reader to review articles such as *Ibrahim et al. (2022), Tan et al. (2022)*, and *Ng et al. (2023)* for more information.

[6]   One major technical advance that has made the rapid development of ML technology possible is the availability of automated differentiable programming (ADP; see discussion by *Shen et al., 2023*).  ADP has made it possible for scientists to rapidly construct and train computational model codes using the back-propagation optimization methodology (*LeCun et al., 1988; Rummelhart, et al., 2013*), and serves as an important bridge between the physics-based and ML-based modeling research communities. As a consequence, any CRR model can be selected as the backbone for a model architecture that can be then improved by inserting neural network layers of various kinds (e.g., ANN, RNN, or CNN layers) to inform the model parameterization (*Tsai et al., 2021; Kraft et al., 2022; Acuña Espinoza et al., 2024*) or to develop substitutes for poorly understood process relationships (*Bhasme et al., 2022; Feng et al., 2022; Höge et al., 2022; Feng et al., 2023b*). Arguably, this enhancement of functional expressivity (if the complexity is well-managed; *Weijs & Ruddell, 2020*), can help to properly regularize the model and improve its ability to learn context-dependent physically-interpretable process relationships from data.

[7]   In this regard, it is worth noting that hydrological modeling efforts over the past decade or so have been driven primarily by a focus on either depth or breadth (*Gupta et al., 2014*). The depth focus seeks to improve process understanding by concentrating the investigative effort on the development of detailed and physically-realistic system representations at specific locations (catchments), whereas the breadth focus seeks to develop models that are generally applicable at multiple locations over large extents (e.g., at the continental and even global scales). Arguably, most of the recent ML-based efforts associated with streamflow prediction have a breadth focus (*Sun et al., 2021; Kratzert et al., 2023; Nearing et al., 2024*). It would seem sensible to balance this with an equal focus on depth (and interpretability) while remaining within the context of the strengths offered by modern ML.

[8]   In this regard, *Wang & Gupta (2024)* recently proposed the formulation for a physically-interpretable computational unit, referred to as the "*Mass-Conserving Perceptron*" (MCP), with the idea that such a unit could form the basis for parsimonious physically-interpretable modeling of dynamical geoscientific systems using off-the-shelf ML technology. The MCP exploits the inherent isomorphism between the directed graph structures underlying both PC models and generic gated RNNs (*Gupta & Nearing, 2014*) to explicitly represent the mass-conserving nature of physical processes while enabling the functional nature of such processes to be directly learned from available data. *Wang & Gupta (2023)* demonstrated that various CRR model architectures based on only a single MCP node (representing a single mass-conserving state variable) can offer comparable performance to several data-based benchmarks when provided with the same input information, while maintaining the same level of physical interpretability that is expected from a more traditional physical-conceptual model. They proposed that the MCP can serve to help unify the PC-based and ML-based modeling approaches, thereby facilitating the development of next-generation physics-informed CRR models by allowing different parsimonious system hypotheses to be progressively tested during model development (*Gong et al., 2013; Nearing et al., 2020a*).

[9]   In this paper, we extend upon *Wang & Gupta (2024)* by using the physically-interpretable MCP computational units as building blocks to construct and test architectural hypotheses that rival the complexity of traditional CRR models, and thereby explore the development of different compound hypotheses regarding the dominant internal dynamics of RR system behaviors. By conceptualizing the RR system as a directed graph architecture consisting of a network of MCP nodes (representing subcomponents of the system) and links (representing the associated flow paths), we propose and test several plausible *"Mass-Conserving Architectures"* (MCAs), and investigate their ability to simulate the input-state-output dynamics of the *Leaf River* Basin.  Specifically, we investigate how different hypotheses regarding system architecture contribute to different aspects of model performance by enhancing or diminishing model performance (*Nearing & Gupta, 2015; Gharari et al., 2021*). As in *Wang & Gupta (2024)* all of the model development, training and testing was implemented in Python version 3.7 using the PyTorch open-source ML differentiable programming framework (*Paszke et al., 2019*). This allowed us to take advantage of the advanced programming resources that PyTorch makes available, including use of optimization algorithms that use backpropagation (*Rumelhart, 1986*) to facilitate efficient parameter learning.

[10] In the next section (Section 2) we briefly review the important features of the MCP unit and how its utility can be enhanced by adding more functional components. Section 3 discusses use of the MCP unit for development of *"HyMod Like"* hydrologic model architectures. Section 4 summarizes the experimental design and procedures for training and evaluation. Comparative results for a variety of proposed architectural hypotheses having different numbers of state variables (nodes) and flow paths (links) are presented in Section 5. Sections 6 and 7 explore enhancements to model performance by restricting capacity of the soil-moisture tank (by applying an input-bypass gate), and by enabling bi-directional mass exchanges with the regional groundwater system. We present the results of model benchmarks in Section 8, summarize our findings and discuss future directions in Section 9.

## 2. The Mass-Conserving Perceptron

### 2.1 Basic Architecture of the MCP

[11] The mass-conserving perceptron (MCP; *Wang & Gupta, 2024*) is an ML-based physically-interpretable computational unit that is isomorphically similar to a single node of a generic gated recurrent neural network, but is different in that it enables mass flows to be conserved at the nodal level. We describe its main aspects briefly below. For a detailed discussion of the architecture, expressive capability, and interpretable nature of the MCP, see *Wang & Gupta (2024).*

[12] *Figure 1a* illustrates the architecture of the MCP node. The node implements a representation of mass-conservative system dynamics via the discrete time update equation:

$$X_{t+1} = X_t - O_t - L_t + U_t \tag{1}$$

whereby the mass state $X_{t+1}$ of the system (node) at time step $t + 1$ is computed by adding the mass of input $U_t$ that enters the node, and subtracting the masses of outputs $O_t$ and $L_t$ that leave the node, during the time interval from $t$ to $t + 1$. For example, in the context of spatially-lumped catchment-scale RR modeling, $U_t$ can represent the precipitation mass input, and $L_t$ and $O_t$ can represent the evapotranspirative and streamflow mass outputs from the system represented by the node.

[13] We further assume that the output $O_t$ and loss $L_t$ depend on the value of the state $X_t$ through the process parameterization equations $O_t = G_t^O \cdot X_t$ and $L_t = G_t^L \cdot X_t$, where $G_t^O$ and $G_t^L$ are context-dependent (see later) time-varying "*output*" and "*loss*" conductivity gating functions respectively, so that Eqn (1) can be rewritten as:

$$X_{t+1} = X_t - G_t^O \cdot X_t - G_t^L \cdot X_t + U_t \tag{2a}$$

$$X_{t+1} = G_t^R \cdot X_t + U_t \tag{2b}$$

where $G_t^R$, referred to hereafter as the "*remember*" gate, represents the fraction of the state $X_t$ that is retained by the system from one time step to the next. Note that to ensure physical realism, we require that the time-evolving values of each of these gates ($G_t^O$, $G_t^L$ and $G_t^R$) must remain both non-negative and less than 1.0 at all times. Further, to ensure conservation of mass we require that $G_t^R + G_t^O + G_t^L = 1$, which means that the remember gate is computed from knowledge of the output and loss gates as $G_t^R = 1 - G_t^O - G_t^L$ (thereby placing a strict constraint on the relative values that $G_t^O$ and $G_t^L$ can take on). Now, assuming knowledge of the initial mass state of the system $X_0$, and given the time history of inputs $U_1, ..., U_t$, Eqn (2) can be used to sequentially update the state $X_t$ of the system if the time-evolving values of the gating functions $G_t^O$ and $G_t^L$ (and therefore $G_t^R$) are also provided.

[14] Given this construction, it is clear that the nature of the time evolution of the gating functions $G_t^O$ and $G_t^L$ completely determines the dynamical behavior of the system. Since these gating functions are, in general, not known a priori, the solution proposed by *Wang & Gupta (2024)* was to parameterize these functions using ML architectures so that their functional forms can be learned directly from available data. For example, in the context of spatially-lumped catchment-scale RR modeling, we can take advantage of prior knowledge (in the form of physical understanding), to parameterize the time evolving value of the output gate $G_t^O$ as being determined by the system state $X_t$, so that $G_t^O = f_{ML}^O(X_t)$ where $f_{ML}^O(\cdot)$ is implemented as an ML architecture that can '*learn*' the functional form of the dependence of $G_t^O$ on $X_t$. Similarly, we can parameterize the loss gate $G_t^L$ as being determined by both the system state $X_t$ (as a measure of water availability) and by potential evapotranspiration $PE_t$ (as a measure of

atmospheric and vegetative demand), so that $G_t^L = f_{ML}^L(X_t, PE_t)$ where $f_{ML}^L(\cdot)$ is implemented as an ML architecture that can '*learn*' the functional form of the dependence of $G_t^L$ on both $X_t$ and $PE_t$.

[15] Of course, alternative functional dependencies can also be learned – such as using near-surface air temperature and/or wind speed instead of $PE_t$ in the computation of $G_t^L$ – which facilitates both the use of alternative sources of information (data) and the testing of alternative hypotheses regarding which variables are important. Further, the choice of ML architecture used to parameterize the gates can reflect both the nature of the available data, and also the degree of complexity and functional nature of the dependency – it can be as simple as a decision tree or shallow feedforward ANN, or as complex as a deep neural network (DNN) incorporating *convolution*, *recurrence*, and/or *attention* mechanisms as appropriate. In other words, by parameterizing the gating functions using trainable ML architectures, we provide the system with the ability to learn how the values of the conductivity functions depend on context.

### 2.2 More Complex MCP Architectures

[16] **Figure 1a** also illustrates some additional features of the MCP node. By inclusion of an "*input*" gate $G_t^U$, Eqn (2b) can be written as:

$$X_{t+1} = G_t^R \cdot X_t + G_t^U \cdot U_t \tag{3}$$

where $G_t^U = f_{ML}^U(\dots)$ is implemented as an ML architecture that can '*learn*' the functional form of the dependence of $G_t^U$ on context. By fixing $G_t^U = 1$ we recover the simpler case discussed above, in which all of the incoming precipitation $U_t$ enters the system and is directly added to the system state $X_t$. However, by allowing $G_t^U$ to depend on the state $X_t$ one can represent a "*saturation excess*" mechanism while allowing $G_t^U$ to depend on both the state $X_t$ and input $U_t$ one can represent an "*infiltration excess*" mechanism, both of which can cause incoming mass to bypass the MCP node. As discussed by *Wang & Gupta (2024),* input gating can also be used to implement a precipitation-bias-correction mechanism to compensate for systematic biases in the input measurement/observational process.

[17] Finally, **Figure 1a** illustrates how the concept of learnable gating can be used to represent the possibility of unobserved mass exchanges with the external environment, such as bi-directional groundwater exchanges. This can be accomplished by implementing an additional gate, referred to as the mass relaxation gate ($G_t^{MR}$). Unlike with the output and loss gates, the range of $G_t^{MR}$ is allowed to vary on $(-1, +1)$ to facilitate both inflows to and outflows from the node (system state $X_t$). Further, its value is constrained such that $G_t^{MR} \leq G_t^R$ so that the system cannot lose more water than is actually available. In the absence of other contextual information, $G_t^{MR}$ can be parameterized as simply depending on the system state $X_t$, such that $G_t^{MR} > 0$ (outflows occur) when the system is very "wet" ($X_t$ exceeds some threshold level $X_t^*$) and $G_t^{MR} < 0$ (inflows occur) when the system is very "dry" ($X_t$ is below the threshold level $X_t^*$). For implementational details, please see *Wang & Gupta (2024).*

### 2.3 Desirable Features of the MCP

[18] The MCP discussed above has been designed to have several desirable features that make it suitable for interpretable physical-conceptual modeling of dynamical systems such as are of interest in hydrology. These features include:

1) Recurrence, enabling the dynamical evolution of the system state (memory) to be represented.

2) The ability to impose conservation principles at the nodal level, as constraints on system evolution.

3) The ability to represent and learn the dynamics of unobserved gains/losses of mass from the system.

4) The ability to learn the forms of the process parameterization equations (gating functions) that govern the dynamical behaviors of the system based on context (current and past conditions).

5) Ease of implementation using off-the-shelf ML technology such as PyTorch in Python (*Paszke et al., 2019*).

## 3. Use of the MCP to Construct Multi-Node Architectural Hypotheses

[19] In previous work (*Wang & Gupta, 2024),* we explored the behavioral expressivity, interpretability, and performance achievable by a *single* MCP node enabled by the learnable gating mechanism. In that very simple

model architecture, the system dynamics are entirely represented by a single cell-state with only one flow-pathway between the cell-state and the observed output (streamflow at the catchment outlet). The focus of that study was on testing gating mechanisms of various levels of complexity (constant, sigmoid, shallow feedforward ANNs), and varying degrees of context dependence, to explore various scientific hypotheses regarding the level of functional complexity necessary to accurately model the input-state-output dynamics of the *Leaf River* basin in Mississippi, USA. Therefore, *all* of the architectural complexity was expressed through the (learned) forms of the gating functions, whose functional forms were varied from simple *time-constant* gating to *time-variable context-dependent Sigmoid-based* gating, and then to more complex *time-variable context-dependent ANN-based* gating.

[20]  In this paper, we instead explore the value of constructing more complex, multiple-node (multi-cell-state), directed-graph system architectures using various versions of the MCP node as building blocks. As before, our interest is in developing *interpretable* physical-conceptual representations of the system that can combine the prior information available from theory with the new information available in data by exploiting the power of ML (including programming and training end-to-end using standard ML-development differentiable programming environments such as PyTorch, so that future developments can benefit from the progressive rapid development thereof).

### 3.1 Example of a Multiple-Node HyMod Like Architecture

[21]  As a simple example consider the two-flow-path *HyMod Like* architecture shown in **Figure 1b**, that is composed of three MCP-based computational units. This architecture consists of an initial node (blue outline) representing surface soil moisture ($X_t^{sm}$) that receives precipitation ($U_t^P$) as input and computes evapotranspiration loss ($L_t^E$), vertical output ($OV_t^{sm}$) as recharge to groundwater, and horizontal output ($OH_t^{sm}$) as flow to the channel. The vertical recharge becomes input to a groundwater node ($X_t^{gw}$) that computes horizontal groundwater flow ($O_t^{gw}$), while the horizontal outflow becomes input to a channel node ($X_t^{ch}$) that computes horizontal channel flow ($O_t^{ch}$). The sum of horizontal channel flow and groundwater flow becomes streamflow $O_t$ at the watershed outlet ($O_t = O_t^{ch} + O_t^{gw}$).

[22]  To implement this *HyMod Like* architecture, we need two kinds of MCP-type nodes as illustrated by **Figure 1c** and **Figure 1d**. The soil moisture node (blue box, **Figure 1c**) extends the basic MCP-node discussed in section 2.1 to include an additional recharge output ($OV_t^{sm}$) and its learnable gating mechanism $G_t^{OVsm}$. This results in four time-varying context-dependent gates to be learned ($G_t^{OHsm}$, $G_t^{LEsm}$, $G_t^{OVsm}$ and $G_t^{Rsm}$), where mass conservation requires that all of the gates are all constrained on [0,1] while the soil moisture remember gate $G_t^{Rsm}$ is additionally constrained to satisfy $G_t^{Rsm} = (1 - G_t^{OHsm} - G_t^{OVsm} - G_t^{LEsm})$. Meanwhile the groundwater storage node and the channel routing node are conceptualized (green box, **Figure 1d**) as simplifications of the basic MCP-node by assuming that no mass loss occurs from either of them (all of the mass loss from the system occurs as evapotranspiration from the surface soil moisture node). Accordingly, the groundwater storage remember gate must satisfy $G_t^{Rgw} = (1 - G_t^{Ogw})$ and the channel routing remember gate must satisfy $G_t^{Rch} = (1 - G_t^{Och})$.

[23]  As was done in *Wang & Gupta, (2024),* we can also implement an additional constraint on the system to ensure that the actual evapotranspirative loss $L_t^E$ remains less than or equal to potential evapotranspiration demand ($D_t^E$) at any given time ($L_t^E \leq D_t^E$). This is achieved by computing a *"physically constrained"* loss gate $G_t^{LE_{sm}^{con}}$ using Eqn (4):

$$G_t^{LE_{sm}^{con}} = G_t^{LE_{sm}} - ReLU(G_t^{LE_{sm}} - \frac{D_t^E}{X_t}) \tag{4}$$

where the rectified linear unit *ReLU* operation is defined as $ReLU = f(x) = \max(0, x)$ *(Liu, 2017),* and then using $G_t^{LE_{sm}^{con}}$ in place of $G_t^{LE_{sm}}$ when computing evapotranspirative loss $L_t^E$ (i.e., using $L_t^E = G_t^{LE_{sm}^{con}} \cdot X_t^{sm}$).

[24]  Overall, therefore, this *HyMod Like* architecture consists of two input fluxes ($U_t^P$ and $D_t^E$), three latent state variables ($X_t^{sm}$, $X_t^{gw}$ and $X_t^{ch}$), four intermediate fluxes ($OV_t^{sm}$, $OH_t^{sm}$, $O_t^{gw}$, $O_t^{ch}$) and one output flux ($O_t$). Using observed time-series data for $U_t^P$, $D_t^E$ and $O_t$, our system identification task is to learn functional forms for the five

gating functions $G_t^{OH_{sm}}$, $G_t^{LE_{sm}}$, $G_t^{OV_{sm}}$, $G_t^{Ogw}$ and $G_t^{Och}$ such that the model (as architecturally conceived) is able to reproduce (as well as possible) the observed input-state-output dynamics encoded in the available data.

[25] Clearly a variety of alternative directed graph architectures can be constructed and used to test different possible scientific hypotheses about the internal dynamics of the system; the *HyMod Like* architecture described above is simply one of many possibilities. In Section 3.2, we design and discuss several progressively more complex system architectural hypotheses (constructed using MCP-based nodal units) that we then train (Sections 4 and 5) using real data from the *Leaf River* Basin to assess how much structural complexity is identifiable from (i.e., supported by) the available data.

### 3.2 MCP-Based Architectural Hypotheses Explored in this Study

[26] In this work, we draw upon the fact that most of the behavioral and performance gains in *Wang & Gupta (2024)* were achieved using context-dependent sigmoid-based gating (the benefits of ANN-based gating were important but relatively small), and therefore use only either *Time-constant* or *Sigmoid-based* gating in the experiments reported here. For instance, the output (loss) conductivity function can be assumed to either be a constant value, or to depend (in monotonic non-decreasing fashion, as required by thermodynamic principles) on the current cell state (and/or PET) in a manner determined by the sigmoid function. This allows us to shift the focus from the behavioral expressivity and performance of each node (see *Wang & Gupta, 2024)* to that achievable using architectures involving multiple cell-states (nodes) and numerous flow paths (connections). While more complex behavioral dynamics could certainly be achieved by using *ANN-based* or other forms of gating (see *Wang & Gupta, 2024*), we leave such explorations for future research. To reiterate, the idea behind the MCP-based approach is that MCP-based *PC* models that are *understandable* (in the same sense that traditional PC-based models are understandable) can be conceived of and constructed using *interpretable* components based in established ML concepts. Such models can then be implemented and trained using standard ML technologies (backpropagation, etc.), thereby enabling rapid testing and comparison of a variety of alternative hypotheses regarding system structure and functioning.

[27] For the purposes of this study, we consider the six progressively more complex model (network) architectures illustrated in *Figure 2* and *Table 1*, referred to as $MA_1$ to $MA_6$. All of these architectures are constructed using MCP-based nodal units (as illustrated by the *HyMod Like* architectural example described in Section 2.4) with *Sigmoid-based* gating. While these six architectures are not exhaustive of the possibilities that can be constructed, they represent a variety of conceptualizations that explore various hypotheses regarding the input-state-output dynamics of the *Leaf River* catchment. Further, as in the *HyMod Like* example discussed above, we account for all evapotranspirative loss in the first node, and constrain it to not exceed potential evapotranspirative demand.

a) $MA_1$: This model consists of a single-cell-state and a single-flow-path architecture, and is identical to the model $MC\{O_\sigma L_\sigma^{con}\}$ discussed in *Wang & Gupta (2024)*. Its' purpose is to serve as a baseline for assessing the benefits achievable using more complex architectures.

b) $MA_2$: This model is obtained from $MA_1$ by adding a second output flow path to streamflow. While it has only one cell-state, the second flow-path can route water at a different rate (controlled by an additional learnable output gate). The purpose is to examine whether a better representation of the input-state-output dynamics of the precipitation-runoff process can be achieved by including more than one parallel flow path to streamflow. Conceptually, one flow path can be interpreted as a "*quicker*" flow pathway (e.g., overland and/or interflow) while the other can be interpreted as a "*slower*" flow pathway (e.g., subsurface flow).

c) $MA_3$: This model is obtained from $MA_1$ by routing the output of the first ($MA_1$) node through a second cell-state (node) arranged in series. It therefore consists of two cell-states, but only one flow-path to streamflow. Conceptually, the first node can be interpreted as performing the function of partitioning incoming precipitation into soil-moisture storage, evapotranspirative loss, and streamflow output, with its cell-state representing lumped-average catchment soil moisture storage. Meanwhile, the second node can be interpreted as performing the function of routing, thereby enabling greater flexibility to reproduce hydrograph timing and shape.

d) $MA_4$: This model is obtained from $MA_2$ by passing one of the two flow paths through a second cell-state (node) whose output is then routed to streamflow. It therefore consists of two cell-states and two flow-paths. Conceptually, the first node can be interpreted as partitioning incoming precipitation into "*surface zone*" soil-moisture storage, evapotranspirative loss, and streamflow output, while the second node can be interpreted as a "*subsurface*" (groundwater aquifer) store that can sustain baseflow in the river during dry periods when the storage content of the first node has been depleted (*Gupta and Sorooshian, 1983*).

e) $MA_5$: This model is obtained from $MA_4$ by passing <u>*each*</u> of the two flow paths through different cell-states (nodes) whose outputs are then routed to streamflow. It therefore consists of three cell-states and two flow-paths. Conceptually, the node on one of these flow paths can be interpreted as performing the function of streamflow routing, while the node on the other flow path can be interpreted as a subsurface store that can sustain baseflow.

f) $MA_6$: This model is obtained from $MA_5$ by adding a third output flow path that goes directly to streamflow. It therefore consists of three cell-states and three flow-paths. Conceptually, this direct path can be interpreted as representing overland flow, while the other two (cell-state-mediated) paths represent streamflow routing and subsurface (groundwater) flow.

# 4. Experimental Setup

## 4.1 Study Site and Data Management

[28] All of the experiments reported here were conducted using the *Leaf River* data set (compiled by the US National Weather Service), which consists of 40 years (WY 1949-1988) of daily data from the humid, 1944 $km^2$, *Leaf River Basin* (LRB) located near Collins in southern Mississippi, USA (*Sorooshian et al., 1983*). The dataset consists of cumulative daily values of observed mean areal precipitation ($PP$; mm/day), potential evapotranspiration ($PET$; mm/day), and watershed outlet streamflow ($QQ$; mm/day). It has been widely used in many studies by the hydrological science community for many decades for model development and testing.

[29] We follow the data-splitting procedure reported in our earlier submission regarding the development of a single MCP unit (*Wang & Gupta, 2024*) and adopt the robust data allocation method proposed and tested by (*Zheng et al., 2022*) that partitions the data ($\mathcal{D}$) to ensure distributional consistency of the observational streamflow records across three subsets of the data to be used for *training* ($\mathcal{D}_{train}$), *selection* ($\mathcal{D}_{select}$), and *testing* ($\mathcal{D}_{test}$). This helps to ensure consistent model performance across each of three independent sets (*Chen et al., 2022; Maier et al., 2023*), and to reasonably neglect the need for procedures such as k-fold cross validation. Accordingly, $\mathcal{D}_{train}$ is used to learn the values of the parameters/weights of the gating functions, $\mathcal{D}_{select}$ is used to select the best model among several trained using different random seeds for parameter initializations, and $\mathcal{D}_{test}$ is used for independent evaluation of the generalization ability of the trained model. The data is partitioned based on the ratio of 2:1:1 resulting in a training subset consisting of 7,306 timesteps, and selection and testing subsets consisting of 3,652 time-steps each.

## 4.2 Metrics Used for Training and Performance Assessment

[30] The metric used for model training was the *Kling-Gupta Efficiency* ($KGE$; Eqn. 5) (*Gupta et al., 2009*) that has been well-established as a suitable training metric by the hydrologic science community (*Gauch et al., 2023*). Each model architecture was trained 10 times by random initialization of the parameters, from which the one having the highest scaled $KGE$ score ($KGE_{ss}$; Eqn. 6) (*Khatami et al., 2020*) computed on the *selection* set was retained. Performance assessment was conducted using $KGE_{ss}$ and the components of $KGE$ (Eqns. 7-9):

$$KGE = 1 - \sqrt{((\rho^{KGE} - 1)^2 + (\beta^{KGE} - 1)^2 + (\alpha^{KGE} - 1)^2)} \tag{5}$$

$$KGE_{ss} = 1 - \frac{(1-KGE)}{\sqrt{2}} \tag{6}$$

$$\alpha^{KGE} = \frac{\sigma_s}{\sigma_o} \tag{7}$$

$$\beta^{KGE} = \frac{\mu_s}{\mu_o} \tag{8}$$

$$\rho^{KGE} = \frac{Cov_{so}}{\sigma_s \sigma_o} \tag{9}$$

where $\sigma_s$ and $\sigma_o$ are the standard deviations, and $\mu_s$ and $\mu_o$ are the means, of the corresponding data-period simulated and observed streamflow hydrographs respectively and, similarly, $Cov_{so}$ is the covariance between the simulated and observed values. Note that $KGE$ (and therefore $KGE_{ss}$) is maximized when $\alpha^{KGE}$, $\beta^{KGE}$ and $\rho^{KGE}$ are all $1.0$. As pointed out by *Knoben et al., (2019)* the adjustment from $KGE$ to $KGE_{ss}$ expressed by Eqn (6) ensures that $KGE_{ss} = 0$ (as opposed to $KGE = -0.414$) when the data-period mean of the observed streamflow series is used as the benchmark for comparison (as in the commonly used $NSE$ metric).

[31] To assess overall model performance under different hydro-climatological conditions (e.g., dry, medium, and wet years), we plot the 40-year distribution of year-wise *annual* $KGE_{ss}$ scores, and display these using Box-and-Whisker Plots. We also report and examine various percentiles (including *min*, 5%, 25%, *median*, 75%, and 95% values) for annual $KGE_{ss}$ and the three associated $KGE$ components. This approach enables distinguishing between different interannual (hydro-climatic) distributions of performance when different models share the same (or similar) average long-term levels of performance skill. In particular, it allows us to examine how well the different model architectures perform on the drier years that are characterized by low precipitation.

[32] In addition, we examine model performance on five different flow-magnitude groups, again using $KGE_{ss}$ and the three associated $KGE$ components. This enables an examination of how different architectural components (nodes and links) affect streamflow levels.

### 4.3 Hyperparameters and Training Procedure

[33] The training procedures used in this study are adopted from our previous work (*Wang & Gupta, 2024).* To initialize the model cell-states, we use a "*three-year*" spin-up that sequentially repeats the first water year data (WY 1949) three times at the start of the overall 40-year simulation period. This helps to minimize the potential effects of state initialization errors (*De la Fuente et al., 2023b*). The gradient-based ADAM optimization algorithm (*Kingma & Ba, 2014*) is used for model training (i.e., to determine optimal values for the parameters of the gating functions). The training metric and its gradient were computed using the streamflow values/timesteps assigned to the training subset. For each model architecture, each of the 10 random initializations of the parameters/weights was trained (or fine-tuned) for 2000 epochs using a single full-batch of the data, with a learning rate of $2.5 \times 10^{-1}$ for the first 300 epochs, reduced thereafter to $1.25 \times 10^{-1}$. From the 10 randomly initialized training runs, the one achieving the "*best*" $KGE_{ss}$ performance on the *selection* data subset was retained.

[34] To obtain relatively consistent results for progressively more complex system architectures, we adopted the following training procedure. We first trained the simplest $MA_1$ architecture. This then served as a benchmark lower bound on performance that we would expect to improve on (or at least not be worse than) when using more complex architectural hypotheses. As demonstrated in *Wang & Gupta (2024),* this simple single-node architecture already provides remarkably good performance on the *Leaf River* catchment. We then trained the $MA_2$ architecture as a progressive improvement to $MA_1$, where only the parameters associated with the recharge-output gate were randomly initialized at the start of training, while the parameters of the previously trained components were initialized to their previously obtained optimal values and then allowed to be further adjusted by the training process. Similarly, architecture $MA_3$ was trained as a progressive improvement to $MA_1$, and architecture $MA_4$ was trained as a progressive improvement to $MA_2$.

[35]  The procedure was altered slightly for architecture $MA_5$ as follows. In this case, the initial parameters for the soil moisture node were inherited from $MA_2$, the initial parameters for the channel routing node were inherited from $MA_3$, and the initial parameters for the groundwater node were inherited from $MA_4$; all parameters were then further adjusted by the training process. Finally, the initial parameters for $MA_6$ were inherited from $MA_5$, and only the gating parameters of the direct quick-flow path to the catchment outlet were initialized randomly.

[36]  Because groundwater dynamics can involve very long residence times, for all architectures with a groundwater node the corresponding cell-state was initialized to the values obtained using a time-constant output gate for the groundwater node, while the cell-states of the other nodes were initialized to zero. This facilitates obtaining a reasonably good guess for the initial value of the groundwater node cell-state by dividing the observed streamflow value for that initial time step by the time-constant value of the groundwater node output gate.

[37]  Finally, to facilitate proper parameter training, it was also found to be important to scale the "*informational quantities*" used as context variables by the various gates. For the surface soil moisture node, we used the scaling factors learned for the $MA_1$ architecture (means and standard deviations of the simulated 40-year time series of the cell-state) in our previous study (*Wang & Gupta, 2024*). For the channel routing and groundwater nodes, the corresponding cell-state scaling factors were obtained from a preliminary training step using time-constant output gating. Preliminary testing showed little sensitivity of the overall results to fairly coarse variations of these scaling factors and we therefore did not attempt to fine-tune them further.

[38]  Overall, adoption of this training strategy significantly reduced the amount of computation required for model development and testing. It also helped to stabilize the results by requiring fewer random initializations and helped to ensure that the progressively more complex model architectures could achieve performance results that were typically not worse than those of the models they were derived from. While we could have instead trained each new model architecture from scratch, progressively increasing architectural complexity would require exponentially increasing numbers of random initializations to obtain good training results, and so we adopted this progressive training strategy for all the experiments reported in this paper and summarized the parameter inheritance for all the cases in **Table S1**.  While possibly not globally optimal, this progressive training strategy should help to retain the performance gains already achieved with the simpler architecture (*Lee et al., 2018; Jia et al., 2019*), and result in an improvement (or at least not a decline) in training period performance. Note that a decline in generalization performance would suggest that overfitting has occurred.

## 5. Results for the Six MCP-Based Architectural Hypotheses Tested

[39]  Performance results for six aforementioned architectural hypotheses are shown visually in **Figure 3** and numerically in **Table 2**. In this discussion, we treat the simplest (single-node) $MA_1$ architecture as our baseline, and assess performance relative to it. In particular, **Figure 3** shows quantile box plots (5%, 25%, median, 75% and 95%) of the distributions of annual $KGE_{ss}$ performance for each year in the total 40-year data set.  Note that the red (+) symbols indicate performance on the outlier (drier) years. For ease of comparison, horizontal lines across the entire plot indicate the median, lower and upper quartiles of performance for the baseline $MA_1$ architecture.

### *5.1 Comparison of Alternative Architectural Hypotheses*

### 5.1.1 Model Architecture $MA_2$ (Single Cell-State, Two Flow Pathways)

[40]  We begin with model $MA_2$. Recall that this model represents the hypothesis that a single cell-state is sufficient to represent system storage dynamics, but that including more than one direct parallel pathway from cell-state to streamflow can provide an improved representation of the dynamical behavior of the *Leaf River* catchment. Compared with $MA_1$, we see improvements to the distribution of annual $KGE_{SS}$, with the distribution becoming tighter and shifting upwards towards the optimal value of 1.0 ($KGE_{SS}^{5\%}$ improves 0.48 →

$0.62$, $KGE_{SS}^{25\%}$ improves $0.78 \rightarrow 0.80$, median $KGE_{SS}^{50\%}$ improves $0.84 \rightarrow 0.85$, and $KGE_{SS}^{75\%}$ improves $0.87 \rightarrow 0.88$), thereby clearly supporting the two-flow-path hypothesis. While $KGE_{SS}^{95\%}$ performance declines very slightly (from $0.92$ to $0.91$), performance improves in 24 of the 40 years, and this improvement occurs across both high percentile (wetter) and low percentile (drier) years. Very notably, performance on the worst year ($KGE_{SS}^{worst}$) improves significantly from $0.30 \rightarrow 0.59$. Overall, we see a noticeable improvement in the low percentile (drier) years, which suggests that further attention to the representation of low flow processes is likely to be beneficial.

### 5.1.2 Model Architecture $MA_3$ (Two Cell-States, Single Flow Pathway)

[41]  Model $MA_3$ represents a complementary hypothesis to $MA_2$ where, instead of incorporating two parallel flow paths, $MA_3$ retains a single flow path but adds a second (streamflow) sequential routing cell-state along that flow path. Compared with $MA_1$ and $MA_2$ $KGE_{SS}^{99\%}$ improves dramatically (to $0.96$), while $KGE_{SS}^{95\%}$ remains at $0.91$ and $KGE_{SS}^{75\%}$ improves slightly (to $0.88$) indicating improved performance on wetter years. On the other hand, we observe an overall increase in the width of the distribution, caused by significant declines in $KGE_{SS}^{median}$, $KGE_{SS}^{25\%}$ and $KGE_{SS}^{5\%}$, indicating worse performance on the drier years. Notably performance on the outlier years is significantly worse than for the single-cell two-flow-path architecture $MA_2$ and very similar to the single-cell one-flow-path architecture $MA_1$.

[42]  Overall, the inclusion of a flow routing node does not actually enhance model performance by improving the overall mass balance, timing, and shape of the streamflow hydrograph (**Table 3**). Comparison of the results for $MA_1$, $MA_2$ and $MA_3$ suggest that while incorporating a representation of routing can improve performance on wetter years, the input-state-output dynamics of the LRB during drier periods is better represented by including more than 1 flow path ($MA_2$).

### 5.1.3 Model Architecture $MA_4$ (Two Cell-States, Two Flow Pathways)

[43]  Model $MA_4$ extends $MA_2$ by maintaining the representation of two parallel flow paths, but explicitly distinguishes between a "*quicker*" direct pathway to streamflow and a "*slower*" flow pathway that "*recharges*" a groundwater storage (a second cell-state) from which long-term baseflow is generated. As mentioned above (Section 4.3), proper training of the $MA_4$ architecture requires special attention to the initialization of the cell-state representing groundwater storage, because of its potentially very long residence times.

[44]  Compared with $MA_2$, we see clear improvement to the distribution of annual performance, with higher $KGE_{ss}$ skill for 24 of the 40 years. $KGE_{ss}$ skill improves for all of the quantiles above $25\%$, with clear improvements to the median, $75\%$ and $95\%$ years, but declines somewhat for the drier $5\%$, and worst years. The variability ratio ($\alpha^{KGE}$) improves in 22 of the years (with 40-year $\alpha^{KGE}$ increasing from $0.97$ to $1.01$), the mass-balance ratio ($\beta^{KGE}$) improves in 16 of the years, and the correlation coefficient improves in 39 of the years (with 40-year $\gamma^{KGE}$ improving from $0.88$ to $0.93$) indicating better reproduction of flow timing and hydrograph shape. Overall, inclusion of the cell-state representing groundwater storage seems to improve the representation of streamflow.

[45]  Note, however, that the $KGE_{SS}$ metric (and in general other "$MSE$" type metrics) for model training and performance evaluation tends to emphasize performance on the larger flow values, over other flow ranges. To take a closer look at what is happening, we partitioned the 40-year discharge data into five flow magnitude ranges based on the log-scale flow-duration (sorted flow) curve (**Figure S1 & Table 3**). Consistent with the nature of $KGE$, we see that only the highest flow range achieves $KGE$ components that are similar to those obtained when computed over all of the data (the entire flow regime). The middle three flow ranges tend to have around $200-400\%$ larger $\alpha^{KGE}$ (flow variability ratio), while the corresponding value for the lowest flow range is about $60-150\%$ larger. Similar results are seen for $\gamma^{KGE}$ (timing and shape) where the highest flow range has $\gamma^{KGE} > 0.8$ whereas the values for the remaining flow ranges are mostly below $0.4$.

[46]  Nonetheless, the fact that representation of the low flows improves with the $MA_4$ architecture indicates that there is clear value to incorporating a representation of groundwater dynamics in the model architecture.

The overall improved performance of $MA_4$ over $MA_2$ supports the hypothesis that it is necessary to both include a subsurface flow path and also to keep track of the groundwater state when modeling the LRB.

### 5.1.4 Model Architecture $MA_5$ (Three Cell-States, Two Flow Pathways)

[47] Model $MA_5$ combines the features included in $MA_2$, $MA_3$ and $MA_4$. Similar to $MA_2$ and $MA_4$ it includes two parallel flow paths, but now each of the flow pathways passes through a different cell-state so that overland/channel routing and long-term baseflow generation can both be simulated. As such this model is conceptually similar (but different in gating structure) to the well-known *HyMod* architecture (*Boyle, 2000*) that is widely used as a parsimonious way to simulate lumped catchment-scale rainfall-runoff dynamics. Accordingly, the $MA_5$ architecture emulates the *"HyMod Like"* architecture while providing greater functional flexibility through the use of context-dependent gating functions. To initialize the parameters of $MA_5$, we used the optimal values previously obtained for architectures $MA_2$ (for node 1), $MA_3$ (for node 2), and $MA_4$ (for node 3).

[48] As can be seen, $MA_5$ achieves overall better $KGE_{ss}$ performance only when compared to the $MA_3$. Further, it has a better $\beta^{KGE}$ mass balance ratios of 0.67 and 0.79 respectively for the two smallest flow groups (**Table 3**). Meanwhile, the worst year $KGE_{SS}^{worst} = 0.58$ shows the same significant improvement over $MA_1$ as was obtained using $MA_4$. As with $MA_4$, the results support the hypothesis that it is necessary to include both surface and subsurface flow paths and also to keep track of the groundwater state when modeling the LRB. Meanwhile, the benefits of including a routing store are not apparent.

### 5.1.5 Model Architecture $MA_6$ (Three Cell-States, Three Flow Pathways)

[49] Finally, we examine the potential benefits of including three flow pathways. $MA_6$ can be viewed as a modification of $MA_5$, in which a direct "*very fast*" flow path links the surface soil-moisture component (node 1) to the output. Conceptually, $MA_6$ can be viewed as representing fast overland flow, medium rate overland/interflow that is mediated by a channel routing process, and recharge to a groundwater storage that generates long-term baseflow.

[50] Compared with $MA_2$, $MA_4$ and $MA_5$ which all have two-flow paths, the three flow-path $MA_6$ architecture shows further improvement in the overall distribution of annual performance (**Table 2**). However, it does not consistently outperform $MA_4$ and $MA_5$ across all five flow regimes (see **Table 3**), indicating that the gain in performance relative to increased architectural complexity may be marginal.

### *5.2 Assessment of Performance across Flow Ranges*

[51] So far, we have focused on aggregate performance metrics computed from the entire streamflow hydrograph. Here, we summarize the effects of varying architectural complexity on the full range of flows (**Table 3**). **Figure 4** shows "*normal-scale*" and "*log-scale*" streamflow hydrograph plots for the driest (WY 1952), median (WY 1953), and wettest (WY 1973) water-years based on peak flow, enabling us to clearly examine the peak flow and baseflow behaviors of various models.

[52] From the upper row of plots (driest year), we see that all of the model architectures are unable to reproduce the first flow peak, and also struggle to reproduce some of the other peaks (**Figure 4a**). While it is possible that some important runoff generation processes are still not suitably represented, it seems likely that these results indicate the possibility of significant errors (under catch and/or non-recording) in the precipitation data for drier years. Nonetheless, as model complexity is progressively increased, particularly with the inclusion of the channel routing cell-state ($MA_3$; yellow line) and groundwater storage cell-state ($MA_4$; purple line; $MA_5$; green line; $MA_6$; light blue line), we see much better representation of baseflow (**Figure 4b**) – compare with the simplest model architectures $MA_1$ (blue line) and $MA_2$ (red line).

[53] From the middle row of plots (median peak flow year) we see that all of the model architectures perform similarly well at simulating the flow peaks, but that the simulation of groundwater recessions is significantly improved when the groundwater recharge, storage, and release processes are represented (**Figure 4c** & **Figure 4d**). Similar results are seen for the bottom row of plots (wettest year), but now the simpler architectures $MA_1$

and $MA_2$ do not perform as well as the more complex architectures ($MA_3$ through $MA_6$) at simulating the largest flood peak, suggesting that the increased architectural complexity does provide practical benefit for peak-flow (flood) prediction under wetter hydroclimatic conditions (**Figure 4e**). Again, inclusion of groundwater processes ($MA_5$ and $MA_6$) improves the representation of baseflow (**Figure 4f**). Qualitatively, architecture $MA_5$ seems to achieve the best overall performance across both high (peak) and low (groundwater recession) flows.

[54] Strictly speaking, considering all of the architectures tested, we only obtained a decent degree of quantitative skill (less than ~10% error for the variability and mass balance ratios, and less than ~20% error for correlation) in the largest flow group corresponding to $0.64\ mmday^{-1} < Q_{obs} \leq 64.01\ mmday^{-1}$ (see **Table 3**). For the three largest flow groups ($Q_{obs} \geq 0.32\ mmday^{-1}$), we obtain reasonably good values (less than ~10% error) for the mass balance ratio, which is encouraging given the mass-conserving nature of these model architectures. Overall, the use of $KGE$ as training metric seems to result in the models being able to closely match more than about 50% of the entire set of data points. In general, variability ratio and correlation performance tend to correlate well with flow magnitude, while including an explicit representation of both channel routing and groundwater processes tends to provide better mass balance ratios than when those processes are not represented.

[55] In summary, the *qualitative* (**Figure 4**) and *quantitative* (**Table 3**) analyses together indicate that the additional complexity associated with explicitly modeling channel routing and groundwater processes (resulting in a three-state representation), while increasing the number of flow-paths to two or more ($MA_5$ and $MA_6$) is clearly warranted. The result is a "*more adequate*" (realistic) representation of the behavior of the *Leaf River* catchment system (*Gupta et al., 2012*). However, if one was concerned only with flood prediction, then the more parsimonious $MA_3$ architecture (with no representation of groundwater) would likely provide sufficient predictive accuracy, and additional gains would likely require improvements in (primarily rainfall) data quality.

[56] Overall, the results do suggest that inclusion of a third (very fast) flow routing path ($MA_6$) can provide some additional degree of benefit for the LRB – it would therefore be worthwhile to examine the usefulness of this architecture for representing major flooding events by conducting tests on a variety of other catchments. On balance, however, the moderately complex "*HyMod Like*" $MA_5$ architecture seem to provide relatively good representations of the overall flow regime (especially including low flows), while contributing to a better understanding of how water flows through the system.

### 5.3 Gating Functions Learned for the HyMod Like Network Architectures

[57] In **Figure 5**, we plot the functional forms of the various gating functions learned for the six network architectures ($MA_1$ to $MA_6$). **Figure 5a** shows the shape of the remember gate of the first node (soil-moisture cell-state), which represents how the catchment partitions incoming precipitation into storage, loss to evapotranspiration, and release to streamflow, as a function of accumulated water ($X_t^{sm}$) and available energy (represented by potential evapotranspiration $PE_t$). We see that the forms of the remember gates are very similar between $MA_1$ and $MA_3$, between $MA_2$ and $MA_4$, and finally between $MA_5$ and $MA_6$.

[58] The $MA_1$ and $MA_3$ architectures tend to release very little of the accumulated water (the remember gate values tends to remain above 0.95 and 0.94 respectively) when the level of the cell-state is smaller than about 613 mm (660 mm) with the release gradually increasing to a value of about 5% (10%) when the cell-state exceeds that level. The role of energy demand ($PE_t$) to determine the release/store of water in the soil-moisture node is less important for $MA_3$ when the cell state is above ~800 mm, than for $MA_1$ for which the remember gate value drops gradually from 0.95 to 0.90 after $PE_t = 2.70\ mm$.

[59] $PE_t$ has a larger impact on the remember gate for the $MA_2$ and $MA_4$ architectures in which the groundwater flow path is added (compare with the single surface flow path architectures $MA_1$ and $MA_3$). For $MA_2$ the remember gate quickly decreases to below 0.90 at about $PE_t = 6.8\ mm$, and the remember gate value when $PE_t$ exceeds this threshold is relatively insensitive to the value of the cell state. In contrast,

whereas the remember gate for $MA_4$ has a similar shape to $MA_2$ when both the cell-state and $PE_t$ are small, at higher values the behavior still depends on both factors and there is a large difference in the minimum remember gate value (0.71 for $MA_2$ and 0.87 for $MA_4$). The conclusion is that addition of the groundwater recharge, storage and release processes has a significant impact on the storage and release dynamics of the soil-moisture cell-state.

[60] Finally, the soil-moisture cell-state remember gates for $MA_5$ and $MA_6$ are behaviorally very similar. Both retain smaller amounts of water (less than ~75%) when the cell state is larger than ~594 $mm$ (645 $mm$) regardless of the value of $PE_t$. When the cell-state is below this threshold, $PE_t$ values larger than 7.38 $mm$ (6.55 $mm$) result in more than 5% of the water being released. Overall, the addition of channel routing and groundwater storage cell-states and flow paths can lead to behaviorally very dissimilar behavioral dynamics for the soil-moisture node.

[61] Of course, we can expect interdependence between the learned representations of the remember gate and those of the other (output and loss) gates due to the fact that the gates do not function independently, and are not learned independently. We therefore further examine how the forms of the various gates change as additional components are progressively added to the model architecture.

[62] **Figure 5b** shows that the addition of an extra output flow path to the soil-moisture node (modifying $MA_1$ to $MA_2$) results in a decrease in the maximum value of the "*quick-flow*" output gate from 0.052 (darker blue line) to 0.038 (orange line). This is because $MA_2$ releases less water along the quick flow path to compensate for the fact that water is now also being released along the second "*slow-flow*" path (see orange line in **Figure 5d**) when the soil-moisture node cell state value is larger than ~702 $mm$. This is accompanied by a drop of the 40-year average cell-state value from 549 $mm$ to 269 $mm$.

[63] As noted earlier, the $MA_3$ model (obtained by adding a channel routing component to the single $MA_1$ output flow path) exhibits very similar behavior to the other models. Specifically, the model tends to "*lose*" about the same amount of water through the evapotranspirative process (see yellow line in **Figure 5c**) at lower values of the energy demand (this is also strongly apparent in the plot of accumulated remember gate flux (see **Figure S2d**). At the same time, the output gate for the single flow pathway begins to increase at about the same value of the cell-state (~640 $mm$) and increases up to a much larger value of 0.089. Also, the $MA_3$ architecture has almost the same 40-year average (550 $mm$) and very similar maximum cell-state value (811 $mm$) compared to $MA_1$ (814 $mm$). However, both cases have the same long-term accumulation of water mass in the soil-moisture tank (depicted by overlapping blue and yellow lines in **Figure S2a**). In fact, 640 $mm$ is about the 90th percentile cell-state value across 40-years for both cases indicating that the active part of the gate function is being utilized less than 10% of the time, hence producing very small amounts of mass in the long term.

[64] From **Figure 5f**, we see that the learned output gate for the $MA_3$ channel routing store has an almost constant value of around 0.81 over most of the cell-state range ($X_t^{rt} \geq 9\ mm$), such that it behaves much like a "*linear*" reservoir. However, further analysis reveals that the channel routing cell-state value is actually below this threshold for ~97% of the time steps (the 40-year average of the channel routing state is ~3.05 $mm$). So, while this architecture results in improved performance compared to $MA_1$ (as reported above; see **Figure 4**), the emphasis seems to have shifted to improved performance on the lower flow levels. This result (together with the results reported below) is suggestive of the importance of creating a separate groundwater flow path so that the channel routing cell-state can better perform its intended role of surface flow routing.

[65] In model $MA_4$ (which includes a cell-state for groundwater accumulation and release, but excludes the cell-state for channel-routing), the output gate for the soil-moisture cell-state saturates at the relatively small value of 0.03 (**Figure 5b**), while the recharge gate saturates at the relatively largest value of 0.183 (**Figure 5d**). Nonetheless, this does not result in more water actually being directed towards the groundwater storage (**Figure S2b**), and instead the accumulated mass flowing through the quick flow path is the largest among all of the cases in which a groundwater flow path is represented (**Figure S2a**). Compared with $MA_2$ (which has a

groundwater flow path but no groundwater cell state), the recharge gate saturation value increases from 0.088 to 0.183. However, the $MA_4$ model seems to prioritize quick flow over groundwater flow, with the output gate being activated (at 445mm) slightly earlier than recharge gate (at 460mm) reaching saturation (**Figure 5b**) value before the recharge gate (**Figure 5d**).

[66] Comparing $MA_5$ with $MA_4$ we see that addition of the channel-routing cell-state results in a large increase in the saturation value from 0.030 to 0.328 (**Figure 5a**), whereas the recharge gate value decreases from 0.183 to 0.060 (**Figure 5d**). In spite of this, $MA_5$ generates smaller amounts of lateral outflow from the soil-moisture tank (**Figure S2a**) and instead recharges more water downward to the groundwater cell-state (**Figure S2b**). This occurs because its recharge gate is activated earlier (at $X_t^{sm} = 395mm$) than for $MA_4$ ($X_t^{sm} = 460mm$) and compared to its own output gate ($X_t^{rt} = 510\ mm$).

[67] Meanwhile addition of the channel routing cell-state (transforming $MA_4$ into $MA_5$) does not result in the corresponding channel routing cell-state output gate behaving much different to that of $MA_3$ (**Figure 5f**). It does, however, result in the groundwater cell-state output gate of $MA_5$ behaving quite differently from that of $MA_4$ (**Figure 5g**). A closer look at long-term water accumulations shows that $MA_5$ tends to send more water to the groundwater node than $MA_4$ (**Figure S2b**) which results in $MA_5$ sending less water to channel routing than $MA_4$ (**Figure S2a**). Overall, $MA_5$ tends to store more water as groundwater which is arguably more consistent with our understanding of the behavior of the LRB.

[68] Finally, while the gating functions for $MA_5$ and $MA_6$ are quite similar for the soil-moisture cell-state, the output gate behaviors for the channel routing and groundwater cell-states are quite different. For $MA_6$, the channel routing output gate is activated earlier (at a smaller cell-state value). Addition of the extra quick flow path results in a decrease in the long-term output flux through the output gate (**Figure S2a**), recharge gate (**Figure S2b**), output gate in the routing node (**Figure S2g**) and the groundwater node (**Figure S2i**). Although this additional architectural complexity of $MA_6$ results in a refined representation of the surface flow process, with improved distribution of annual $KGE_{ss}$ compared to $MA_5$, it performs worse performance than $MA_5$ in the lower flow regimes for all of the years - dry, median and wet (**Figure 4b & Figure 4d & Figure 4e**).

# 6. Addition of an Input-Bypass Gate

[69] The results presented in **Section 5** suggest that it is more important to include multiple flow paths in the model architecture than to increase the number of nodes (cell-states). In traditional CRR model development, one important mechanism that can result in a time-varying number of flow paths is when the "*surface zone*" soil-moisture storage is represented as having a finite capacity, so that complete saturation of this storage causes a portion of the incoming precipitation to bypass the storage as "*saturation-excess*" overland flow. Alternatively, a similar effect can be achieved if the infiltration rate of the soil at the land surface is exceeded, leading to "*infiltration-excess*" overland flow.

[70] While these two processes are causally different (the former is due to the storage filling up and becoming saturated, while the latter is due to precipitation intensity exceeding some threshold rate), their behavioral effects can be difficult to distinguish in that they both result in the emergence of an overland flow path during longer and more intense precipitation events. And, of course, both processes can occur during any given precipitation event.

## 6.1 Structure of the Input-Bypass Gate

[71] To investigate the potential added benefit of enabling the model to simulate such processes, we augmented the model architecture to include a time-varying "*input-bypass gate*" $G_t^{U_{sm}}$ on the "*surface zone*" soil-moisture storage node. Specifically, we tested two different forms for this learnable input-bypass gate.

[72] The first formulation (referred to using the notation $BP_1$), is based on learning a threshold parameter value $\Theta_c$ (constrained to be positive) that represents the maximum capacity of the soil-moisture cell-state. This results in a "*saturation-excess*" bypass flow rate $U_t^P = G_t^{U_{sm}} \cdot U_t$ being generated by the soil-moisture tank

whenever it becomes fully saturated (i.e., $U_t + X_t^{sm} > \Theta_c$), where $U_t$ is the input precipitation and $X_t^{sm}$ is the unscaled cell state. In this case, the gating function $G_t^{U_{sm}}$ takes the mathematical form $G_t^{U_{sm}} = \frac{U_t^P}{U_t}$, $U_t^P = ReLU(U_t + X_t^{sm} - \Theta_c)$. In practice, we instead learn the parameter $\theta_c = \frac{\Theta_c}{w_s}$, where $w_s$ is a scaling constant (treated as a hyperparameter) that we set to $500$ (based, empirically, on trying various integer values from 100 to 500). To ensure positivity, and to avoid exploding gradients, we actually train on $\exp(\theta_c)$ while initializing $\theta_c$ randomly on $[-1, 1]$. Note that $0 \leq G_t^{U_{sm}} < 1$, so that $G_t^{U_{sm}} = 0$ corresponds to no saturation-excess bypass being generated by the system ($U_t^P = 0$).

[73] The second formulation (referred to using the notation $\boldsymbol{BP_2}$), is based on learning an input-bypass gating value $G_t^{U_{sm}}$ that can depend directly on *both* the cell-state and the precipitation intensity (therefore representing some combination of both "*saturation-excess*" and "*infiltration-excess*" mechanisms) but without the need to explicitly learn a value for the maximum soil-moisture cell-state capacity. In this case we parameterize the input-bypass gate simply as $G_t^{U_{sm}} = \sigma(b_t^{sm} + a_t^{sm} \cdot (\widetilde{X}_t^{sm} + \widetilde{U}_t))$, where $a_t^{sm}$ and $b_t^{sm}$ are trainable parameters, $\widetilde{X}_t^{sm}$ is the standardized value of the cell state and $\widetilde{U}_t$ is precipitation normalized by the 40-year maximum value.

### *6.2 Results of including an Input-Bypass*

[74] Overall, we found that the second ($\boldsymbol{BP_2}$) formulation provided better results. This is interesting, because the first ($\boldsymbol{BP_1}$) formulation more closely matches the mechanism typically built into conceptual rainfall-runoff models. However, the second formulation is arguably more general, in that it can learn to represent the combined effects of both the "*saturation-excess*" and "*infiltration-excess*" runoff generation processes, as and when they occur in a catchment. We therefore only present here the results obtained using the $\boldsymbol{BP_2}$ formulation; for the $\boldsymbol{BP_1}$ results, please see **Figure S3** in the supplementary materials where we include the results from $\boldsymbol{MA_1}$ to $\boldsymbol{MA_6}$ as examples.

[75] The results (**Figure 6**) are shown in the form of scatterplots of model performance where the x-axes represent the un-augmented models, and the y-axes represent the input-bypass-augmented models. The individual dots represent performance on each of the 40 available water-years. The four columns show scatterplots for the $KGE_{ss}$ (overall skill; 1 is optimal), $A^{KGE} = 1 - |1 - \alpha^{KGE}|$ (based on the variability ratio; 1 is optimal), ), $B^{KGE} = 1 - |1 - \beta^{KGE}|$ (based on the mass-balance ratio; 1 is optimal), and $\gamma^{KGE}$ (cross-correlation; 1 is optimal) metrics. Accordingly, values that fall above the $1{:}1$ line correspond to years in which adding the input-bypass mechanism improves model performance as measured by each metric. Each row corresponds to a different baseline model architecture ($\boldsymbol{MA_1}, \boldsymbol{MA_2}, \boldsymbol{MA_3}, \boldsymbol{MA_4}, \boldsymbol{MA_5}$ and $\boldsymbol{MA_6}$).

[76] Overall, it seems clear that there is significant benefit to incorporating the $\boldsymbol{BP_2}$ input-bypass mechanism on the "*surface zone*" soil-moisture cell-state, to create an additional (context-dependent) flow pathway. The improvement is particularly noticeable for architecture $\boldsymbol{MA_3}$, in which there was originally only one flow path (passing through two cell-states) – while the variability and mass-balance ratios are clearly improved, especially on the drier (lower-performance) years, there is significant improvement to hydrograph timing and shape ($\gamma^{KGE}$) across all of the water-years (wet and dry). Similar overall improvements are seen for the *HyMod Like* architecture $\boldsymbol{MA_5}$, and although there is deterioration to variability ratio in some of the years, we again see significant improvement to hydrograph timing and shape across (almost) all of the water-years. More detailed analysis is provided in the supplementary materials (**Text S1**).

## 7. Allowing for Unobserved Exchanges of Mass with the Environment through the Groundwater System

[77] *Wang & Gupta (2024)* showed that the performance of the single-node $\boldsymbol{MA_1}$ model architecture for the LRB can be enhanced, particularly during drier years, by allowing for the possibility of unobserved mass exchanges with the surrounding environment (*Schaller & Fan, 2009; Fan, 2019*). In other words, we allow for the possibility that the LRB exports water to the environment through the subsurface under very wet

conditions, while receiving (importing) water through the subsurface under very dry conditions due to differences in hydraulic head. Here, we implement the possibility of such behaviors by adding a learnable "*mass relaxation*" (MR) gate $G_t^{MR_{gw}}$ to the groundwater cell-state, so as to allow for the possibility of context-dependent bi-directional groundwater exchanges with the environment.

### 7.1 Structure of the Mass Relaxation Gate

[78] The flux through the MR gate is formulated as $q_t^{MR_{gw}} = G_t^{MR_{gw}} \cdot abs\left(X_t^{gw} - c_{MR_{gw}}\right)$ where $-1 < G_t^{MR_{gw}} < 1$ so that positive (negative) values represent fluxes out of (into) the groundwater cell-state. The parameter $c_{MR_{gw}}$ is in units of the cell state and represents the equilibrium cell state at which there is no mass exchange with the environment. The value of $c_{MR_{gw}}$ is normally set to be larger than $0$, but can (in principle) be allowed to be negative so that water always flows out of the node, or larger than the maximum value of the cell-state $max\left(X_t^{gw}\right)$, so that water always enters the node.

[79] To ensure physical consistency, the MR gate is actually implemented as $G_t^{MR_{gw}} = f_t^{MR_{gw}} - ReLU\left(f_t^{MR_{gw}} - G_t^{R_{gw}}\right)$ where $f_t^{MR_{gw}}$ is the "*learned*" mass relaxation (MR) gating function (see below), whose value is then adjusted to $G_t^{MR}$ to ensure that it is impossible to lose more water from the system than the total cell state available in the node. Here, based on the results reported in *Wang and Gupta (2024)*, we test only the cell-state-dependent version of the mass relaxation gate. Accordingly, we have $f_t^{MR_{gw}} = \kappa_{MR_{gw}} \cdot Tanh\left(a_{MR_{gw}} \cdot \left(\hat{X}_t^{gw} - \tilde{c}_{MR_{gw}}\right)\right)$ where $\kappa_{MR_{gw}}$, $a_{MR_{gw}}$ and $\tilde{c}_{MR_{gw}}$ are trainable parameters, and $\hat{X}_t^{gw}$ is the scaled cell-state time series. The value of $\kappa_{MR_{gw}}$ is constrained to lie on $[0,1]$, and $a_{MR}$ is constrained to be larger than $0$. Finally, the *remember* gate of the groundwater cell-state is represented as $G_t^{R_{gw}} = \left(1 - G_t^{O_{gw}} - G_t^{MR_{gw}}\right)$.

[80] The aforementioned MR gate was added to the groundwater cell-state of the $MA_4$, $MA_5$ and $MA_6$ architectures and we trained the models by either constraining the $c_{MR_{gw}}$ value to be positive, or by giving it the flexibility to possibly learn to be negative. Each version of the model was trained 10 times, by initializing with different random seeds. Overall, we found that allowing $c_{MR_{gw}}$ to be either negative or larger than the maximum 40-year groundwater tank cell-state was not helpful in improving model performance, and therefore we only report the cases when $0 \leq c_{MR_{gw}} \leq max\left(X_t^{gw}\right)$, so that water can be both added and removed from the groundwater node due to unobserved cell-state-dependent mass exchanges with the environment.

### 7.2 Results of including Mass Relaxation

[81] Overall, for all three architectures, inclusion of the MR gate significantly improves low-flow hydrograph performance, particularly when the flow is below $10^{-1}$ $mmday^{-1}$ (**Figure S4**). The improvement is most significant for the dry year but is also noticeable for the medium year (**Figure S4d** to **Figure S4f**) while having only a small impact on the wet year due to the generally high magnitude of the flows (**Figure S4g** to **Figure S4i**). Here, we choose to report detailed results only for the $MA_5$ architecture, since it provides the overall best performance in the previous section. The MR-gate augmented version of $MA_5$ is referred to as $MA_5MR_{gw}$.

[82] The top row of **Figure 7** shows (log-scale) hydrographs for the $MA_5$ architecture, so that we can see the dry-, medium-, and wet-year effects of including (dashed lines) and not-including (solid lines) the learned groundwater-cell-state MR gate. As might be expected, the benefit of MR-gating is most significant during the dry year (**Figure 7a**) where the mass relaxation gate consistently adds water into the groundwater node (**Figure 7g**), while the effects of precipitation on the groundwater cell-state are quite muted (**Figure 7d**) compared to the medium (**Figure 7e**) and wet (**Figure 7f**) years.

[83] In contrast, during the medium (**Figure 7h**) and wet (**Figure 7i**) years water either enters or leaves the groundwater node depending on the magnitude of its cell state relative to a threshold value of $c_{MR_{gw}} = 129.76\ mm$; water enters the system when the cell state is larger than this value and vice versa. In general, low flow performance improves for all three years, although the results show that further improvements in performance are still possible. A more nuanced analysis of the effects of MR-gating on the $MA_4$, $MA_5$ and $MA_6$ architectures is provided in the supplementary materials (**Text S2**).

## 8. Benchmarking Against Physical-Conceptual Models and LSTM Networks

### 8.1 Benchmark Models

[84] To benchmark the performance of our MCP-based approach on the *Leaf River* dataset, we compare against the pure data-based LSTM network and the three CRR model architectures (two of which are very well-established) listed below. All of these benchmark models use exactly the same input information (i.e., precipitation and PET) as our MCP-based architectures.

a) **$HyMOD\ Like$**: The seven-parameter spatially-lumped **$HyMOD\ Like$** model, reported in *Wang & Gupta (2024)*, can be considered to be something like a "*lower*" performance benchmark for daily-time-step catchment-scale CRR modeling. Although it shares the same three-cell-state directed-graph architectural representation as $MA_5$ (see section 5.1.4), consisting of a soil-moisture tank and surface routing tank for surface flow, and a groundwater tank for subsurface flow, the output gate conductivities are assumed to be *time-constant* values (not dependent on other contextual variables) as is common in CRR modeling. However, like other CRR models, the loss gate conductivity *does* vary with time to compute evapotranspiration losses, in a manner that depends on PET. The overall **$HyMOD\ Like$** architecture contains seven parameters that must be calibrated. We should expect most existing CRR type models to achieve better performance than this model.

b) **$GR4J$**: The four-parameter spatially-lumped **$GR4J$** CRR model architecture consists of two cell-states and two unit hydrographs (*Perrin et al., 2003*). The four parameters that must be calibrated include a 1) production store maximal capacity, 2) catchment water exchange coefficient, 3) day maximal capacity of the routing reservoir, and 4) unit hydrograph time base. The two cell-states represent a "*production*" store and a "*routing*" store. This parsimonious model architecture has been extensively tested and found to provide excellent performance across a wide range of catchments (*Anshuman et al., 2021; De la Fuente er al., 2023b*).

c) **$SACSMA$**: Then sixteen-parameter spatially-lumped **$SACSMA$** *Sacramento Soil Moisture Accounting* CRR model architecture (*Burnash & Ferral, 1973; Peck, 1976*) is one of the key components of the US *National Weather Service River Forecast System* (NWSRFS). Consistent with *Sorooshian et al. (1993),* the version adopted here consists of 6 cell-states (distributed between upper and lower soil zones), that route water in a complex manner along several flow paths (surface, subsurface and groundwater). This model architecture has been extensively used to model streamflow across the continental US (*Kratzert et al., 2019ab*).

d) **$LSTM$**: The purely data-based *Long Short-Term Memory Network* (**$LSTM$**; *Hochreiter & Schmidhuber, 1997*) is a special type of recurrent neural network that includes memory cells that can store information over long periods of time, and uses three gating operations (input, forget, output). Various forms of the **$LSTM$** network with increasing capacity (numbers of cell-states) were tested. Due to the lack of any physically-based restrictions being placed on its architectural form, the performance of the **$LSTM$** network can be taken to represent an approximate upper benchmark on precipitation-to-streamflow conversion performance achievable using the *Leaf River* dataset (*Nearing et al., 2020a*). Numbers of trainable parameters vary from 43 for the single-layer 2-cell-state version to 223 for the corresponding 6-cell-state version.

[85] We implemented the ***HyMOD Like*** and ***LSTM*** models using the differentiable computational environment discussed earlier, and trained them using the backpropagation algorithm, following the procedures presented in section 4.3. To calibrate the parameters of the ***GR4J*** and the ***SACSMA*** models, we used the *Shuffled Complex Evolution* (*SCE*; *Duan et al., 1992*) global optimization algorithm implemented in the open-source SPOTPY package (*Houska et al., 2015*). All models were trained using the same data-splitting procedure, KGE objective function, and 3-year spin-up period as was used for developing the MCP-based architectures reported earlier. **Text S3** in the supplementary materials provided more detailed information regarding the three benchmark CRR models.

### *8.2 Benchmarking Results*

[86] To conclude this analysis, we compare performance of the MCP-based model architectures against the benchmark models listed in section 8.1. Specifically, we focus on the following four MCP-based models:

a) $MA_1$: The seven-parameter single cell-state $MA_1$ architecture reported in section 2.1 is the simplest of the MCP-based architectural hypotheses explored in this study. To keep the architecture as minimal as possible, we do not include either the input-bypass gate or the mass relaxation gate. Accordingly, its' architecture contains one flow path and uses simple Sigmoid-function gating on the output and loss gates, with ET-loss constrained to not exceed PET.

b) $MA_5$: The eighteen-parameter two flow-path three cell-state $MA_5$ architecture reported in section 5.1.4 includes a representation of surface soil-moisture, surface routing, and groundwater storages. While architecturally similar to the ***Hymod Like*** CRR model, it uses time-variable Sigmoid-function gating, and ET-loss is constrained to not exceed PET.

c) $MA_5MR_{gw}^{\sigma}$: The twenty-one-parameter two flow-path three cell-state $MA_5MR_{gw}^{\sigma}$ architecture reported in section 7.1 is an extension of $MA_5$ to include a time-variable groundwater-cell-state-dependent mass relaxation gate that enables unobserved exchanges of groundwater with the surrounding environment.

d) $MA_5BP_2$: The twenty-parameter two flow-path three cell-state $MA_5BP_2$ architecture reported in section 6.1 is also an extension of $MA_5$, this time to include time-variable sigmoid-based input-bypass gating, which controls how much water enters the surface soil-moisture tank, and how much water bypasses it.

[87] **Figure 8** and **Table 4** summarize the 40-year annual $KGE_{ss}$ performance for all of the above-mentioned models using box whisker plots. Results for other metrics, including the three $KGE$ components, Nash–Sutcliffe Efficiency ($NSE$), Root Mean Square Error ($RMSE$), and Mean Absolute Error ($MAE$) on the annual basis are provided in **Table S3**.

[88] The leftmost subplot in **Figure 8** shows performance results for the benchmark CRR models. Here, we see significant progressive improvement from the ***HyMOD Like*** (three cell-states, seven parameters) to ***GR4J*** (two cell-states, four parameters) to ***SACSMA*** (six cell-states, sixteen parameters) model architectures, with $KGE_{ss}^{median}$ increasing from $0.53 \rightarrow 0.77 \rightarrow 0.81$, and $KGE_{ss}^{worst}$ improving from $0.11 \rightarrow 0.26 \rightarrow 0.48$. These results highlight the facts that:

- The well-established and parsimonious ***GR4J*** model performs significantly better than the ***HyMOD Like*** model indicating an architectural design that is much more efficient in its ability to exploit the information provided by the precipitation and potential evapotranspiration data. One reason for this is that the ***GR4J*** model enables unobserved groundwater mass exchange with the environment (the ***HyMOD Like*** model does not) which, as discussed in section 7, can significantly improve performance, especially during drier years.

- Conversely, performance of the ***SACSMA*** model (which also does not model unobserved groundwater mass exchange with the environment) is only slightly better than that of the ***GR4J*** model in terms of the $KGE_{SS}^{median}$ (0.77 → 0.81), although quite a bit better in terms of $KGE_{SS}^{worst}$ (0.26 → 0.48).

[89] The center subplot in **Figure 8** shows performance results for the four MCP-based model architectures listed above ($MA_1$, $MA_5$, $MA_5MR_{gw}^{\sigma}$ and $MA_5BP_2$). As noted in sections 5 and 6, we see progressive improvement with better dry- and wet-year performance as the model architectural hypothesis is enhanced. Note however that:

- The simplest single-cell-state MCP-based $MA_1$ architecture with time-variable gating achieves significantly better performance than the three-cell-state $Hymod\ Like$ architecture with time-constant output gating. Further it also performs distributionally better than the traditional two cell-state $GR4J$ model architecture across all (dry, medium, and wet) years, and is better than the six cell-state $SACSMA$ model on all but the two driest years.

- The three cell-state MCP-based $MA_5$, $MA_5MR_{gw}^{\sigma}$ and $MA_5BP_2$ architectures (all of which include representation of groundwater) with context-dependent time-variable gating achieve significantly better distributional performance than the best benchmark CRR model (the six cell-state $SACSMA$).

[90] Finally, the rightmost subplot in **Figure 8** shows performance results for four purely data-based single-layer $LSTM$ model architectures ($LSTM(N_H)$, for $N_H = 2, 3, 5$ and 6), where $N_H$ indicates the number of cell states/nodes. The distributionally best performance for the LRB was obtained with $N_H = 6$ (223 parameters), and testing of $N_H$ up to 10 did not result in further performance improvements (see supplementary materials). From these results we note that:

- The 223-parameter six-cell-state $LSTM(6)$ model performs only marginally better than the 20-parameter three-cell-state MCP-based $MA_5BP_2$ architecture across the entire distribution of years, and when using the same number of cell-states (three) as the $MA_5BP_2$ architecture, its median performance is lower. The fact that best $LSTM$ performance is achieved with six informational cell-states (and not higher) might suggest that as many as six cell-states are needed to optimally represent the RR dynamics of the LRB. However, achieving that (marginally better) level of performance while still maintaining an interpretable architecture could be challenging, and is left for future exploration.

- The single layer $LSTM$ shows the highest overall skill when $N_H = 5 - 6$, and it performs generally better than the $MA_5BP_2$. This can imply that we may need as many as 5 or 6 state variables to properly represent the RR dynamics of the LRB. Note that, as demonstrated in *Wang & Gupta (2024)*, the single node $LSTM$ has worse performance and poor physical interpretability when compared with the generic single node $MCP$ (i.e., $MA_1$) architecture. The $LSTM$ results suggest that it may be beneficial to explore different PC architectures with more state variables (e.g., the 6-state variable architecture of the $SACSMA$) using MCP-like nodes as the basic computational units.

[91] Overall, these results show that MCP-based mass-conserving architectures with learnable context-dependent time-variable gating can achieve superior performance to traditional physical-conceptual CRR model architectures, while maintaining similar levels of interpretability. Further, their performance can approach that of purely data-based machine-learning architectures that are not constrained to obey physical principles other than cell-state recurrence.

# 9. Conclusions

## 9.1 Summary & Discussion

[92] This study used the physically-interpretable MCP-based computational unit proposed by *Wang & Gupta (2024)* to explore a variety of model architectural hypotheses for the rainfall-runoff dynamics of the LRB in Mississippi. The various model structures were developed in a progressive manner, thereby facilitating hypothesis testing regarding a parsimonious representation of the underlying nature of the dominant

processes giving rise to the observed rainfall-runoff dynamics. In contrast with the breadth-focus, where a predetermined model architecture is tested on a large sample of catchments, we adopted a depth-focus wherein the performance of different architectures with varying numbers of state variables (nodes) and flow paths (links) was comprehensively examined in terms of a number of diagnostic metrics, including the variability ratio, mass balance ratio, and timing of the hydrograph over different years (dry, medium and wet) and flow magnitude ranges. This approach better facilitates analysis of both peak- and low-flow performance, revealing the relative value of progressively incorporating different nodal components (cell-states) and links (flow-paths), and experimental support for other hypothesis such as an input-bypass mechanism to simulate saturation/infiltration excess processes at the land surface, and a mechanism for bi-directional groundwater mass exchange with the surrounding environment.

[93] Compared to the single node $MA_1$ architecture explored in detail by *Wang & Gupta (2024),* we found noticeable improvement in the low percentile (drier) years when a groundwater flow path is added ($MA_2$). However, when only a routing tank was added ($MA_3$) we observed that improvement in the high percentile (wetter) years was achieved at the expense of significant performance declines in the low percentile (drier) years. By adding a groundwater tank to independently track the state of the groundwater system ($MA_4$) the overall performance was further improved, clearly highlighting the value of including a subsurface flow path for modeling the RR dynamics of the LRB. On balance, the three cell-state and two flow path *"HyMod Like"* $MA_5$ architecture that represents separate flow paths for overland/channel routing and groundwater dynamics achieved the best overall performance, maintaining the overall annual $KGE_{ss}$ levels achieved by $MA_4$ but with improvements to skill in the representation of low flow magnitudes and dynamics. Finally, the addition of a *"very fast"* overland flow path ($MA_6$) resulted in a slightly better distribution of annual $KGE_{ss}$ but without consistently outperforming $MA_4$ and $MA_5$ across all five flow regimes. Overall, our results clearly emphasize the importance of using multiple diagnostic metrics for performance evaluation (*Gupta et al., 1998*) when comparing alternative architectural hypotheses for which long-term median $KGE_{ss}$ performance can be very similar.

[94] Further inclusion of an input-bypass mechanism to the "*surface zone*" soil-moisture cell-state was found to be particularly useful for improving performance of the $MA_3$ and $MA_5$ architectures, mainly by improving the timing of the peak flows. This mechanism adds an additional (context-dependent) flow pathway by representing the combined effects of "*saturation-excess*" and "*infiltration-excess*" runoff generation processes. Similarly, inclusion of a cell-state-dependent mass-relaxation gating mechanism to the $MA_5$ architecture, to enable bidirectional (gaining and losing) groundwater exchanges between the groundwater cell-state and the surrounding environment, significantly improved low flow ( $Q < 10^{-1}\ mmday^{-1}$ ) performance, especially in drier periods when water was gained from the environment while allowing water losses to the environment during wet periods.

[95] Note that the results reported in this study were obtained by training the models using the $KGE$ metric, which (similarly to $MSE$) is known to emphasize high-flow (peak-flow) performance while giving relatively little importance to matching low-flow dynamics. In spite of this metric-driven bias to the model development process, we found that by adding a separate architectural representation for groundwater dynamics (groundwater cell-state and flow path) significant overall improvements to the representation of RR dynamics was achieved. However, our experience with this study suggests that it may be valuable to explore the development and use of alternative metrics (*Gupta et al.,1999*) that will result in better information extraction and training of hydrologic models (and for other systems such as this where the distribution of the system output is highly skewed and can vary over several orders of magnitude) by achieving a more balanced emphasis on reproducing the entire range of (high, medium and low) flows. By doing so, it might be possible to reduce the need to examine so many different diagnostic metrics, and better facilitate the breadth-focus development of models using large samples of catchments representing a wide range of hydroclimatic regimes.

[96] Following *Wang & Gupta (2024)* the training procedure used in this study involved stagewise model development wherein the initialization values for weights/parameters were inherited from those estimated at

previous stages. While an important advantage of this approach is to reduce computational cost, there is a potential danger that by prioritizing the training of different parts of the model it can lead to multiple-plausible inferred results for the forms of the gating functions. Future work should place more emphasis on ensuring the robustness of model interpretability, particularly when applying MCP-based architectures to large sample studies that span diverse hydro-geo-climate conditions.

[97]  All of the models reported in this study were trained using the well-established $KGE$ metric (Gupta et al., 2009). While other metrics including $NSE$, RMSE, MAE, and the 3 components of $KGE$ are summarized in Tables S2 and S3 (for comparison against other studies), it is important to recognize the inherent multi-objective nature of the RR problem (*Gupta et al., 1998*) and future studies should take into consideration the potential benefits of multiple training metrics (*Gupta et al., 1999*). We suggest, therefore, that the information-theoretic paradigm be used as a basis for the selection of the model training metric, so as to minimize information loss while enhancing model training efficiency and generalizability (*Hodson et al., 2024*). For example, criteria such as AIC, BIC (*Akaike, 1974; Schwarz, 1978*), and minimum description length (*Weijs & Ruddell, 2020*) may be useful for objectively addressing the tradeoff between predictive accuracy and model complexity.

[98]  Finally, in response to a review comment, we should note that the results generated using MCP-based models are "*interpretable*" in the exact same sense that results generated using traditional PC-based CRR models are interpretable. Such models simulate time series of quantities that are conceptually referred to as "*surface soil moisture*", "*actual evapotranspiration*", and "*groundwater storage*" etc., and that individually and collectively obey certain regularizing principles such as "*mass-conservation*", "*fluxes in response to thermodynamic gradients*", "*flux-rate conductivities that are less than 1*", and so on. However, we must recognize that all quantities represented by such models are only <u>*analogically*</u> related to physical quantities assumed to exist in the system being represented and must therefore be interpreted as such. This, unavoidably, holds true for mathematical- and computer-based models of all types, including those that are ML-based, PC-based, or even (so-called) "*physics-based*". As such, the cell-states of a model must be understood to be only *analogically* representative of real physical quantities. Without access to actual physical measurements of the target quantities at the appropriate (effective) scale of the model, the best we can do is try to establish that the conceptual model quantities are "*behaviorally informative*" about the real-world quantities they are intended to represent. In lumped catchment-scale CRR modeling, such data (e.g., about upper or lower zone soil moisture) are virtually impossible to obtain, and so we must rely on physical intuition to establish whether the state variables behave in an "*informationally reasonable*" manner. For example, *Kratzert et al., (2018)* demonstrated that certain cell-states in their LSTM-based model behave in a manner that is informative about snowmelt dynamics.

### 9.2 Future Extensions

### 9.2.1 Towards System Architecture Search for Hydrologic Models at Catchment-Scale

[99]  By viewing a catchment system as consisting of conceptually-interpretable subcomponents that can be represented via relatively low-order (minimum description length) directed graphs consisting of nodes and links, an important future direction will be to employ network architecture search procedures (*Elsken et al., 2019*) whereby evolutionary/genetic algorithms are combined with differentiable programming to searching the model space for "optimal" (plausible) physically-Informed network architectures, thereby achieving a more complete exploitation of the information available in data via structural assimilation (*Stanley and Miikkulainen, 2002*; *Liu et al., 2018*; *Lan et al., 2022*). In this regard, it may also be productive to explore approaches such as Gaussian Sliding Windows Regression for hydrologic inference of the number of flow paths inside the catchment (*Schrunner et al., 2023*).

### 9.2.2 Towards Regionalization for CONUS-Wide Modeling

[100]     From the breadth perspective, future large catchment-sample investigations should investigate the bare minimum model complexity required at different locations, by exploring a variety of different

parsimonious MCP-based network architectures for their suitability in different hydro-climatic regimes (*Gupta et al., 2014*). Recent work (e.g., *Kratzert et al., 2024*) has demonstrated that it is possible to develop regional, continental, and even global scale hydrological models by leveraging data-based deep learning approaches trained on large heterogeneous datasets. Nonetheless, it seems conceptually reasonable that hydrological understanding can be advanced by exploring the possibility of regional, continental, and even global scale model development by leveraging concepts commonly employed by traditional physical-conceptual modeling (*Samaniego et al., 2010*) and AI-physics hybrid approaches (*Jiang et al., 2020*). To support physical-conceptual understanding, such regional (etc.) approaches must necessarily allow for the representation of, and selection from, a variety of different interpretable dominant-process configurations (structural architectures). We expect the interpretability and functional flexibility afforded by the MCP-based time-variable gating approach to prove very useful as a backbone for the development of such models. In this regard, future work may also exploit concepts of data clustering and the use of similarity functions (*De la Fuente et al., 2023a*), and especially the power afforded by the *Yang & Chui (2023a,b)* strategy for latent-space encoding of large-sample catchment data sets.

[101]    Finally, one important limitation of the MCP-based architectures investigated in this current study is their lack of applicability to cold and high-elevation regions where snow accumulation and melt dynamics play an important role. While a relatively simple approach would be to couple the MCP with an existing snow module (e.g., *Patil & Stieglitz, 2014; Aghakouchak & Habib, 2010*), an approach that would be much more consistent with the philosophy underlying the work reported here would be to augment the current set of plausible MCP-based architectures to enable learning the forms of snow accumulation and melt gating functions driven by relevant meteorological contextual variables (*Jennings et al., 2018; Wang et al., 2019*). In this regard, it has already been established that accurate continental-scale representation of snow dynamics via deep learning is in principle possible (*Wang et al., 2022*). Doing so would set the stage for MCP-based models to be developed and applied to the interpretable physical-conceptual investigation of the potential hydrological impacts of climate change (*Hale et al., 2023*).

## Acknowledgments

The authors would like to thank the WRR editorial team, including *Marc Bierkens* (Editor), an anonymous Associate Editor, and reviewers Joseph *Janssen* and *Arik Tashie, plus two anonymous reviewers,* for providing constructive comments. The first author (*YHW*) would like to thank the late *Thomas Meixner*, as well as *Jennifer Mcintosh, Martha Whitaker, Eyad Atallah, Dale Ward, Ty Ferré, Jim Yeh,* and *Chris Castro*, for their support, and acknowledge the TA and outreach assistantship support provided by *the Department of Hydrology and Atmospheric Sciences* and the *University of Arizona Data Science Institute* during the final two years of his Ph.D. study, which made the finalization of this work possible. YHW extends the gratitude to *Derrick Zwickl* for his invaluable consultation on high-performance computing, and to *Johnathan Frame*, *Martin Gauch*, and *Grey Nearing* for the email discussion on the SAC-SMA model. We also thank *University of Arizona Data Science Institute* for providing HPC computation resources. The second author (HVG) acknowledges partial support by the *Australian Centre of Excellence for Climate System Science* (CE110001028), the inspiration and encouragement provided by members of the *Information Theory in the Geosciences* group (geoinfotheory.org), and support for a 4-month research visit to the *Karlsruhe Institute of Technology,* Germany provided by the *KIT International Excellence Fellowship Award* program.

## Open Research

**Table 1:** Detailed information for the multi-flow-path and multi-cell-state Mass-Conserving architectures examined in this study.

| Model Architectures | Nodes: States | Nodes: Gate Functions | Flow Paths | Output Flux Generated by Gates |
|---|---|---|---|---|
| $MA_1$ | Soil-Moisture Node: $x_t^{SM}$ | **Soil-Moisture Node:** Output, Loss, Remember Gates ($G_t^{O_{SM}}$, $G_t^{L_{SM}}$, $G_t^{R_{SM}}$) | Surface Flow Path | $L_t$: Loss gate flux in soil-moisture tank<br>$O_t^o$: Output gate flux in soil-moisture tank<br>$O_t$: System streamflow output |
| $MA_2$ | Soil-Moisture Node: $x_t^{SM}$ | **Soil-Moisture Node:** Output, Loss, Remember Gates ($G_t^{O_{SM}}$, $G_t^{L_{SM}}$, $G_t^{R_{SM}}$) | Surface Flow Path<br>Groundwater Flow Path | $L_t$: Loss gate flux in soil-moisture tank<br>$O_t^o$: Output gate flux in soil-moisture tank<br>$O_t^{RG}$: Recharge (Second Output) Gate flux in soil-moisture tank<br>$O_t$: System streamflow output |
| $MA_3$ | Soil-Moisture Node: $x_t^{SM}$<br>Surface Channel Routing Node: $x_t^{CH}$ | **Soil-Moisture Node:** Output, Loss, Remember Gates ($G_t^{O_{SM}}$, $G_t^{L_{SM}}$, $G_t^{R_{SM}}$)<br>**Surface Channel Routing Node:** Output, Remember Gates ($G_t^{O_{CH}}$, $G_t^{R_{CH}}$) | Surface Flow Path | $L_t$: Loss gate flux in soil-moisture tank<br>$O_t^o$: Output gate flux in soil-moisture tank<br>$O_t^{CH}$: Output gate flux in surface routing tank<br>$O_t$: System streamflow output |
| $MA_4$ | Soil-Moisture Node: $x_t^{SM}$<br>Groundwater Node: $x_t^{GW}$ | **Soil-Moisture Node:** Output, Loss, Remember Gates ($G_t^{O_{SM}}$, $G_t^{L_{SM}}$, $G_t^{R_{SM}}$)<br>**Groundwater Node:** Output, Remember Gates ($G_t^{O_{GW}}$, $G_t^{R_{GW}}$) | Surface Flow Path<br>Groundwater Flow Path | $L_t$: Loss gate flux in soil-moisture tank<br>$O_t^o$: Output gate flux in soil-moisture tank<br>$O_t^{RG}$: Recharge (Second Output) Gate f lux in soil-moisture tank<br>$O_t^{GW}$: Output gate flux in groundwater tank<br>$O_t$: System streamflow output |
| $MA_5$ | Soil-Moisture Node: $x_t^{SM}$<br>Surface Channel Routing Node: $x_t^{CH}$<br>Groundwater Node: $x_t^{GW}$ | **Soil-Moisture Node:** Output, Loss, Remember Gates ($G_t^{O_{SM}}$, $G_t^{L_{SM}}$, $G_t^{R_{SM}}$)<br>**Surface Channel Routing Node:** Output, Remember Gates ($G_t^{O_{CH}}$, $G_t^{R_{CH}}$)<br>**Groundwater Node:** Output, Remember Gates ($G_t^{O_{GW}}$, $G_t^{R_{GW}}$) | Surface Flow Path<br>Groundwater Flow Path | $L_t$: Loss gate flux in soil-moisture tank<br>$O_t^o$: Output gate flux in soil-moisture tank<br>$O_t^{CH}$: Output gate flux in surface routing tank<br>$O_t^{RG}$: Recharge (Second Output) Gate flux in soil-moisture tank<br>$O_t^{GW}$: Output gate flux in groundwater tank<br>$O_t$: System streamflow output |
| $MA_6$ | Soil-Moisture Node: $x_t^{SM}$<br>Surface Channel Routing Node: $x_t^{CH}$<br>Groundwater Node: $x_t^{GW}$ | **Soil-Moisture Node:** Output, Loss, Remember Gates ($G_t^{O_{SM}}$, $G_t^{L_{SM}}$, $G_t^{R_{SM}}$)<br>**Surface Channel Routing Node:** Output, Remember Gates ($G_t^{O_{CH}}$, $G_t^{R_{CH}}$)<br>**Groundwater Node:** Output, Remember Gates ($G_t^{O_{GW}}$, $G_t^{R_{GW}}$) | Surface Flow Path<br>Groundwater Flow Path<br>Surface Quick Flow Path | $L_t$: Loss gate flux in soil-moisture tank<br>$O_t^o$: Output gate flux in soil-moisture tank<br>$O_t^{FP}$: Quick-flow Path (third Output) Gate flux in soil-moisture tank<br>$O_t^{CH}$: Output gate flux in surface routing tank<br>$O_t^{RG}$: Recharge (Second Output) Gate flux in soil-moisture tank<br>$O_t^{GW}$: Output gate flux in groundwater tank<br>$O_t$: System streamflow output |

**Table 2:** $KGE_{ss}$ Statistics and numbers of parameter for the *"HyMod"* like mass-conserving architectural hypotheses investigated in this paper (best scores across architectures are shown in red, second-best scores are shown in blue)

| | $MA_1$ | $MA_1$ $BP_1$ | $MA_1$ $BP_2$ | $MA_2$ | $MA_2$ $BP_1$ | $MA_2$ $BP_2$ | $MA_3$ | $MA_3$ $BP_1$ | $MA_3$ $BP_2$ | $MA_4$ | $MA_4$ $BP_1$ | $MA_4$ $BP_2$ | $MA_5$ | $MA_5$ $BP_1$ | $MA_5$ $BP_2$ | $MA_6$ | $MA_6$ $BP_1$ | $MA_6$ $BP_2$ |
|---|---|---|---|---|---|---|---|---|---|---|---|---|---|---|---|---|---|---|
| $KGE_{ss}^{worst}$ | 0.30 | 0.28 | 0.49 | 0.59 | 0.56 | *0.68* | 0.30 | 0.57 | 0.55 | 0.58 | 0.58 | 0.59 | 0.58 | 0.51 | *0.63* | 0.57 | 0.53 | 0.60 |
| $KGE_{ss}^{5\%}$ | 0.48 | 0.47 | 0.56 | 0.62 | 0.59 | *0.70* | 0.47 | 0.62 | 0.60 | 0.60 | *0.68* | 0.63 | 0.59 | 0.56 | 0.64 | 0.63 | 0.64 | 0.67 |
| $KGE_{ss}^{25\%}$ | 0.78 | 0.77 | 0.79 | 0.80 | 0.80 | 0.81 | 0.75 | 0.76 | 0.81 | 0.80 | *0.82* | 0.81 | 0.78 | 0.79 | *0.82* | 0.81 | 0.78 | *0.83* |
| $KGE_{ss}^{50\%}$ | 0.84 | 0.84 | 0.84 | 0.85 | 0.85 | 0.85 | 0.82 | 0.84 | *0.88* | 0.86 | 0.84 | 0.86 | 0.84 | 0.83 | *0.89* | 0.86 | 0.84 | 0.87 |
| $KGE_{ss}^{75\%}$ | 0.87 | 0.87 | 0.87 | 0.88 | 0.87 | 0.88 | 0.88 | 0.87 | *0.91* | 0.90 | 0.87 | *0.91* | 0.88 | 0.86 | *0.92* | *0.91* | 0.88 | *0.91* |
| $KGE_{ss}^{95\%}$ | 0.91 | 0.92 | 0.91 | 0.91 | 0.91 | 0.91 | 0.91 | 0.91 | *0.95* | 0.94 | 0.91 | 0.94 | 0.92 | 0.91 | *0.96* | 0.94 | 0.92 | 0.94 |
| No. of Parameters | 7 | 8 | 9 | 10 | 11 | 12 | 11 | 12 | 13 | 14 | 15 | 16 | 18 | 19 | 20 | 21 | 22 | 23 |

**Table 3:** Statistics of KGE components and KGE$_{ss}$ (shown in red are $\alpha^{KGE}$ scores between 0.5 and 1.5, $\beta^{KGE}$ scores between 0.8 and 1.2, $\gamma^{KGE}$ above 0.8 inclusive, and KGE$_{ss}$ scores above 0.8 inclusive). Note that the Leaf River dataset contains 14,610 timesteps in total. The flow is separated into five groups according to magnitude, and each group has 2922 data points, with the *"First Group"* representing the **lowest** 20th percentile of streamflow values and the "Fifth Group" representing the **highest** 20th percentile of streamflow values.

| | MA$_1$ | MA$_1$BP$_1$ | MA$_1$BP$_2$ | MA$_2$ | MA$_2$BP$_1$ | MA$_2$BP$_2$ | MA$_3$ | MA$_3$BP$_1$ | MA$_3$BP$_2$ | MA$_4$ | MA$_4$BP$_1$ | MA$_4$BP$_2$ | MA$_5$ | MA$_5$BP$_1$ | MA$_5$BP$_2$ | MA$_6$ | MA$_6$BP$_1$ | MA$_6$BP$_2$ |
|---|---|---|---|---|---|---|---|---|---|---|---|---|---|---|---|---|---|---|
| | | | | | | | $\alpha^{KGE}$ | | | | | | | | | | | |
| First Group | 1.67 | 1.62 | 2.93 | 1.65 | *1.49* | *1.44* | 2.53 | 3.01 | 2.65 | 1.72 | 1.71 | *1.33* | 1.95 | 3.25 | 2.50 | 1.62 | 2.23 | 1.84 |
| Second Group | 4.20 | 4.12 | 4.09 | 4.36 | 4.07 | 4.26 | 5.07 | 2.45 | 4.37 | 3.96 | 4.02 | 3.73 | 3.60 | 5.07 | 3.54 | 3.86 | 4.48 | 3.09 |
| Third Group | 4.72 | 4.67 | 4.47 | 4.95 | 4.72 | 4.68 | 5.20 | 3.29 | 3.97 | 4.54 | 4.55 | 4.67 | 4.11 | 4.77 | 3.07 | 4.40 | 4.71 | 3.46 |
| Fourth Group | 3.24 | 3.24 | 3.13 | 3.25 | 3.19 | 3.08 | 3.38 | 3.70 | 2.96 | 3.11 | 3.19 | 3.11 | 3.05 | 5.30 | 2.54 | 2.89 | 3.00 | 2.72 |
| Fifth Group | *0.97* | *0.97* | *0.98* | *0.94* | *0.95* | *0.97* | *1.03* | *0.98* | *1.00* | *1.00* | *0.99* | *0.99* | *1.03* | *1.02* | *1.00* | *1.00* | *1.00* | *0.99* |
| | | | | | | | $\beta^{KGE}$ | | | | | | | | | | | |
| First Group | 0.15 | 0.14 | 0.30 | 0.15 | 0.13 | 0.13 | 0.33 | 0.69 | 0.42 | 0.21 | 0.30 | 0.15 | 0.67 | 0.80 | 0.56 | 0.31 | 0.56 | 0.54 |
| Second Group | 0.47 | 0.46 | 0.55 | 0.51 | 0.47 | 0.47 | 0.70 | *0.82* | 0.69 | 0.50 | 0.61 | 0.43 | 0.79 | *0.93* | *0.81* | 0.60 | 0.78 | 0.72 |
| Third Group | *1.02* | *1.00* | *1.02* | *1.13* | *1.06* | *0.99* | *1.20* | *0.77* | *0.95* | *1.01* | *1.04* | *1.00* | *1.09* | *1.11* | *0.98* | *1.09* | *1.20* | *1.00* |
| Fourth Group | 1.28 | 1.27 | 1.26 | 1.38 | 1.32 | 1.24 | 1.36 | *0.98* | *1.18* | 1.24 | 1.23 | 1.26 | 1.24 | *1.14* | *1.13* | 1.26 | 1.29 | *1.16* |
| Fifth Group | *0.97* | *0.97* | *0.98* | *0.97* | *0.95* | *1.00* | *0.94* | *1.03* | *1.00* | *0.99* | *0.99* | *0.99* | *0.96* | *0.96* | *1.00* | *0.97* | *0.95* | *0.98* |
| | | | | | | | $\gamma^{KGE}$ | | | | | | | | | | | |
| First Group | 0.25 | 0.25 | 0.16 | 0.27 | 0.27 | 0.26 | 0.29 | 0.45 | 0.31 | 0.28 | 0.39 | 0.27 | 0.39 | 0.32 | 0.40 | 0.35 | 0.34 | 0.40 |
| Second Group | 0.31 | 0.31 | 0.27 | 0.33 | 0.33 | 0.30 | 0.31 | 0.29 | 0.29 | 0.32 | 0.30 | 0.31 | 0.32 | 0.21 | 0.36 | 0.34 | 0.32 | 0.38 |
| Third Group | 0.39 | 0.39 | 0.39 | 0.41 | 0.41 | 0.38 | 0.37 | 0.27 | 0.35 | 0.39 | 0.37 | 0.40 | 0.40 | 0.30 | 0.40 | 0.41 | 0.39 | 0.42 |
| Fourth Group | 0.35 | 0.35 | 0.35 | 0.35 | 0.35 | 0.36 | 0.31 | 0.31 | 0.40 | 0.34 | 0.33 | 0.34 | 0.33 | 0.26 | 0.44 | 0.34 | 0.31 | 0.38 |
| Fifth Group | *0.82* | *0.82* | *0.82* | *0.82* | *0.82* | *0.82* | *0.81* | *0.80* | *0.89* | *0.88* | *0.81* | *0.88* | *0.82* | *0.81* | *0.90* | *0.88* | *0.82* | *0.89* |

**Table 4:** $KGE_{ss}$ Statistics and numbers of parameter for the *benchmark models* (best scores across architectures are shown in red, second-best scores are shown in blue)

| | *HyMOD Like* | *GR4J* | *SACSMA* | *$MA_1$* | *$MA_5$* | *$MA_5BP_2$* | *$MA_5MR_{gw}^{\sigma}$* | *LSTM(2)* | *LSTM(3)* | *LSTM(5)* | *LSTM(6)* |
|---|---|---|---|---|---|---|---|---|---|---|---|
| $KGE_{ss}^{worst}$ | 0.11 | 0.26 | 0.48 | 0.30 | 0.58 | 0.63 | 0.61 | *0.68* | *0.66* | *0.66* | *0.68* |
| $KGE_{ss}^{5\%}$ | 0.25 | 0.40 | 0.57 | 0.48 | 0.59 | 0.64 | 0.64 | 0.72 | *0.75* | *0.78* | *0.78* |
| $KGE_{ss}^{25\%}$ | 0.45 | 0.69 | 0.76 | 0.78 | 0.78 | 0.82 | 0.78 | 0.83 | *0.84* | *0.85* | *0.84* |
| $KGE_{ss}^{50\%}$ | 0.53 | 0.77 | 0.81 | 0.84 | 0.84 | *0.89* | 0.84 | 0.87 | 0.88 | *0.90* | *0.90* |
| $KGE_{ss}^{75\%}$ | 0.60 | 0.81 | 0.84 | 0.87 | 0.88 | *0.92* | 0.88 | *0.92* | *0.93* | *0.93* | *0.93* |
| $KGE_{ss}^{95\%}$ | 0.68 | 0.88 | 0.89 | 0.91 | 0.92 | *0.96* | 0.93 | 0.95 | *0.96* | *0.98* | *0.96* |
| No. of Parameters | 7 | 4 | 16 | 7 | 18 | 20 | 21 | 43 | 76 | 166 | 223 |

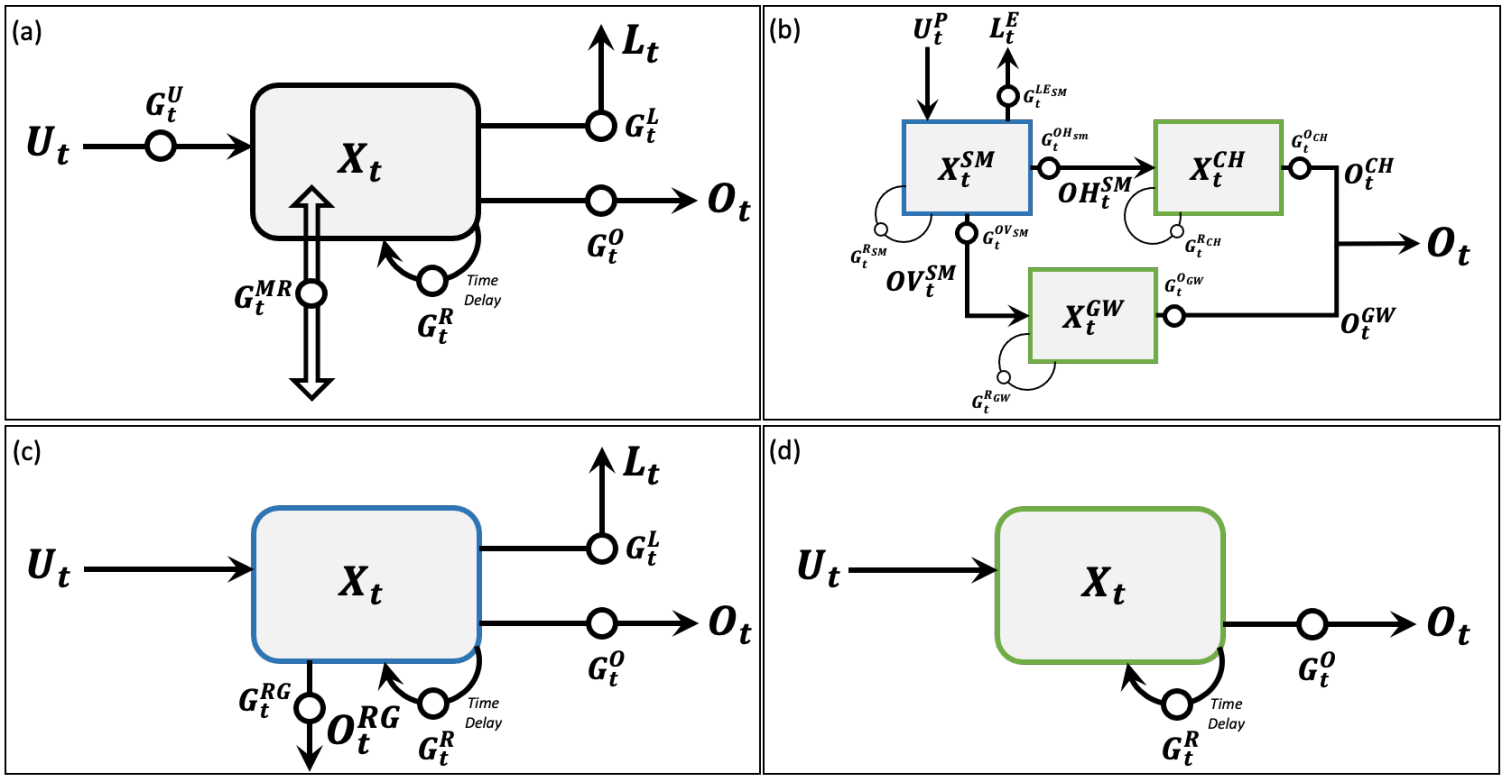

**Figure 1:** Direct-graph representation for a) single node generic mass-conserving perceptron (MCP), b) *HyMod* Like hydrologic model, and the subcomponents of a *HyMod* Like architecture including c) surface soil moisture tank/node, and d) surface routing and groundwater tank/node. * Note that the number of output gates needed in the main soil moisture tank depends on the number of flow paths in the architecture. For instance, the architecture requires two output gates in the main soil moisture tank when the flow paths include surface and groundwater flow (i.e., the HyMod-Like Architecture). In this illustrative case, compared to subplot (b), subplot (c) is used as the soil moisture tank with $G_t^O$ representing the output gate in the surface (horizontal) flow path ($G_t^{OH_{sm}}$), $O_t$ representing the streamflow generated through this (horizontal) surface flow path ($OH_t^{SM}$). Similarly, the soil moisture tank with $G_t^{RG}$ *represents* the output (recharge) gate in the subsurface (vertical) flow path ($G_t^{OV_{sm}}$), $O_t^{RG}$ representing the streamflow generated through this (vertical) subsurface flow path ($OV_t^{SM}$). Subplot (d) is used for both surface routing nodes and groundwater nodes, and the difference compared against subplot (a) lies in the lack of a loss gate in this architecture. The input $U_t$ equals to $OH_t^{SM}/OV_t^{SM}$, output gate $G_t^O$ represents $G_t^{O_{CH}}/G_t^{O_{GW}}$, and the output flow $O_t$ refers to $O_t^{CH}$ (surface channel routing outflow)/$O_t^{GW}$ (groundwater outflow) when subplot (d) architecture is used as surface (channel routing)/subsurface (groundwater) node. Similarly, we can easily map other state variables and gating functions between the *HyMod* Like architecture (subplot b) and the associated subcomponents (subplot c & d).

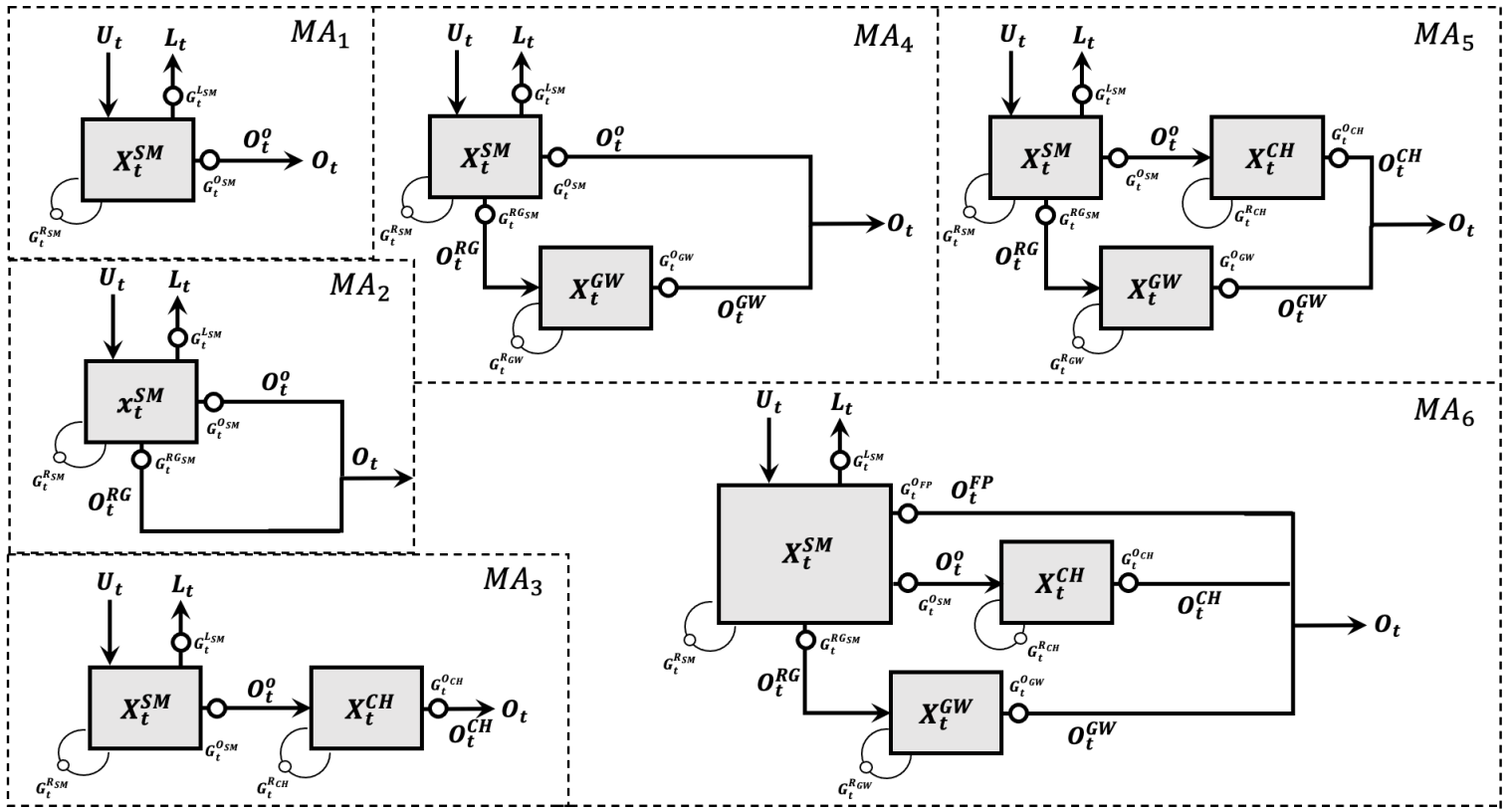

**Figure 2:** The multi-flow-path and multi-cell-state Mass-Conserving architectures examined in this study.

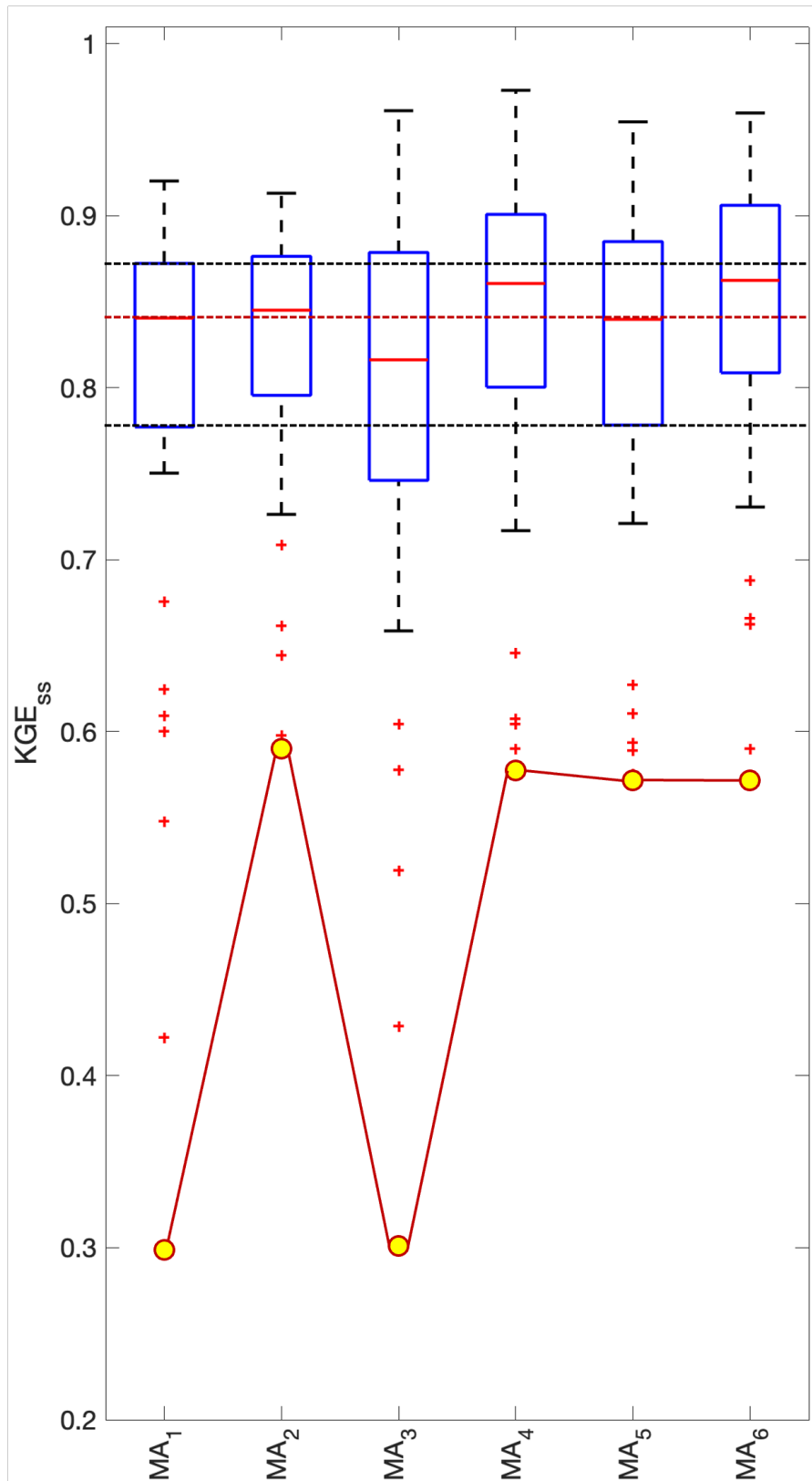

**Figure 3**: Box and whisker plots of distributions of annual $KGE_{ss}$ values for progressively more complex MCP-based architectures. Note that the red dashed horizontal line represents the median (50 percentiles), and the dashed blue lines show the 25 and 75 percentiles of the annual $KGE_{ss}$ for the benchmark $MA_1$ architecture. The yellow circles indicate the worst-year performance of the annual $KGE_{ss}$ for each architecture with red solid lines connected.

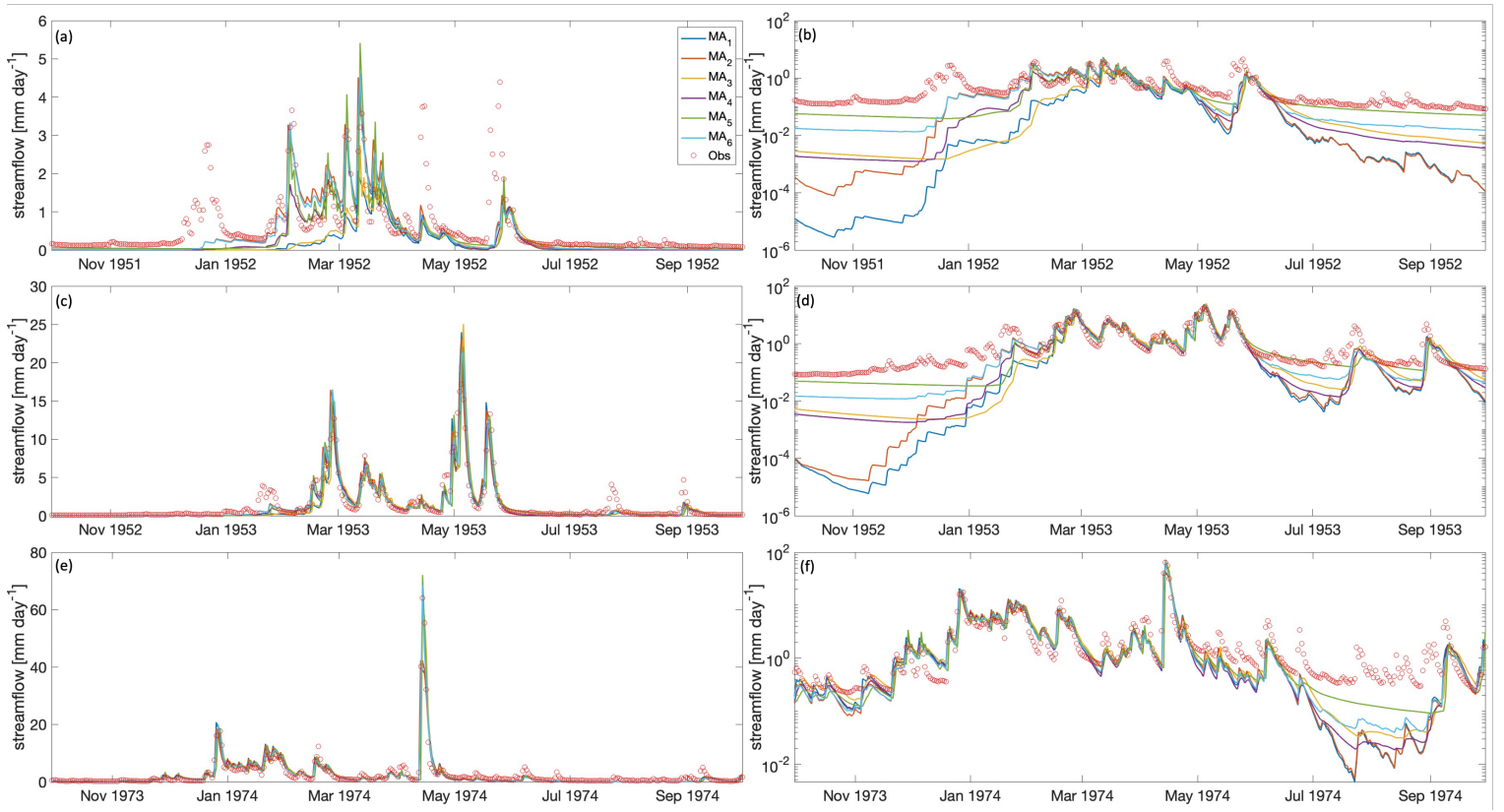

**Figure 4:** Streamflow hydrographs shown in both linear and log space for (a & b) the driest year WY 1952, (c & d) a median year WY 1953, and (e & f) the wettest year WY 1973 based on the annual flow peak. Red circles indicate the observed data, and solid lines indicate simulations provided by various Mass-Conserving architectures ($MA_1$ to $MA_6$).

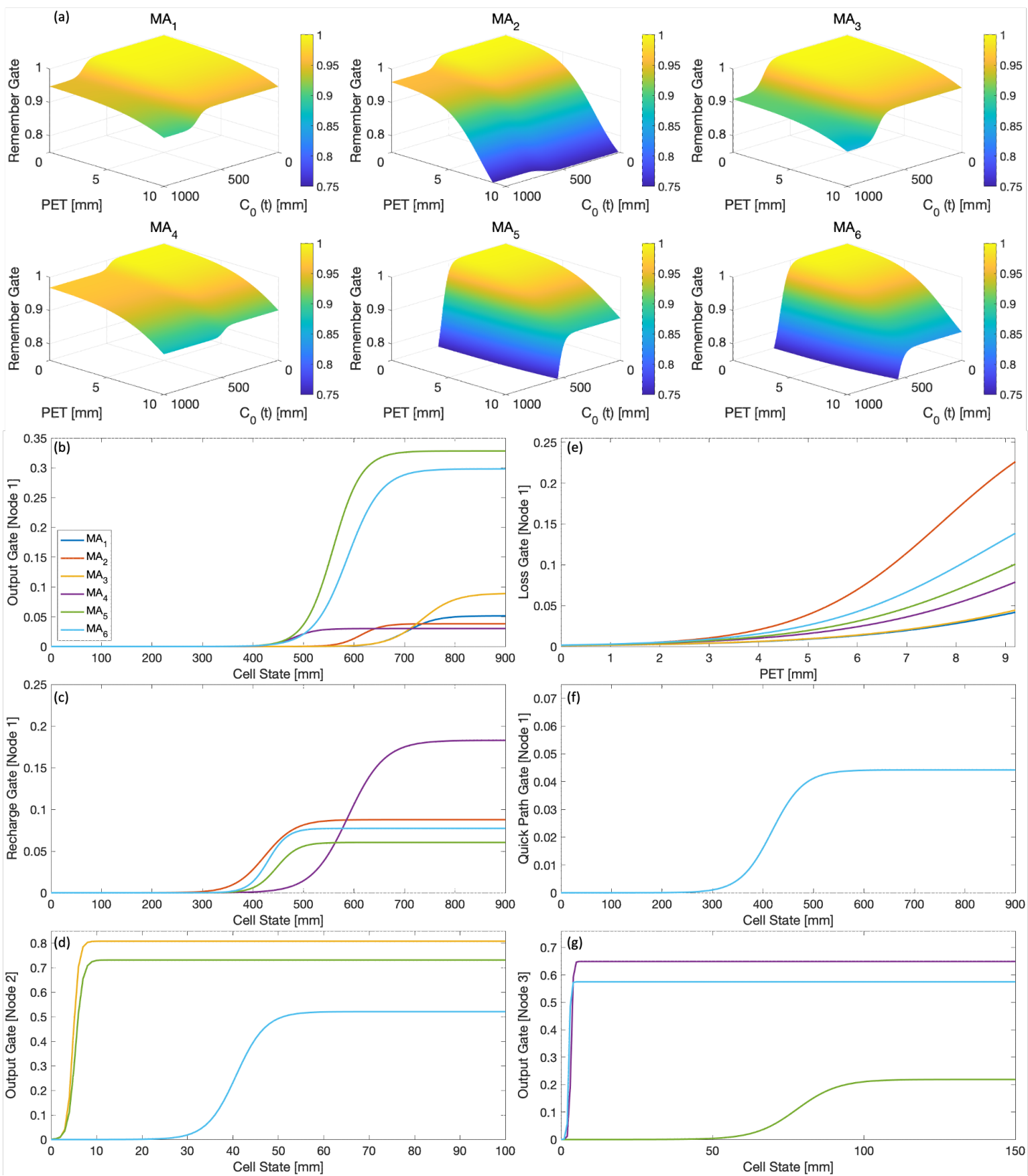

**Figure 5:** Learned gating functions associated with various Mass-Conserving architectures ($MA_1$ to $MA_6$); a) remember gate, b) output gate, c) loss gate, d) recharge gate, and e) quick path gate of node 1, and the f) output gate of node 2 and the g) output gate of node 3.

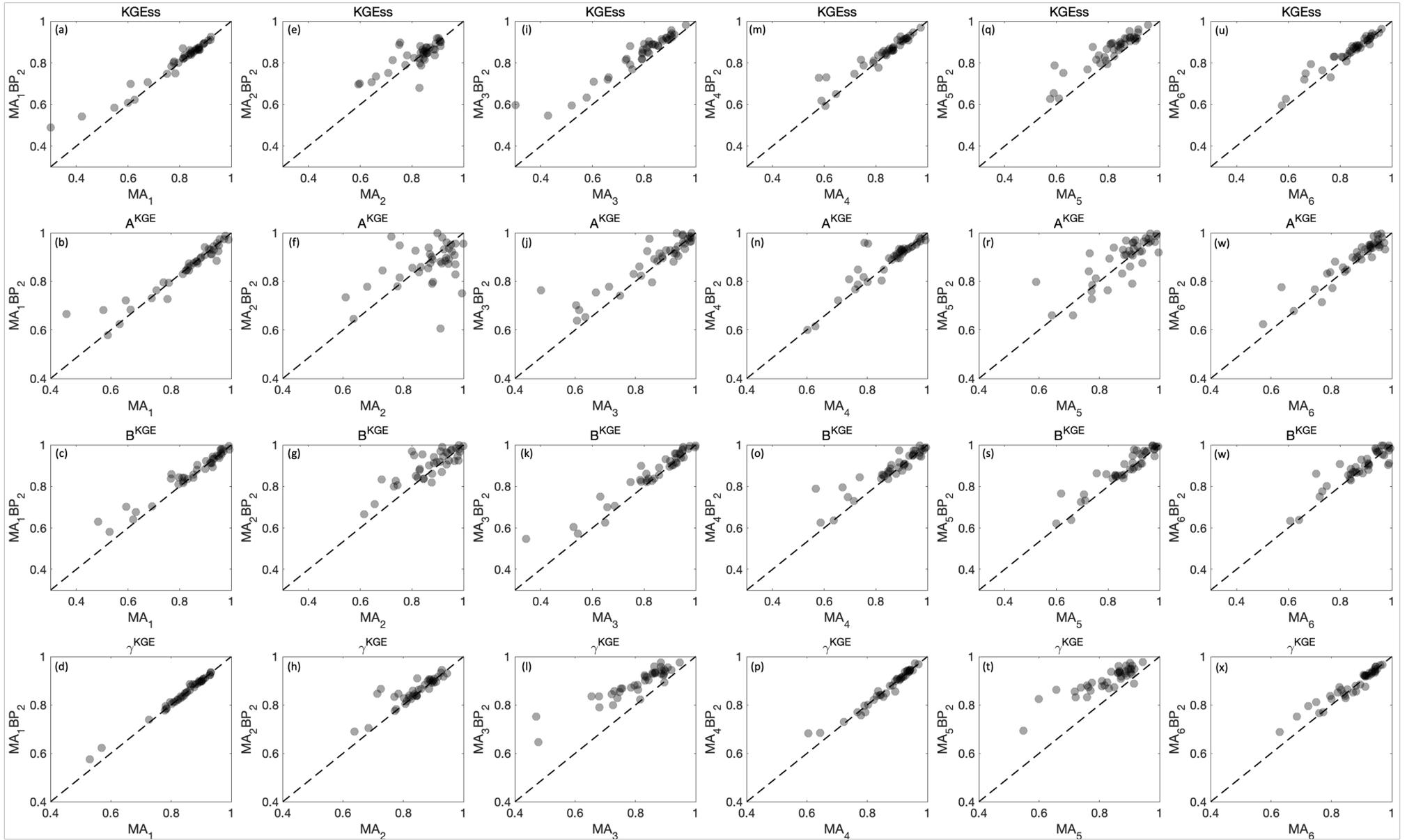

**Figure 6:** Scatter plots of annual $KGE_{ss}$, $A^{KGE}$, $B^{KGE}$ and $\gamma^{KGE}$; (a to d) Comparing $MA_1$ with $MA_1BP_2$; (e to h) Comparing $MA_2$ with $MA_2BP_2$; (i to l) Comparing $MA_3$ with $MA_3BP_2$; (m to p) Comparing $MA_4$ with $MA_4BP_2$; (q to t) Comparing $MA_5$ with $MA_5BP_2$; (u to x) Comparing $MA_6$ with $MA_6BP_2$.

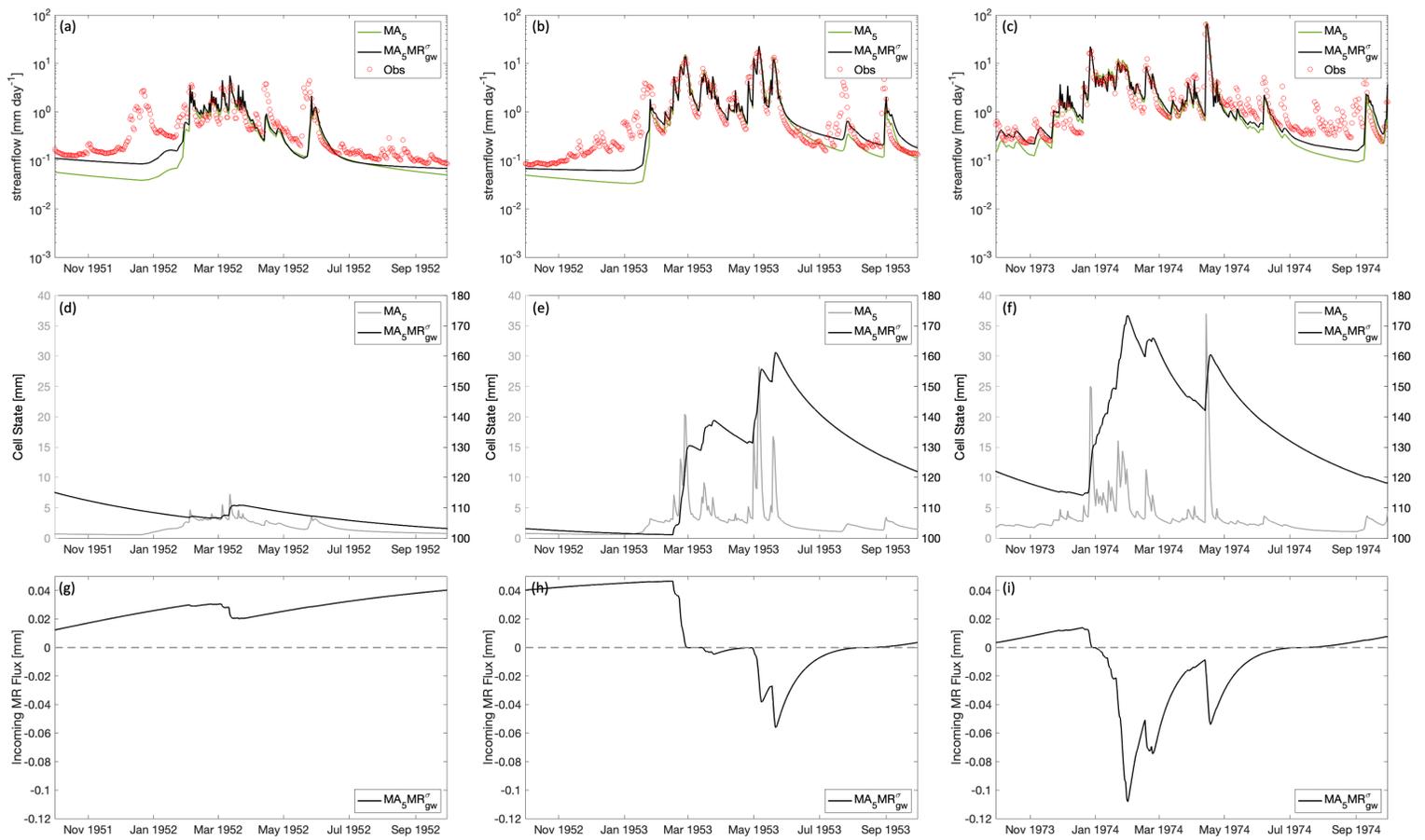

**Figure 7:** Plots showing changes to dry (WY 1952; left column), medium (WY 1953; middle column) and wet (WY 1974; right column) simulations when a mechanism enabling cell-state-dependent bi-directional mass groundwater exchanges is added to the *HyMod* like $MA_5$ architecture. Subplots (a to c) show hydrographs without (solid line) and with (dashed line) mass-relaxation enabled; Subplots (d to e) show corresponding values of the groundwater cell-state; Subplots (g to i) show the magnitude to groundwater mass exchange - positive (negative) values indicate estimated inflows (outflows) to the catchment.

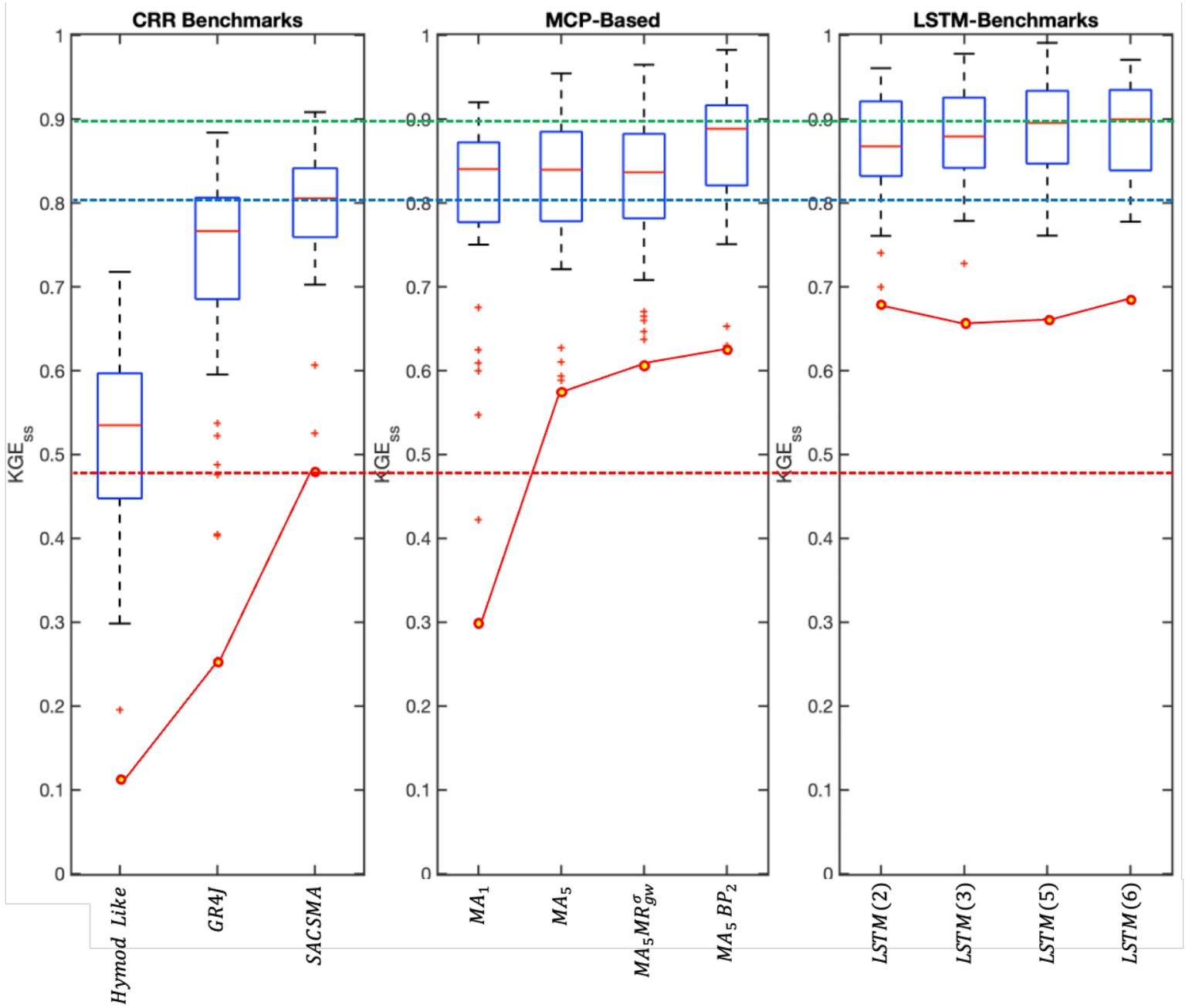

**Figure 8:** Benchmark comparison of four MCP-based mass-conserving architectures (*MAs*; center subplot) with three physical-conceptual rainfall-runoff model architectures (CRRs; left subplot) and four purely data-based single-layer LSTM-based models (right subplot) having different numbers of cell-states ($N_H = 2, 3, 5$ & $6$). All models use exactly the same input drivers (current observed values of precipitation ($U_t$) and potential evapotranspiration ($PE_t$). Box and whisker plots show the 40-year distributions of annual $KGE_{ss}$ performance metric values for each model; the yellow-filled "red dot" indicates the (dry) year with worst $KGE_{ss}$ skill. To aid in comparison, the blue and red dashed lines indicates median- and worst-year performance levels for the "best" CRR model (*SACSMA*), and the green-dashed line indicates median-year performance level for the "best" data-based model (*LSTM(6)*).

Supporting Information for

Towards Interpretable Physical-Conceptual Catchment-Scale Hydrological Modeling using the Mass-Conserving-Perceptron

Yuan-Heng Wang[1,2], Hoshin V. Gupta[1]

[1] Department of Hydrology and Atmospheric Science, The University of Arizona, Tucson, AZ
[2] Environmental Sciences Area, Lawrence Berkeley National Laboratory, Berkeley, CA

Corresponding Author: Yuan-Heng Wang, Ph.D.
Email: yhwang0730@gmail.com | YuanHengWang@lbl.gov | yhwang0730@arizona.edu

Contents of this file

## Introduction

This Supporting Information provides 3 supplementary texts, 3 supplementary tables, and 6 supplementary figures to support the discussions in the main text. The contents of these supplementary materials are as follows.

**Text S1.** Detailed findings regarding the implementation of input-bypass gate in the soil-moisture node system reported in section 6.2 of the manuscript.

**Text S2.** Detailed findings regarding the implementation of mass-relaxation in the groundwater system reported in section 7.2 of the manuscript.

**Text S3.** Summary of benchmark physical-conceptual CRR models in section 8 of the manuscript.

**Table S1.** Summary of parameter inheritance for all the MA model cases listed in this study.

**Table S2.** Summary of skill metrics for the MA models reported in this study.

**Table S3.** Summary of skill metrics for the benchmark models reported in this study.

**Figure S1**. Log-scale flow-duration (sorted flow) curve using 40-year daily streamflow data in Leaf River Basin

**Figure S2.** Accumulative gate flux over time of each Mass-Conserving Architecture (MA) unit including the flux of a) output gate, b) recharge gate, c) quick path gate, d) remember gate, e) loss gate, and f) constraint loss gate of the node 1, and the g) output gate, h) remember gate of node 2, and the i) output gate, j) remember gate of node 3.

**Figure S3.** Scatter plots for the annual performance skill of $KGE_{ss}$, $A^{KGE}$, $B^{KGE}$, $\gamma^{KGE}$ between $MA_N$ and $MA_N BP_1$ where $N = 1$ to 6.

**Figure S4**. Comparison of hydrographs with the implementation of $MR_{gw}^{\sigma}$ on $MA_4$ to $MA_6$ architectures in a dry year WY 1952 (subplot a to c), median year WY 1953 (subplot d to f), and wet year WY 1974 (subplot g to i).

**Figure S5:** Box and whisker plots of the 40-year distributions of annual $KGE_{ss}$ performance metric values for various node numbers ($N_H = 1$ to $10$) of single-layer $LSTM$ networks.

**Figure S6:** Time series plots of the $MA_5$ architecture that includes soil moisture ($SM$) state in dry, median, and wet year from subplot (a) to (c), actual evapotranspirative loss ($AET$) from subplot (d) to (f), outflow from soil-moisture tank ($O_t^{SM}$) from subplot (g) to (i),

outflow from surface channel routing tank ($O_t^{CH}$) from subplot (j) to (l), outflow from groundwater tank ($O_t^{GW}$) from subplot (m) to (o).

**Figure S7:** Box and whisker plots of distributions of annual $KGE_{ss}$ values for incorporating input-bypass gate for the six mass-conserving architectures.

**Text S1: Detailed findings regarding the implementation of input-bypass gate in the soil-moisture node system reported in section 6.2 of the manuscript.**

*Model Architecture $MA_1$ (Single Cell-States, Single Flow Pathway) with Input-Bypass Gate*

Implementation of the second type of input-bypass gate ($MA_1 - BP_2$) was found to be more useful than the first type ($MA_1 - BP_1$) for improving performance of the $MA_1$ architecture (**Figure 5S**). We found that performance remains very similar between $MA_1$ and $MA_1 - BP_1$ for most of the percentiles (**Figure S3a** to **Figure S3b**) with decreases for $KGE_{SS}^{\min}$ from $0.30 \rightarrow 0.28$ and increases for $KGE_{SS}^{95\%}$ from $0.91 \rightarrow 0.92$. In contrast, the use of $BP_2$ on $MA_1$ helps to improve the performance, especially for below median percentile (**Table 1**), such as $KGE_{SS}^{\min}$ ($0.30 \rightarrow 0.49$), $KGE_{SS}^{5\%}$ ($0.48 \rightarrow 0.56$), and $KGE_{SS}^{25\%}$ ($0.78 \rightarrow 0.79$). This reflects somehow evenly on the main $KGE$ (**Figure 6a**) and all its three components (**Figure 6b** to **Figure 6d**). Overall, the results show that it is necessary to consider the storage capacity of soil moisture tank and it seems the $BP_2$ interpretation (due to precipitation intensity exceeding some threshold rate) can be a better reasonable interpretation than the $BP_1$ (due to the storage filling up and becoming saturated).

*Model Architecture $MA_2$ (Single Cell-States, Two Flow Pathway) with Input-Bypass Gate*

We found a very similar tendency of performance improvement based on box plots between $MA_2$ and $MA_1$ when implementing both input bypass gating mechanisms. The use of $BP_2$ shows a larger and more significant improvement than $BP_1$ and especially note that the $KGE_{SS}^{\min}$ (0.68) and $KGE_{SS}^{5\%}$ (0.70) skills of $MA_2 - BP_2$ are among the best out of all the architectures listed in **Table 1** including those with multi-state and multi-path architectures. The $MA_2 - BP_2$ outperforms $MA_2$ mostly in mass balance (**Figure 6g**), and correlation (**Figure 6h**) compared to variability (**Figure 6f**). Since implementing a bypass is similar to considering a new flow route when the soil tank is filled or saturated, the result implies the relative importance of adding a flow path over that of adding a cell state (node), especially for the low percentile years.

*Model Architecture $MA_3$ (Two Cell-States, Single Flow Pathway) with Input-Bypass Gate*

We show the scatter plot of $MA_3 - BP_2$ vs. $MA_3$ including $KGE_{SS}$ and three components of $KGE$ from **Figure 6i** to **Figure 6l**. Each black dot represents the annual skill from total 40-years and if the dots are above the 1 to 1 line (black dash line) then it means the $MA_3 - BP_2$ outperforms the $MA_3$ for that specific year and vice versa. The main take-away here is that the inclusion of input-bypass ($MA_3 - BP_2$) on the flow routing node enhances model performance by improving the overall $KGE_{SS}$ (**Figure 6i**) and the timing (**Figure 6l**) of the streamflow hydrograph. We notice that mass balance (**Figure 6k**) and the shape of the streamflow hydrograph (**Figure 6j**) improve particularly for the dry to medium years. Hence, we report more details through comparing the cases with or without the input-bypass gating in **Figure S7**. The performance improvement achieved by implementing the first type of bypass ($MA_3 - BP_1$) is limited in that only $KGE_{SS}^{25\%}$ and $KGE_{SS}^{50\%}$ improve to 0.76, and 0.84 respectively.

Particularly noticeable for the $MA_3 - BP_2$ is the extent to which performance improves in the high percentile (wetter) years, with $KGE_{SS}^{75\%}$ improving from $0.88 \rightarrow 0.91$ and $KGE_{SS}^{95\%}$ improving from $0.91 \rightarrow 0.95$. Also compared with $MA_1$, we see performance improvement in the worst year ($KGE_{SS}^{worst}$ improves from $0.30 \rightarrow 0.55$), but a decline relative to $MA_2$. Further examination revealed that the two years with the largest improvement were both dry years, with a $\Delta KGE_{SS} =$

$0.30$ improvement for the lowest average flow year (WY 1952) and a $\Delta KGE_{SS} = 0.12$ improvement for WY 1963 which has the smallest average value for annual discharge.

In addition, relative to $MA_1$, significant improvements were seen for $MA_3-BP_2$ in the components of $KGE$, with 29 of the years showing improved values for $\alpha^{KGE}$ (variability ratio), 27 years showing improved values for $\beta^{KGE}$ (mass-balance ratio) and 37 years showing improved values for $\gamma^{KGE}$ (correlation). Overall, the 40-year average $1-|1-\alpha^{KGE}|$ improved from $0.85 \rightarrow 0.89$, the 40-year average $1-|1-\beta^{KGE}|$ improved from $0.83 \rightarrow 0.86$, and the 40-year average $\gamma^{KGE}$ improved from $0.84 \rightarrow 0.89$.

The results suggest that the LRB is better represented by more than 1 flow path, and that the soil moisture tank without bypass (i.e., $MA_3$) representing the saturation-excess overflow is not a proper assumption (i.e., the soil moisture tank should be treated as having a finite storage capacity as $MA_3-BP_1$ or $MA_3-BP_2$.

### *Model Architecture $MA_4$ (Two Cell-States, Two Flow Pathways) with Input-Bypass Gate*

There is a less amount of performance improvement of implementing input-bypass gating on the $MA_4$ than $MA_3$ architecture. Though we still observed the $KGE_{SS}$ improvement at the below-medium quantile years (**Figure 6m)** where the shape of the hydrograph (**Figure 6n**) and the mass balance ratio (**Figure 6o**) contribute to the skill enhancement as opposed to the timing of the hydrograph (**Figure 6p**), the improvement is limited in the sense that only $\alpha^{KGE}$ shows the skill elevates in the particular flow group (i.e., the two smallest) but this is not even necessarily for the $\beta^{KGE}$ of $MA_4-BP_2$ (**Table 2**).

From **Figure S7**, both $MA_4-BP_1$ and $MA_4-BP_2$ outperform the original $MA_4$ case. The former provides a tighter annual $KGE_{ss}$ distribution and the latter shows an overall skill increase. Although $MA_4-BP_2$ seems to perform quite well, it does not completely outperform $MA_3-BP_2$, having lower $KGE_{ss}^{median}$ ($0.86\ vs.\ 0.88$) and $KGE_{SS}^{95\%}$ ($0.95\ vs.\ 0.94$). The result implies the deficiency of merely using two-nodes and so the high possibility of requiring three nodes (state variables) for simulating the R-R dynamics in LRB.

### *Model Architecture $MA_5$ (Three Cell-States, Two Flow Pathways) with Input-Bypass Gate*

As shown earlier, the model $MA_5$ overall only outperforms the $MA_3$ for the case without considering the capacity of the soil-moisture tank. The implementation of bypass functions helps to elevate the performance, where even $MA_5-BP_1$ improves the interquartile range of the skill distribution of $MA_5$. The $MA_5-BP_2$ has higher $KGE_{SS}$ (**Figure 6q**) than $MA_5$ which came from the overall improvement brought from the $\beta^{KGE}$ (**Figure 6s**) and $\gamma^{KGE}$ (**Figure 6t**) components across all the flow regimes.

Compared $MA_5-BP_2$ with the two-node $MA_4-BP_2$ architecture we see clear improvements in annual $KGE_{SS}$, with the distribution shifting upwards towards the optimal value of 1.0, $KGE_{ss}$ is improved in 30 of the years, thereby clearly supporting the three-node hypothesis. Notably, low-to-high year performance values are mostly improved. The mass-balance ratio ($\beta^{KGE}$) is improved in 16 of the years, variability ratio ($\alpha^{KGE}$) is improved in 22 of the years. The most consistent improvement is seen in the correlation coefficient ($\gamma^{KGE}$; timing and shape), with improvements in 33 of the 40 years.

It should be noted however that in **Table 2** there still contains skill deterioration for $MA_5 - BP_2$ when compared to $MA_5$ for the smallest flow group for $\alpha^{KGE}$ and $\beta^{KGE}$ suggesting the consideration of balancing the enhancement of model architecture in both the surface and subsurface components. Overall, incorporating both groundwater and surface water routing components synergistically achieves the combined benefits that were realized by $MA_3$ and $MA_4$ types of architecture and improves over both cases.

### Model Architecture $MA_6$ *(Three Cell-States, Three Flow Pathways) with Input-Bypass Gate*

**Figure 6u** shows the overall improvement of implementing input-bypass gate is marginal for $MA_6$ compared to the $MA_3$ or $MA_5$. Based on Table 2, the major improvement appears to be the low flow regime of the mass balance ratio ($\beta^{KGE}$; **Figure 6w**) as well as the timing of the hydrograph ($\gamma^{KGE}$; **Figure 6x**). Namely, the very fast flow path is not fully capable of functionally regulating the soil-moisture tank capacity which results in the relatively poor performance in this regard.

The basic $MA_6$ architecture even has higher $KGE_{ss}$ than $MA_5 - BP_1$ for totally 29 years. Although it performs worse than $MA_5 - BP_2$, this finding provides strong evidence that the LRB might need more than two flow paths since the $MA_6$ architecture can be interpreted as that $MA_5$ with bypass but the connection is skipped pass over the surface routing tank. Notice however that the input bypass and the quick flow path are two different encoded assumptions, and the similar skill might potentially lead to multiple plausible interpretations of the system, which should be further investigated.

Since the use of input bypass gate improves the original architecture for the $MA_6$ case, we therefore further compare $MA_6 - BP_2$ against $MA_5 - BP_2$. We see that $MA_6 - BP_2$ performs better than $MA_5 - BP_2$ for $KGE_{ss}^{25\%}$, and $KGE_{ss}^{median}$ but not the rest of the listed percentile years (**Table 1**). Further, $MA_6 - BP_2$ has overall better performance for the two smallest flow groups (**Table 2**). The variability ratio ($\alpha^{KGE}$) improves in 19 of the years, the mass-balance ratio ($\beta^{KGE}$) improves in 22 of the years, but the correlation coefficient only improves in 1 of the years. The poor correlation may come from the fact that there are some compensations occurred between overflow of soil-moisture tank and the quick surface flow pathways. Overall, balancing model performance against additional architectural complexity, the results do not fully support adoption of the three-flow-pathway hypothesis. Of course, this could change if the climate of the catchment changed so that the prevalence of fast-overland flow during high intensity precipitation were to increase.

**Text S2: Detailed findings regarding the implementation of mass-relaxation in the groundwater system reported in section 7.2 of the manuscript.**

The major observations include:

- There seems to have compensation between architectural component occurred in medium and wet year. For instance, there is a skill deterioration between June 1953 to September 1953 in $MA_5MR_{gw}^{\sigma}$ (**Figure 7b**) since the $MR_{gw}^{\sigma}$ has learned to loose water in the groundwater system but the positive bias is still increasing.

- There is no further significant skill deterioration found at high flow in $MA_6MR_{gw}^{\sigma}$ (**Figure S3g** to **Figure S3i**). The result suggests $MA_6MR_{gw}^{\sigma}$ leads to higher robustness in high flow than $MA_5MR_{gw}^{\sigma}$ that is likely because that has one more flow path in the surface (quick path gate) so the entire model architecture is less sensitive to the implementation of $MR_{gw}^{\sigma}$ in the groundwater node.

- The low flow improvement for the $MA_6MR_{gw}^{\sigma}$ covers relative fewer timesteps than other two cases during the wet year (**Figure S3g** to **Figure S3i**) which is likely because the original $MA_6$ hydrograph has a relative accurate flow scale at the low flow. However, the $MA_6$ (as well as $MA_4$) family of architecture is found to have flashy cell state time series in all years that does not behave as like the internal storage of groundwater tank (not shown). This is because these two cases do not have a sufficient physical interpretable output gate function in the groundwater node (**Figure 5g**) where it tends to release water very quickly compared to the output gate of $MA_5$ family.

- The boxplot (**Figure S4**) generated based on annual statistics shows that the overall skill when implementing groundwater mass relaxation on $MA_4MR_{gw}^{\sigma}$ to $MA_6MR_{gw}^{\sigma}$ becomes slightly worse than its original architecture of $MA_4$ to $MA_6$. The scatter plot in **Figure S7** shows that the skills of $MA_4MR_{gw}^{\sigma}$ including $KGE_{ss}$ and all three $KGE$ components slightly decrease in the low percentile years. The $MA_5MR_{gw}^{\sigma}$ however shows improvement in several years for all skills especially for the $\beta^{KGE}$ component.

- The $MA_6MR_{gw}^{\sigma}$ has the minimum differences where most of the years maintain very similar skills compared to $MA_6$. The result suggests the use of more than single metric for performance evaluation is necessary for model evaluation (Vrugt, 2023), and the further investigation of using different training metric (Gupta et al., 1998) is useful to understand how the architectural component and training objective will have interactive effect on the entire hydrograph.

**Text S3. Summary of benchmark physical-conceptual CRR models in section 8 of the manuscript.**

This section summarizes the information of CRR models including several well-established architectures such as $HyMOD\ Like$, $GR4J$, and $SACSMA$ models. We list all the associated parameters and state variables used for each approach and provide a comprehensive description regarding their physical meaning from below **Table TS3.1** to **Table TS3.3**.

**Table TS3.1:** Summary of parameters and state variables of $HyMOD$ Like model used for this study

| HyMOD Like | | |
|---|---|---|
| Parameters | Description (unit) | Range Used for Optimization |
| $\kappa_o^{SM}$ | Output gate parameters in soil-moisture tank | |
| $\kappa_l^{SM}$ | | |
| $a_o^{SM}$ | Loss gate parameters in soil-moisture tank | |
| $b_o^{SM}$ | | Initialize between -1 to 1 without constrain on the value range during search |
| $\kappa_r^{SM}$ | Remember gate parameter in soil-moisture tank | |
| $\kappa_o^{CH}$ | Output gate parameter in channel routing tank | |
| $\kappa_o^{GW}$ | Output gate parameter in groundwater tank | |
| State Variables | Description (unit) | Initialization |
| $x_t^{SM}$ | state variables in soil-moisture tank | |
| $x_t^{CH}$ | state variables in channel routing tank | Initialize from 0 with follow up 3-year spin up |
| $x_t^{GW}$ | state variables in groundwater tank | |

**Table TS3.2:** Summary of parameters and state variables of GR4J model used for this study.

| GR4J | | |
|---|---|---|
| Parameters | Description (unit) | Range Used for Optimization |
| $x_1$ | Maximum Capacity of Production Store (mm) | 1-5000 |
| $x_2$ | Groundwater change coefficient (mm) | -1.0-1.0 |
| $x_3$ | Maximum store routing capacity (mm) | 1-1500 |
| $x_4$ | Time peak ordinate of the unit hydrograph (mm) | 0.501-4.50 |
| State Variables | Description (unit) | Initialization |
| $x_t^p$ | Production Store state | Initialize from 0 with follow up 3-year spin up |
| $x_t^r$ | Routing Store State | |

**Table TS3.3:** Summary of parameters and state variables of SACSMA model used for this study.

| | SACSMA | |
|---|---|---|
| Parameters | Description unit | Range Used for Optimization |
| $UZTWM$ | Maximum capacity of the upper zone tension water storage, mm | 10-56 |
| $UZFWM$ | Maximum capacity of the upper zone free water storage, mm | 10-46 |
| $LZTWM$ | Maximum capacity of the lower zone tension water storage, mm | 75-131 |
| $LZFPM$ | Maximum capacity of the lower zone free water primary storage, mm | 50-162 |
| $LZFSM$ | Maximum capacity of the lower zone free water supplemental storage, mm | 10-23 |
| $ADIMP$ | Additional impervious area, decimal fraction | 0-0.173 |
| $UZK$ | Upper zone free water lateral depletion rate, $day^{-1}$ | 0.2-0.245 |
| $LZPK$ | Lower zone primary free water depletion rate, $day^{-1}$ | 0.001-0.009 |
| $LZSK$ | Lower zone supplemental free water depletion rate, $day^{-1}$ | 0.02-0.043 |
| $PCTIM$ | Impervious fraction of the watershed area, decimal fraction | 0-0.043 |
| $ZPERC$ | Maximum percolation rate, dimensionless | 5-226 |
| $REXP$ | Exponent of the percolation equation, dimensionless | 1.1-3.65 |
| $PFREE$ | Fraction of water percolating from upper zone which goes directly to lower zone free water storage, decimal fraction | 0-0.063 |
| $RSERV$ | Fraction of lower zone free water not transferable to lower zone tension water, decimal fraction | 0.29-0.31 |
| $RIVA$ | Riparian vegetation area, decimal fraction | 0-0.01 |
| $SIDE$ | Ratio of deep recharge to channel base flow, dimensionless | 0-0.01 |
| State Variables | Description (unit) | Initialization |
| $UZTWC_t$ | upper zone tension water contents, mm | |
| $UZFWC_t$ | upper zone free water contents, mm | |
| $LZTWC_t$ | lower zone tension water contents, mm | Initialize from 0 with follow up 3-year spin up |
| $LZFPC_t$ | lower zone free primary contents, mm | |
| $LZFSC_t$ | lower zone free supplemental contents, mm | |
| $ADIMC_t$ | tension water contents of the ADIMP area, mm | |

**Table S1:** Summary of parameters inheritance for all the MA model cases listed in this study.

| Model Name | Model Inheritance | Parameter Inheritance | Random Seeds No. | Epoch | Input Standardization in Output/Loss Gate | BP type |
|---|---|---|---|---|---|---|
| $MA_1$ | $MC\{O_\sigma L_\sigma^{con}\}$ | NA | NA | NA | Yes | NA |
| $MA_2$ | $MA_1$ | All Weights and Bias+ | 10 | 2000 | Yes | NA |
| $MA_3^*$ | $MA_1$ | All Weights and Bias+ | 10 | 2000 | Yes | NA |
| $MA_3$ | $MA_1$ | All Weights and Bias+ | 10 | 2000 | Yes | NA |
| | $MA_3^*$ | Scaling factors of cell state in Surface Routing Tank | | | | |
| $MA_4^*$ | $MA_2$ | All Weights and Bias+ | 10 | 2000 | Yes | NA |
| $MA_4$ | $MA_2$ | All Weights and Bias+ | 10 | 2000 | Yes | NA |
| $MA_4$ | $MA_2^*$ | All Weights and Bias+ | 10 | 2000 | Yes | NA |
| $MA_5$ | $MA_2$ | Weights and Bias in Soil Moisture Tank | NA | 2000 | Yes | NA |
| | $MA_3$ | Weights and Bias in Surface Routing Tank | | | | |
| | $MA_4$ | Weights and Bias in Groundwater Tank | | | | |
| $MA_6$ | $MA_5$ | All Weights and Bias+ | 10 | 2000 | Yes | NA |
| $MA_1 - BP$ | $MA_1$ | All Weights and Bias+ | 10 | 2000 | Yes | $BP_1/BP_2$ |
| $MA_2 - BP$ | $MA_2$ | All Weights and Bias+ | 10 | 2000 | Yes | $BP_1/BP_2$ |
| $MA_3 - BP$ | $MA_3$ | All Weights and Bias+ | 10 | 2000 | Yes | $BP_1/BP_2$ |
| $MA_4 - BP$ | $MA_4$ | All Weights and Bias+ | 10 | 2000 | Yes | $BP_1/BP_2$ |
| $MA_5 - BP$ | $MA_5$ | All Weights and Bias+ | 10 | 2000 | Yes | $BP_1/BP_2$ |
| $MA_6 - BP$ | $MA_6$ | All Weights and Bias+ | 10 | 2000 | Yes | $BP_1/BP_2$ |

Notice that "All Weights and Bias+" represents all the parameters but without including the parameters in the newly added components for the target model. $MC\{O_\sigma L_\sigma^{con}\}$ is the single node generic case proposed in Wang and Gupta (2024) that denotes as $MCA_1$ in current study. $MCA_3^*$ and $MCA_4^*$ refer to the case of constant output gating in the routing and groundwater node respectively.

**Table S1:** Summary of parameters inheritance for all the MA model cases listed in this study (Part 2).

| Model Name | Model Inheritance | Parameter Inheritance | Random Seeds No. | Epoch | Input Standardization in Output/Loss Gate |
|---|---|---|---|---|---|
| $MA_4 - MR_{\mathrm{gw}}$ | $MA_4$ | All Weights and Bias+ | 10 | 2000 | Yes |
| $MA_5 - MR_{\mathrm{gw}}$ | $MA_5$ | All Weights and Bias+ | 10 | 2000 | Yes |
| $MA_6 - MR_{\mathrm{gw}}$ | $MA_6$ | All Weights and Bias+ | 10 | 2000 | Yes |

Notice that "All Weights and Bias+" represents all the parameters but without including the parameters in the newly added components for the target model. $MC\{O_\kappa L_\sigma\}$ is the single node case with time-constant output gate and time-variable loss gate (without constraining on the loss gate flux) proposed in Wang and Gupta (2024) that denotes as $MA_1 - PC^*$ in current study. $MA_3^*$ and $MA_4^*$ refer to the case of constant output gating in the routing and groundwater node respectively. The asterisk superscript refers to the cases without constraining the evapotranspirative flux.

**Table S2:** Summary skill metrics for the MA models reported in this study. Numbers in *red color* indicate skill < 0.5, and numbers in *blue color* indicate $0.85 \leq$ skill $\leq 1.0$.

| Percentiles | $MA_1$ | $MA_1BP_1$ | $MA_1BP_2$ | $MA_2$ | $MA_2BP_1$ | $MA_2BP_2$ | $MA_3$ | $MA_3BP_1$ | $MA_3BP_2$ | $MA_4$ | $MA_4BP_1$ | $MA_4BP_2$ | $MA_5$ | $MA_5BP_1$ | $MA_5BP_2$ | $MA_6$ | $MA_6BP_1$ | $MA_6BP_2$ |
|---|---|---|---|---|---|---|---|---|---|---|---|---|---|---|---|---|---|---|
| *KGE* | | | | | | | | | | | | | | | | | | |
| $KGE_{ss}^{worst}$ | 0.01 | -0.02 | 0.28 | 0.42 | 0.37 | 0.55 | 0.01 | 0.40 | 0.36 | 0.41 | 0.40 | 0.42 | 0.40 | 0.30 | 0.47 | 0.40 | 0.33 | 0.43 |
| $KGE_{ss}^{5\%}$ | 0.27 | 0.25 | 0.38 | 0.46 | 0.42 | 0.57 | 0.26 | 0.47 | 0.43 | 0.43 | 0.55 | 0.48 | 0.42 | 0.37 | 0.49 | 0.47 | 0.49 | 0.54 |
| $KGE_{ss}^{25\%}$ | 0.68 | 0.68 | 0.70 | 0.71 | 0.71 | 0.74 | 0.64 | 0.66 | 0.74 | 0.72 | 0.74 | 0.74 | 0.69 | 0.71 | 0.75 | 0.73 | 0.69 | 0.76 |
| $KGE_{ss}^{50\%}$ | 0.77 | 0.77 | 0.78 | 0.78 | 0.79 | 0.79 | 0.74 | 0.77 | 0.83 | 0.80 | 0.78 | 0.80 | 0.77 | 0.77 | 0.84 | 0.81 | 0.77 | 0.82 |
| $KGE_{ss}^{75\%}$ | 0.82 | 0.82 | 0.81 | 0.83 | 0.82 | 0.83 | 0.83 | 0.82 | 0.87 | 0.86 | 0.81 | 0.87 | 0.84 | 0.81 | 0.88 | 0.87 | 0.83 | 0.88 |
| $KGE_{ss}^{95\%}$ | 0.88 | 0.88 | 0.87 | 0.87 | 0.88 | 0.88 | 0.88 | 0.87 | 0.93 | 0.91 | 0.87 | 0.91 | 0.88 | 0.87 | 0.94 | 0.92 | 0.88 | 0.92 |
| $A^{KGE}$ | | | | | | | | | | | | | | | | | | |
| $KGE_{ss}^{worst}$ | 0.45 | 0.43 | 0.58 | 0.61 | 0.57 | 0.61 | 0.49 | 0.62 | 0.64 | 0.60 | 0.57 | 0.60 | 0.59 | 0.52 | 0.66 | 0.57 | 0.44 | 0.62 |
| $KGE_{ss}^{5\%}$ | 0.58 | 0.58 | 0.64 | 0.66 | 0.66 | 0.69 | 0.60 | 0.64 | 0.67 | 0.67 | 0.73 | 0.67 | 0.68 | 0.66 | 0.69 | 0.65 | 0.69 | 0.70 |
| $KGE_{ss}^{25\%}$ | 0.79 | 0.81 | 0.79 | 0.85 | 0.85 | 0.83 | 0.82 | 0.76 | 0.83 | 0.80 | 0.85 | 0.85 | 0.83 | 0.83 | 0.84 | 0.84 | 0.86 | 0.85 |
| $KGE_{ss}^{50\%}$ | 0.89 | 0.89 | 0.88 | 0.91 | 0.91 | 0.88 | 0.91 | 0.85 | 0.92 | 0.91 | 0.88 | 0.92 | 0.91 | 0.87 | 0.92 | 0.92 | 0.90 | 0.93 |
| $KGE_{ss}^{75\%}$ | 0.94 | 0.94 | 0.93 | 0.95 | 0.96 | 0.93 | 0.96 | 0.93 | 0.96 | 0.93 | 0.92 | 0.94 | 0.94 | 0.91 | 0.96 | 0.95 | 0.95 | 0.95 |
| $KGE_{ss}^{95\%}$ | 0.98 | 0.98 | 0.97 | 0.98 | 0.99 | 0.98 | 0.99 | 0.99 | 1.00 | 0.98 | 0.98 | 0.98 | 0.99 | 0.99 | 0.99 | 0.97 | 0.99 | 0.99 |
| $B^{KGE}$ | | | | | | | | | | | | | | | | | | |
| $KGE_{ss}^{worst}$ | 0.30 | 0.28 | 0.49 | 0.61 | 0.57 | 0.67 | 0.34 | 0.69 | 0.55 | 0.57 | 0.72 | 0.63 | 0.60 | 0.64 | 0.62 | 0.61 | 0.65 | 0.64 |
| $KGE_{ss}^{5\%}$ | 0.51 | 0.49 | 0.61 | 0.67 | 0.64 | 0.76 | 0.53 | 0.73 | 0.59 | 0.61 | 0.78 | 0.68 | 0.64 | 0.79 | 0.68 | 0.67 | 0.72 | 0.69 |
| $KGE_{ss}^{25\%}$ | 0.80 | 0.80 | 0.83 | 0.83 | 0.84 | 0.85 | 0.79 | 0.86 | 0.83 | 0.83 | 0.87 | 0.85 | 0.83 | 0.86 | 0.85 | 0.84 | 0.83 | 0.86 |
| $KGE_{ss}^{50\%}$ | 0.88 | 0.89 | 0.89 | 0.89 | 0.89 | 0.92 | 0.89 | 0.91 | 0.90 | 0.89 | 0.91 | 0.90 | 0.90 | 0.92 | 0.90 | 0.90 | 0.88 | 0.90 |
| $KGE_{ss}^{75\%}$ | 0.95 | 0.95 | 0.95 | 0.95 | 0.95 | 0.97 | 0.94 | 0.96 | 0.96 | 0.95 | 0.95 | 0.95 | 0.94 | 0.95 | 0.97 | 0.96 | 0.94 | 0.97 |
| $KGE_{ss}^{95\%}$ | 0.98 | 0.99 | 0.98 | 0.98 | 0.99 | 0.99 | 0.99 | 0.99 | 0.99 | 0.99 | 1.00 | 0.99 | 0.98 | 0.99 | 0.99 | 0.99 | 0.99 | 1.00 |
| $r^{KGE}$ | | | | | | | | | | | | | | | | | | |
| $KGE_{ss}^{worst}$ | 0.53 | 0.52 | 0.57 | 0.64 | 0.63 | 0.69 | 0.47 | 0.48 | 0.65 | 0.60 | 0.66 | 0.68 | 0.55 | 0.49 | 0.69 | 0.63 | 0.59 | 0.69 |
| $KGE_{ss}^{5\%}$ | 0.65 | 0.65 | 0.68 | 0.70 | 0.70 | 0.74 | 0.57 | 0.61 | 0.77 | 0.68 | 0.69 | 0.71 | 0.63 | 0.59 | 0.83 | 0.70 | 0.63 | 0.76 |
| $KGE_{ss}^{25\%}$ | 0.81 | 0.81 | 0.81 | 0.81 | 0.81 | 0.83 | 0.75 | 0.79 | 0.86 | 0.83 | 0.80 | 0.84 | 0.77 | 0.78 | 0.87 | 0.83 | 0.78 | 0.84 |
| $KGE_{ss}^{50\%}$ | 0.86 | 0.86 | 0.86 | 0.85 | 0.84 | 0.86 | 0.83 | 0.85 | 0.91 | 0.90 | 0.84 | 0.90 | 0.86 | 0.84 | 0.92 | 0.91 | 0.86 | 0.90 |
| $KGE_{ss}^{75\%}$ | 0.90 | 0.90 | 0.90 | 0.90 | 0.89 | 0.90 | 0.89 | 0.90 | 0.93 | 0.93 | 0.90 | 0.94 | 0.89 | 0.90 | 0.94 | 0.93 | 0.89 | 0.93 |
| $KGE_{ss}^{95\%}$ | 0.93 | 0.93 | 0.93 | 0.93 | 0.93 | 0.93 | 0.91 | 0.92 | 0.97 | 0.96 | 0.93 | 0.96 | 0.92 | 0.92 | 0.97 | 0.96 | 0.93 | 0.96 |

**Table S2:** Summary skill metrics for the MA models reported in this study (Part 2). For $NSE$, numbers in *red color* indicate skill < 0.5, and numbers in *blue color* indicate $0.85 \leq$ skill $\leq 1.0$.

| Percentiles | $MA_1$ | $MA_1BP_1$ | $MA_1BP_2$ | $MA_2$ | $MA_2BP_1$ | $MA_2BP_2$ | $MA_3$ | $MA_3BP_1$ | $MA_3BP_2$ | $MA_4$ | $MA_4BP_1$ | $MA_4BP_2$ | $MA_5$ | $MA_5BP_1$ | $MA_5BP_2$ | $MA_6$ | $MA_6BP_1$ | $MA_6BP_2$ |
|---|---|---|---|---|---|---|---|---|---|---|---|---|---|---|---|---|---|---|
| | | | | | | | $NSE$ | | | | | | | | | | | |
| $KGE_{ss}^{worst}$ | 0.03 | 0.01 | 0.24 | 0.34 | 0.33 | 0.22 | -0.02 | -0.17 | 0.37 | 0.28 | 0.25 | 0.31 | 0.21 | 0.09 | 0.46 | 0.33 | 0.17 | 0.43 |
| $KGE_{ss}^{5\%}$ | 0.30 | 0.31 | 0.32 | 0.40 | 0.40 | 0.38 | 0.19 | 0.19 | 0.51 | 0.32 | 0.34 | 0.40 | 0.24 | 0.20 | 0.62 | 0.37 | 0.22 | 0.49 |
| $KGE_{ss}^{25\%}$ | 0.56 | 0.56 | 0.56 | 0.56 | 0.57 | 0.59 | 0.47 | 0.40 | 0.68 | 0.60 | 0.55 | 0.61 | 0.48 | 0.44 | 0.72 | 0.61 | 0.48 | 0.67 |
| $KGE_{ss}^{50\%}$ | 0.68 | 0.68 | 0.69 | 0.68 | 0.68 | 0.70 | 0.64 | 0.67 | 0.78 | 0.77 | 0.66 | 0.77 | 0.69 | 0.68 | 0.81 | 0.77 | 0.69 | 0.79 |
| $KGE_{ss}^{75\%}$ | 0.77 | 0.77 | 0.78 | 0.77 | 0.77 | 0.78 | 0.74 | 0.78 | 0.86 | 0.85 | 0.77 | 0.86 | 0.77 | 0.77 | 0.87 | 0.86 | 0.77 | 0.85 |
| $KGE_{ss}^{95\%}$ | 0.86 | 0.86 | 0.85 | 0.86 | 0.86 | 0.86 | 0.83 | 0.83 | 0.93 | 0.91 | 0.86 | 0.92 | 0.83 | 0.83 | 0.93 | 0.91 | 0.86 | 0.92 |
| | | | | | | | $RMSE$ | | | | | | | | | | | |
| $KGE_{ss}^{worst}$ | 2.18 | 2.18 | 2.18 | 2.25 | 2.29 | 2.25 | 2.66 | 2.38 | 1.84 | 1.88 | 2.38 | 1.83 | 2.62 | 2.30 | 1.77 | 1.84 | 2.61 | 1.80 |
| $KGE_{ss}^{5\%}$ | 2.14 | 2.14 | 2.17 | 2.15 | 2.17 | 2.21 | 2.48 | 2.29 | 1.70 | 1.75 | 2.25 | 1.75 | 2.38 | 2.24 | 1.62 | 1.77 | 2.36 | 1.68 |
| $KGE_{ss}^{25\%}$ | 1.54 | 1.54 | 1.55 | 1.55 | 1.55 | 1.65 | 1.68 | 1.64 | 1.25 | 1.33 | 1.65 | 1.32 | 1.61 | 1.62 | 1.24 | 1.31 | 1.55 | 1.29 |
| $KGE_{ss}^{50\%}$ | 1.19 | 1.18 | 1.22 | 1.14 | 1.13 | 1.21 | 1.31 | 1.26 | 0.98 | 1.07 | 1.20 | 1.05 | 1.26 | 1.27 | 0.92 | 1.05 | 1.25 | 1.02 |
| $KGE_{ss}^{75\%}$ | 0.96 | 0.96 | 0.93 | 0.94 | 0.95 | 0.91 | 1.10 | 1.06 | 0.81 | 0.88 | 0.97 | 0.88 | 1.03 | 1.05 | 0.74 | 0.86 | 1.04 | 0.81 |
| $KGE_{ss}^{95\%}$ | 0.67 | 0.68 | 0.65 | 0.66 | 0.66 | 0.60 | 0.77 | 0.84 | 0.61 | 0.65 | 0.68 | 0.65 | 0.71 | 0.78 | 0.54 | 0.64 | 0.71 | 0.62 |
| | | | | | | | $MAE$ | | | | | | | | | | | |
| $KGE_{ss}^{worst}$ | 1.11 | 1.11 | 1.10 | 1.12 | 1.11 | 1.09 | 1.39 | 1.20 | 0.98 | 1.04 | 1.11 | 1.02 | 1.31 | 1.12 | 0.90 | 1.00 | 1.23 | 0.93 |
| $KGE_{ss}^{5\%}$ | 1.08 | 1.08 | 1.06 | 1.11 | 1.10 | 1.05 | 1.25 | 1.16 | 0.95 | 0.99 | 1.07 | 0.98 | 1.17 | 1.10 | 0.86 | 0.96 | 1.09 | 0.89 |
| $KGE_{ss}^{25\%}$ | 0.79 | 0.78 | 0.77 | 0.78 | 0.77 | 0.79 | 0.83 | 0.80 | 0.59 | 0.68 | 0.74 | 0.69 | 0.79 | 0.80 | 0.52 | 0.66 | 0.79 | 0.60 |
| $KGE_{ss}^{50\%}$ | 0.61 | 0.61 | 0.58 | 0.61 | 0.60 | 0.57 | 0.65 | 0.62 | 0.49 | 0.56 | 0.58 | 0.55 | 0.59 | 0.63 | 0.44 | 0.54 | 0.60 | 0.48 |
| $KGE_{ss}^{75\%}$ | 0.49 | 0.48 | 0.47 | 0.49 | 0.49 | 0.44 | 0.55 | 0.47 | 0.39 | 0.45 | 0.45 | 0.44 | 0.47 | 0.52 | 0.35 | 0.43 | 0.48 | 0.38 |
| $KGE_{ss}^{95\%}$ | 0.35 | 0.35 | 0.34 | 0.35 | 0.36 | 0.33 | 0.38 | 0.35 | 0.30 | 0.34 | 0.35 | 0.35 | 0.33 | 0.36 | 0.24 | 0.34 | 0.35 | 0.29 |

**Table S3:** Summary skill metrics for the benchmark models reported in this study.

| Percentiles | HyMOD Like | GR4J | SACSMA | $MA_1$ | $MA_5$ | $MA_5BP_2$ | $MA_5MR_{gw}^{\sigma}$ | LSTM(2) | LSTM(3) | LSTM(5) | LSTM(6) |
|---|---|---|---|---|---|---|---|---|---|---|---|
| | | | | | $KGE$ | | | | | | |
| $KGE_{ss}^{worst}$ | -0.25 | -0.05 | 0.26 | 0.01 | 0.40 | 0.47 | 0.44 | 0.55 | 0.52 | 0.52 | 0.55 |
| $KGE_{ss}^{5\%}$ | -0.07 | 0.16 | 0.39 | 0.27 | 0.42 | 0.49 | 0.49 | 0.60 | 0.64 | 0.68 | 0.69 |
| $KGE_{ss}^{25\%}$ | 0.22 | 0.56 | 0.66 | 0.68 | 0.69 | 0.75 | 0.69 | 0.76 | 0.78 | 0.78 | 0.77 |
| $KGE_{ss}^{50\%}$ | 0.34 | 0.67 | 0.73 | 0.77 | 0.77 | 0.84 | 0.77 | 0.81 | 0.83 | 0.85 | 0.86 |
| $KGE_{ss}^{75\%}$ | 0.43 | 0.73 | 0.78 | 0.82 | 0.84 | 0.88 | 0.83 | 0.89 | 0.89 | 0.91 | 0.91 |
| $KGE_{ss}^{95\%}$ | 0.55 | 0.82 | 0.84 | 0.88 | 0.88 | 0.90 | 0.90 | 0.93 | 0.94 | 0.97 | 0.95 |
| | | | | | $A^{KGE}$ | | | | | | |
| $KGE_{ss}^{worst}$ | 0.30 | 0.02 | 0.49 | 0.45 | 0.59 | 0.66 | 0.63 | 0.64 | 0.62 | 0.65 | 0.68 |
| $KGE_{ss}^{5\%}$ | 0.49 | 0.19 | 0.63 | 0.58 | 0.68 | 0.69 | 0.65 | 0.67 | 0.69 | 0.78 | 0.74 |
| $KGE_{ss}^{25\%}$ | 0.65 | 0.64 | 0.79 | 0.79 | 0.83 | 0.84 | 0.85 | 0.86 | 0.84 | 0.87 | 0.87 |
| $KGE_{ss}^{50\%}$ | 0.74 | 0.85 | 0.87 | 0.89 | 0.91 | 0.92 | 0.92 | 0.93 | 0.91 | 0.91 | 0.94 |
| $KGE_{ss}^{75\%}$ | 0.90 | 0.92 | 0.93 | 0.94 | 0.94 | 0.96 | 0.97 | 0.94 | 0.96 | 0.96 | 0.96 |
| $KGE_{ss}^{95\%}$ | 0.97 | 0.97 | 0.99 | 0.98 | 0.99 | 0.99 | 0.99 | 0.99 | 0.99 | 0.99 | 1.00 |
| | | | | | $B^{KGE}$ | | | | | | |
| $KGE_{ss}^{worst}$ | -0.05 | 0.67 | 0.63 | 0.30 | 0.60 | 0.62 | 0.67 | 0.76 | 0.71 | 0.67 | 0.70 |
| $KGE_{ss}^{5\%}$ | 0.05 | 0.70 | 0.73 | 0.51 | 0.64 | 0.68 | 0.72 | 0.79 | 0.75 | 0.75 | 0.76 |
| $KGE_{ss}^{25\%}$ | 0.38 | 0.76 | 0.81 | 0.80 | 0.83 | 0.85 | 0.84 | 0.85 | 0.84 | 0.85 | 0.85 |
| $KGE_{ss}^{50\%}$ | 0.53 | 0.86 | 0.88 | 0.88 | 0.90 | 0.90 | 0.90 | 0.93 | 0.92 | 0.92 | 0.93 |
| $KGE_{ss}^{75\%}$ | 0.70 | 0.93 | 0.92 | 0.95 | 0.94 | 0.97 | 0.94 | 0.96 | 0.96 | 0.96 | 0.96 |
| $KGE_{ss}^{95\%}$ | 0.93 | 0.98 | 0.95 | 0.98 | 0.98 | 0.99 | 0.98 | 0.99 | 0.99 | 0.99 | 0.99 |
| | | | | | $r^{KGE}$ | | | | | | |
| $KGE_{ss}^{worst}$ | 0.50 | 0.68 | 0.47 | 0.53 | 0.55 | 0.69 | 0.58 | 0.75 | 0.86 | 0.85 | 0.88 |
| $KGE_{ss}^{5\%}$ | 0.53 | 0.69 | 0.60 | 0.65 | 0.63 | 0.83 | 0.63 | 0.78 | 0.88 | 0.87 | 0.89 |
| $KGE_{ss}^{25\%}$ | 0.60 | 0.78 | 0.78 | 0.81 | 0.77 | 0.87 | 0.78 | 0.84 | 0.92 | 0.92 | 0.92 |
| $KGE_{ss}^{50\%}$ | 0.64 | 0.81 | 0.83 | 0.86 | 0.86 | 0.92 | 0.86 | 0.92 | 0.95 | 0.95 | 0.95 |
| $KGE_{ss}^{75\%}$ | 0.68 | 0.85 | 0.86 | 0.90 | 0.89 | 0.94 | 0.89 | 0.96 | 0.97 | 0.97 | 0.97 |
| $KGE_{ss}^{95\%}$ | 0.74 | 0.92 | 0.91 | 0.93 | 0.92 | 0.97 | 0.92 | 0.98 | 0.98 | 0.99 | 0.98 |

Numbers in *red color* indicate skill < 0.5, and numbers in *blue color* indicate $0.85 \leq skill \leq 1.0$.

**Table S3:** Summary skill metrics for the benchmark models reported in this study (Part 2). For $NSE$, numbers in *red color* indicate skill $< 0.5$, and numbers in *blue color* indicate $0.85 \leq skill \leq 1.0$.

| Percentiles | HyMOD Like | GR4J | SACSMA | $MA_1$ | $MA_5$ | $MA_5BP_2$ | $MA_5MR_{gw}^\sigma$ | LSTM(2) | LSTM(3) | LSTM(5) | LSTM(6) |
|---|---|---|---|---|---|---|---|---|---|---|---|
| | | | | | | $NSE$ | | | | | |
| $KGE_{ss}^{worst}$ | -1.17 | -1.18 | 0.11 | 0.03 | 0.21 | 0.46 | 0.08 | 0.51 | 0.68 | 0.70 | 0.73 |
| $KGE_{ss}^{5\%}$ | -0.35 | -0.43 | 0.29 | 0.30 | 0.24 | 0.62 | 0.24 | 0.58 | 0.72 | 0.73 | 0.74 |
| $KGE_{ss}^{25\%}$ | 0.05 | 0.18 | 0.49 | 0.56 | 0.48 | 0.72 | 0.52 | 0.68 | 0.81 | 0.81 | 0.82 |
| $KGE_{ss}^{50\%}$ | 0.32 | 0.53 | 0.61 | 0.68 | 0.69 | 0.81 | 0.69 | 0.82 | 0.89 | 0.87 | 0.89 |
| $KGE_{ss}^{75\%}$ | 0.41 | 0.66 | 0.70 | 0.77 | 0.77 | 0.87 | 0.77 | 0.90 | 0.93 | 0.94 | 0.93 |
| $KGE_{ss}^{95\%}$ | 0.47 | 0.84 | 0.82 | 0.86 | 0.83 | 0.93 | 0.84 | 0.95 | 0.96 | 0.96 | 0.96 |
| | | | | | | $RMSE$ | | | | | |
| $KGE_{ss}^{worst}$ | 4.67 | 2.56 | 2.89 | 2.18 | 2.62 | 1.77 | 2.62 | 1.82 | 1.71 | 1.53 | 1.43 |
| $KGE_{ss}^{5\%}$ | 4.24 | 2.49 | 2.85 | 2.14 | 2.38 | 1.62 | 2.41 | 1.63 | 1.37 | 1.28 | 1.25 |
| $KGE_{ss}^{25\%}$ | 2.48 | 2.25 | 1.76 | 1.54 | 1.61 | 1.24 | 1.56 | 1.01 | 0.84 | 0.80 | 0.84 |
| $KGE_{ss}^{50\%}$ | 1.83 | 1.65 | 1.37 | 1.19 | 1.26 | 0.92 | 1.24 | 0.85 | 0.70 | 0.68 | 0.70 |
| $KGE_{ss}^{75\%}$ | 1.44 | 1.20 | 1.06 | 0.96 | 1.03 | 0.74 | 1.01 | 0.72 | 0.60 | 0.59 | 0.55 |
| $KGE_{ss}^{95\%}$ | 1.14 | 0.80 | 0.72 | 0.67 | 0.71 | 0.54 | 0.71 | 0.56 | 0.40 | 0.41 | 0.40 |
| | | | | | | $MAE$ | | | | | |
| $KGE_{ss}^{worst}$ | 2.24 | 1.35 | 1.26 | 1.11 | 1.31 | 0.90 | 1.22 | 0.86 | 0.73 | 0.72 | 0.67 |
| $KGE_{ss}^{5\%}$ | 2.03 | 1.29 | 1.18 | 1.08 | 1.17 | 0.86 | 1.13 | 0.74 | 0.63 | 0.66 | 0.64 |
| $KGE_{ss}^{25\%}$ | 1.49 | 0.93 | 0.89 | 0.79 | 0.79 | 0.52 | 0.74 | 0.51 | 0.42 | 0.41 | 0.39 |
| $KGE_{ss}^{50\%}$ | 1.16 | 0.70 | 0.62 | 0.61 | 0.59 | 0.44 | 0.55 | 0.44 | 0.35 | 0.33 | 0.34 |
| $KGE_{ss}^{75\%}$ | 0.92 | 0.51 | 0.51 | 0.49 | 0.47 | 0.35 | 0.45 | 0.36 | 0.30 | 0.28 | 0.30 |
| $KGE_{ss}^{95\%}$ | 0.72 | 0.38 | 0.37 | 0.35 | 0.33 | 0.24 | 0.32 | 0.29 | 0.24 | 0.21 | 0.21 |

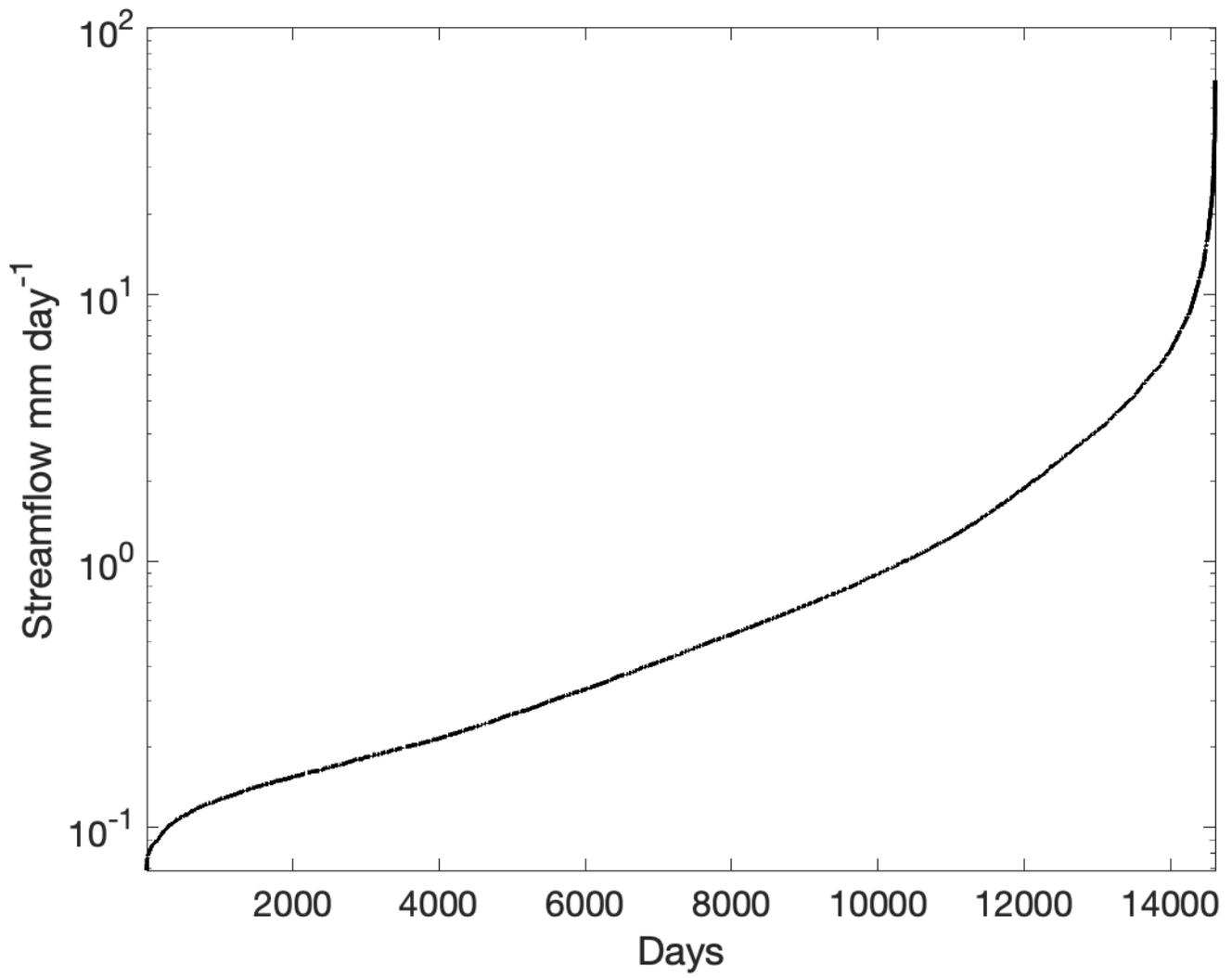

**Figure S1:** Log-scale flow-duration (sorted flow) curve using 40-year daily streamflow data in Leaf River Basin

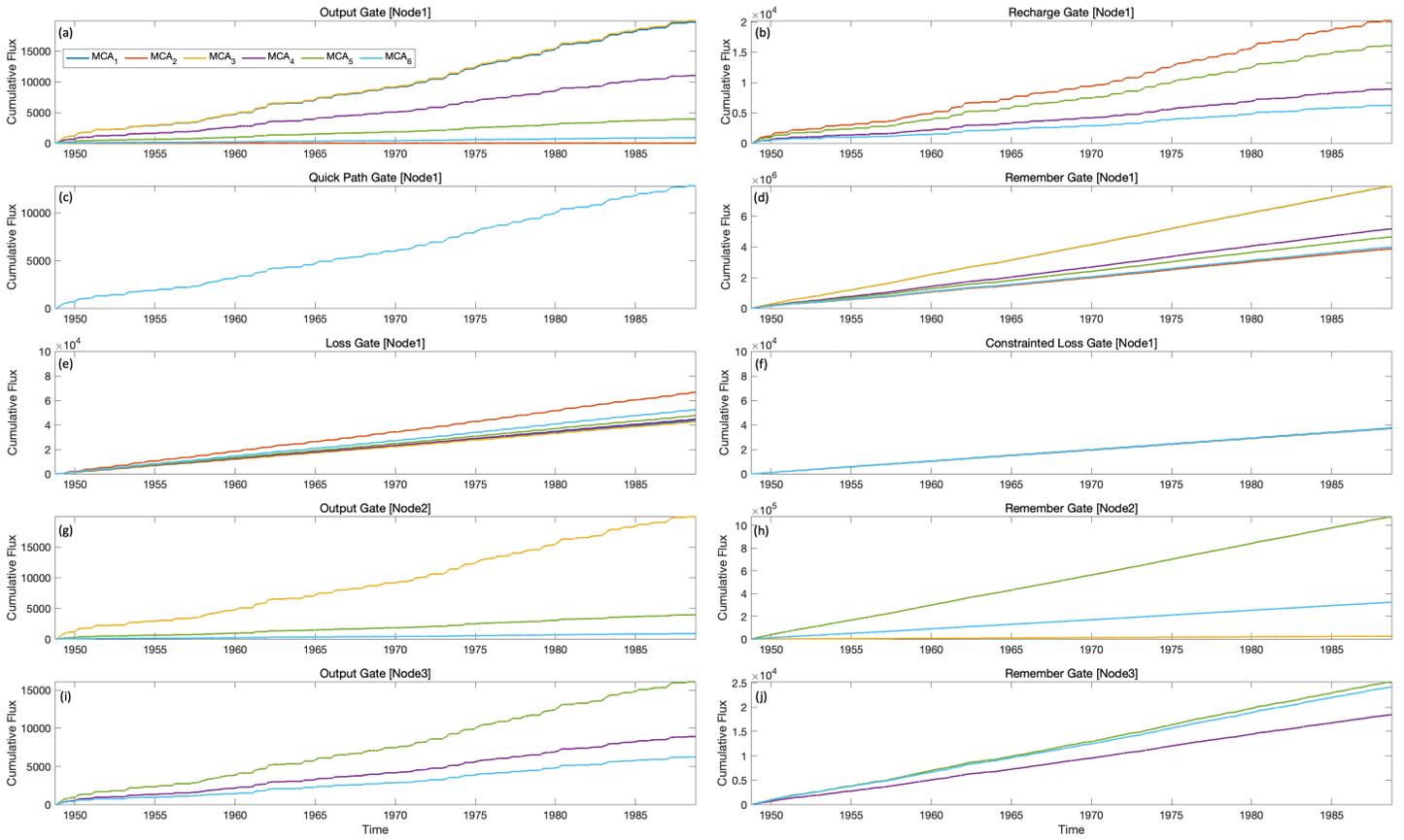

**Figure S2:** Accumulative gate flux over time of each Mass-Conserving Architecture (MA) unit including the flux of a) output gate, b) recharge gate, c) quick path gate, d) remember gate, e) loss gate, and f) constraint loss gate of the node 1, and the g) output gate, h) remember gate of node 2, and the i) output gate, j) remember gate of node 3.

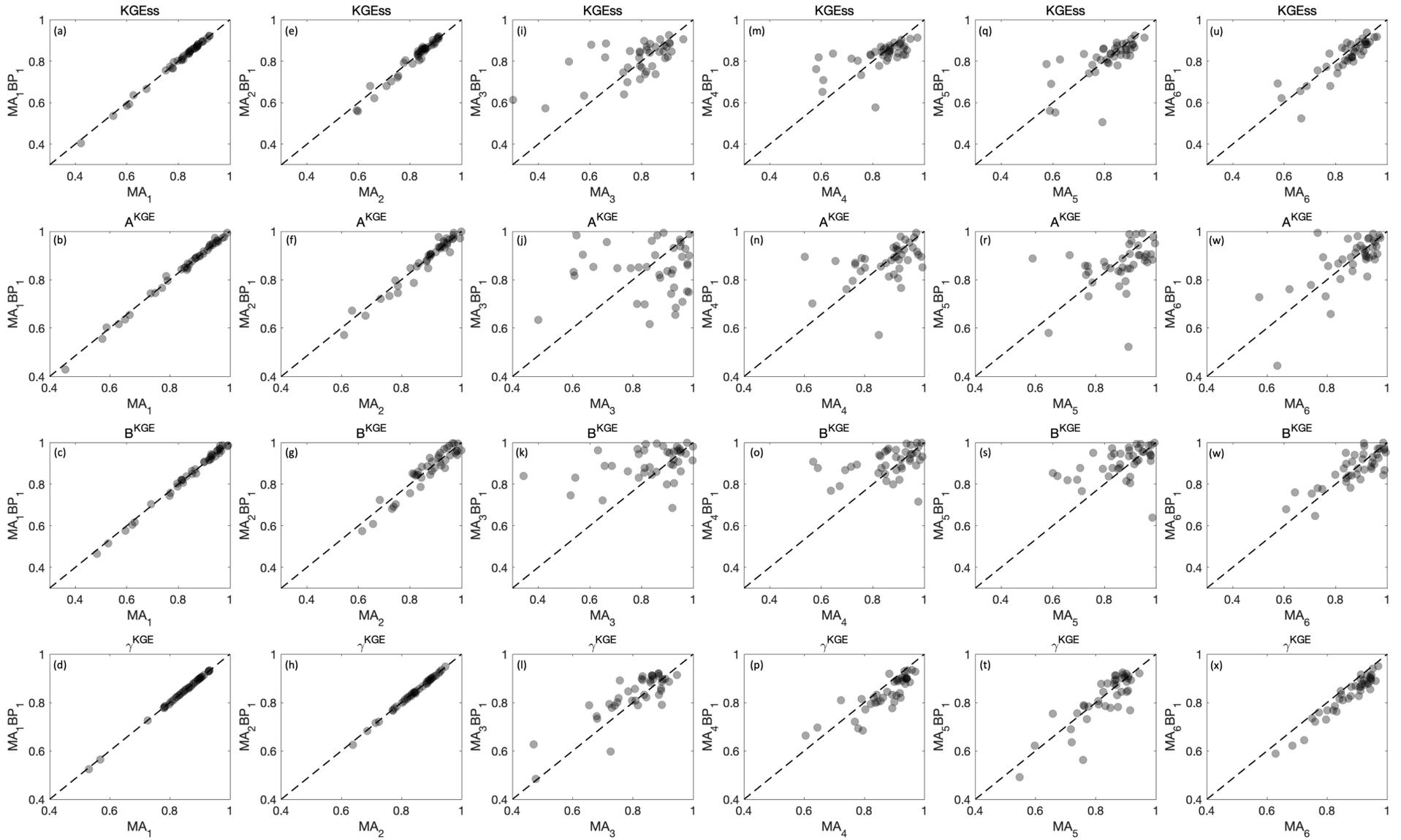

**Figure S3:** Scatter plots for the annual performance skill of $KGE_{ss}$, $A^{KGE}$, $B^{KGE}$, $\gamma^{KGE}$ between $MA_1$ and $MA_3BP_1$ in subplot from (a) to (d), $MA_2$ and $MA_4BP_2$ in subplot from (e) to (h), $MA_3$ and $MA_3BP_1$ in subplot from (i) to (l), $MA_4$ and $MA_4BP_1$ in subplot from (m) to (p), $MA_5$ and $MA_5BP_1$ in subplot from (q) to (t), and $MA_6$ and $MA_6BP_1$ in subplot from (u) to (x).

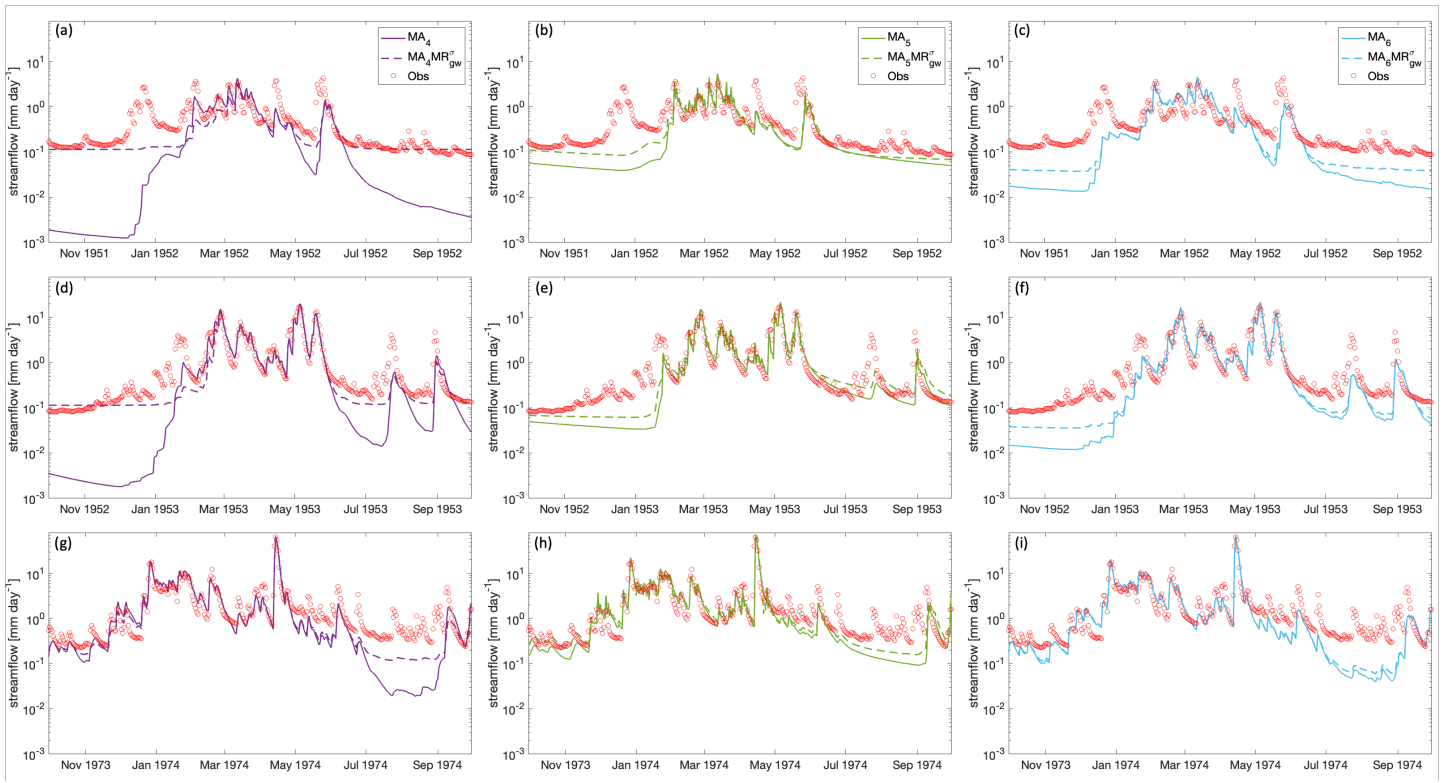

**Figure S4:** Comparison of hydrograph with the implementation of $MR_{gw}^\sigma$ on $MA_4$ to $MA_6$ architecture in dry year WY 1952 (subplot a to c), median year WY 1953 (subplot d to f), and wet year WY 1974 (subplot g to i).

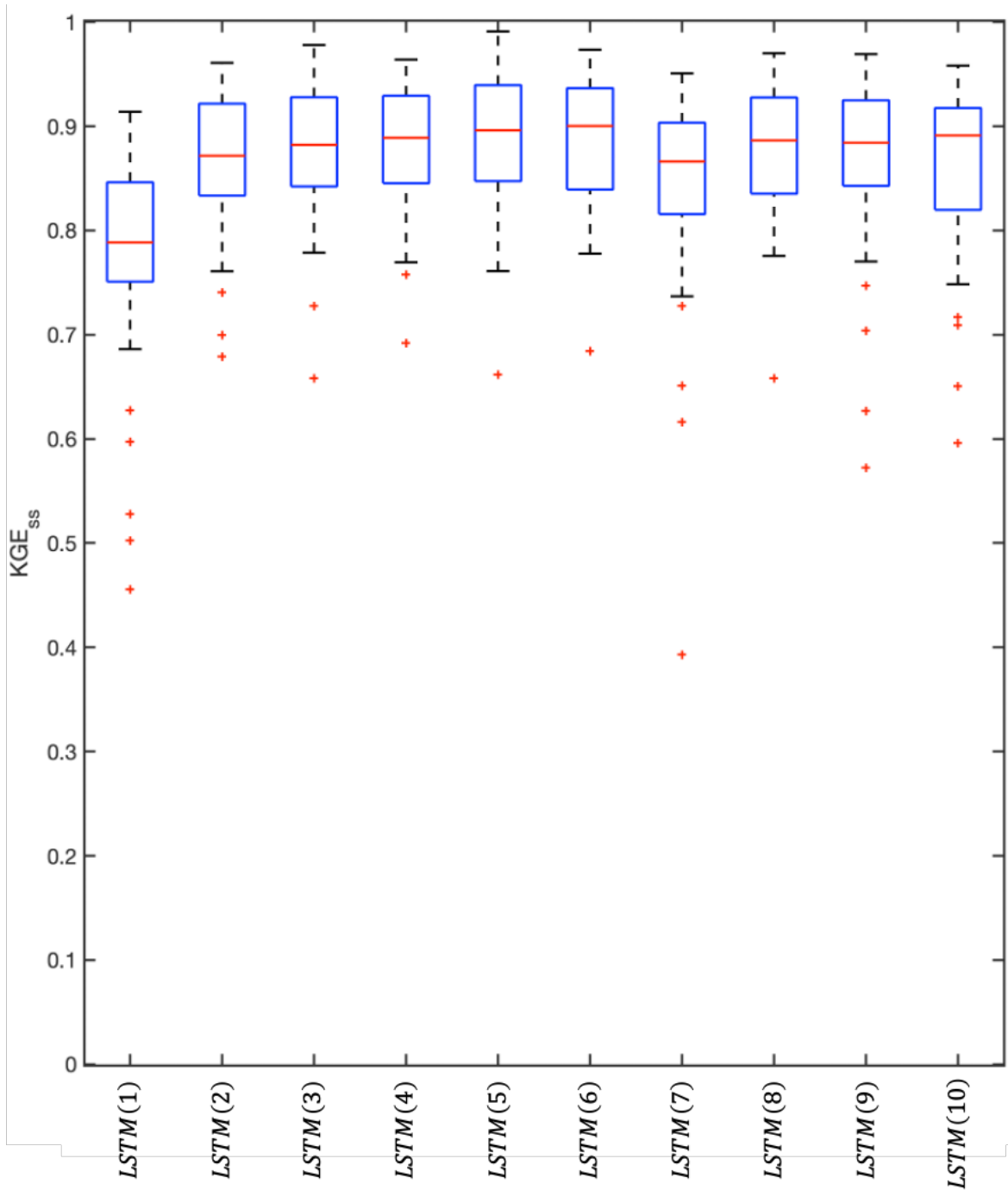

**Figure S5:** Box and whisker plots of the 40-year distributions of annual $KGE_{ss}$ performance metric values for various node numbers ($N_H = 1$ to $10$) of single-layer $LSTM$ networks.

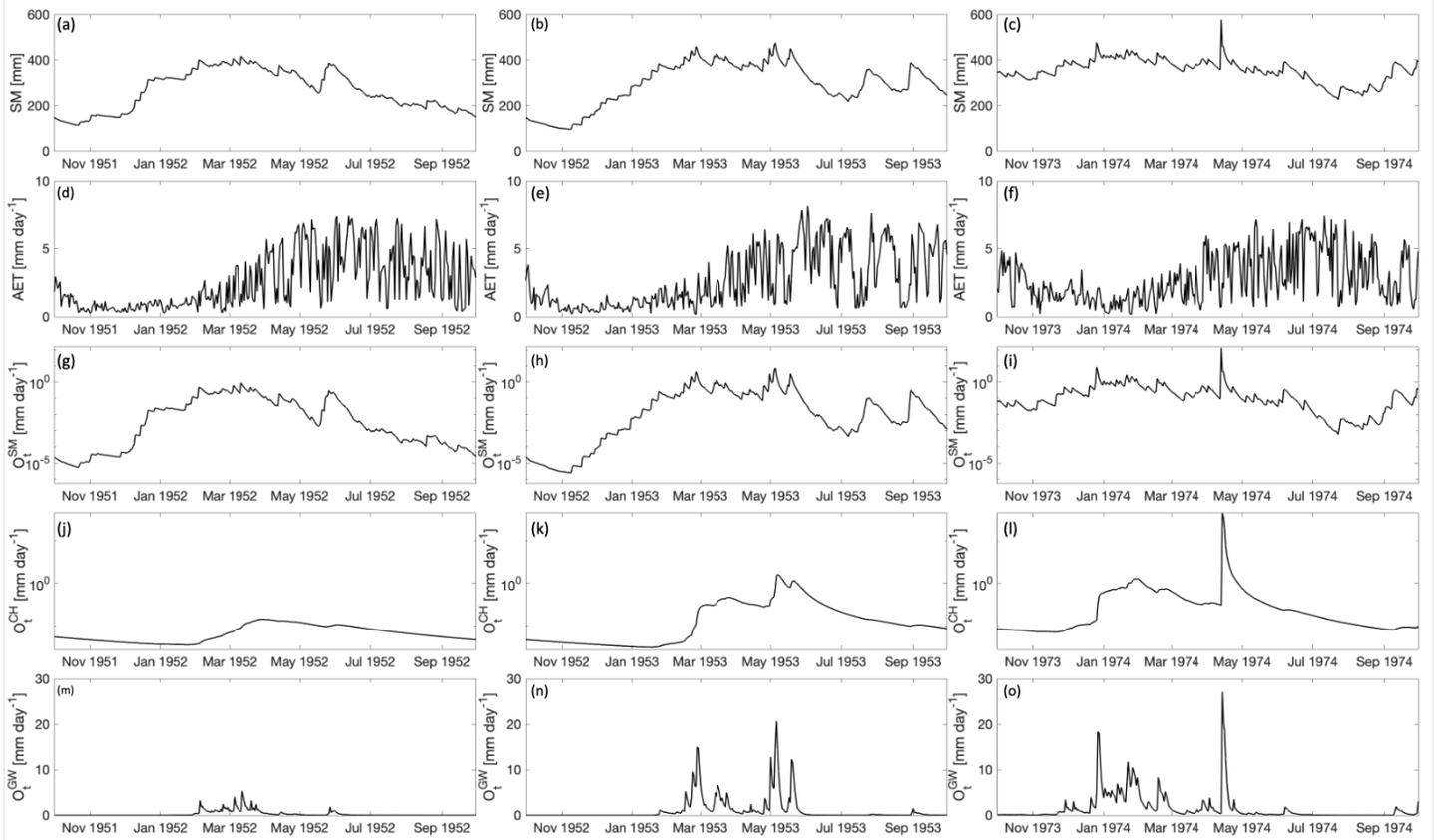

**Figure S6:** Time series plots of the $MA_5$ architecture including soil moisture ($SM$) state in dry, median, and wet year from subplot (a) to (c), actual evapotranspirative loss ($AET$) from subplot (d) to (f), outflow from soil-moisture tank ($O_t^{SM}$) from subplot (g) to (i), outflow from surface channel routing tank ($O_t^{CH}$) from subplot (j) to (l), outflow from groundwater tank ($O_t^{GW}$) from subplot (m) to (o).

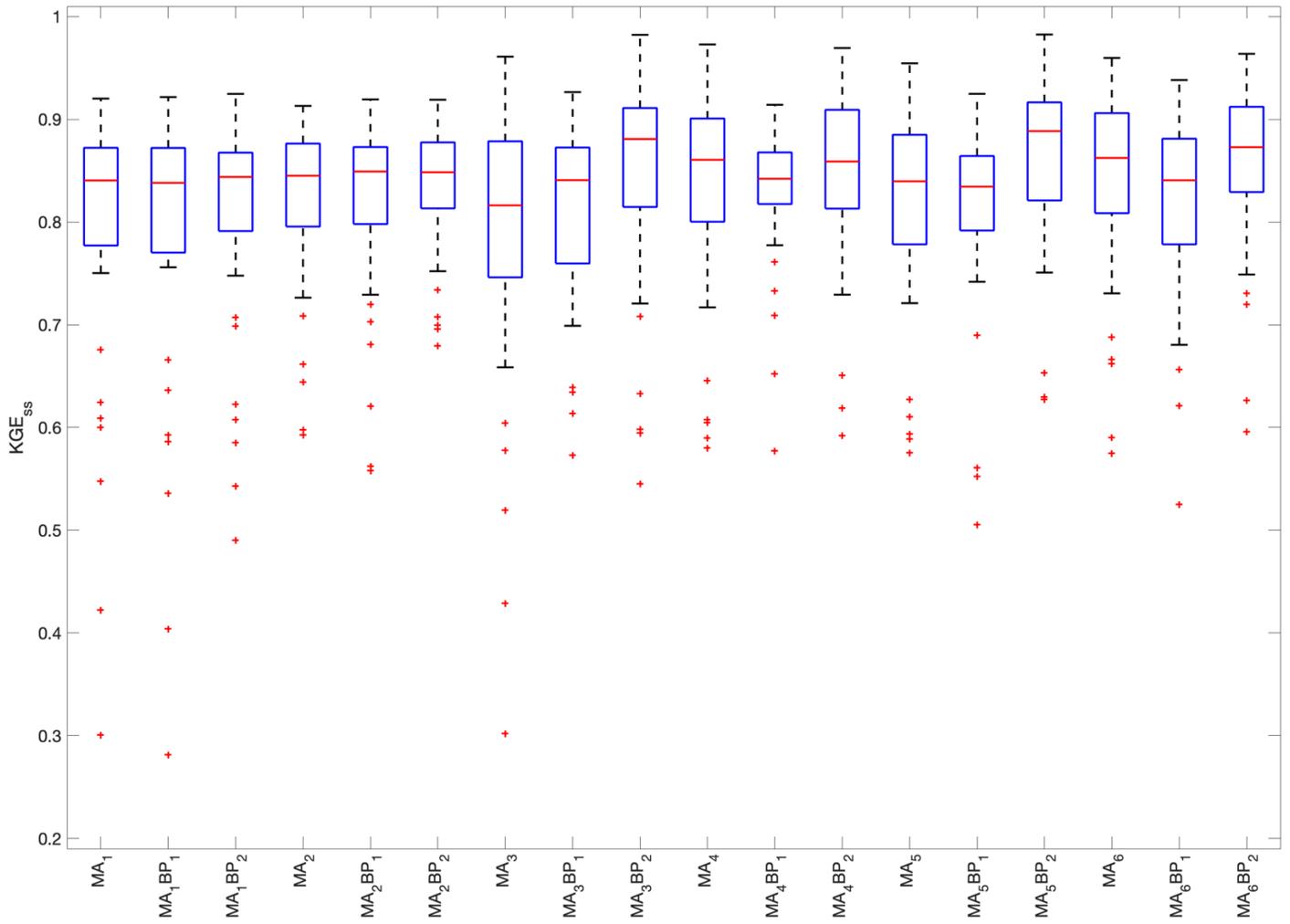

**Figure S7:** Box and whisker plots of distributions of annual $KGE_{ss}$ values for incorporating input-bypass gate for the six mass-conserving architectures.